\documentclass[10pt,twocolumn]{article}

\usepackage{cvpr}

\usepackage{framed,multirow}

\usepackage{graphicx}

\usepackage{amsmath,amssymb}
\usepackage{latexsym}

\usepackage{url}
\usepackage{xcolor}

\usepackage[labelformat=empty]{subfig}
\usepackage[export]{adjustbox}

\usepackage{array}
\usepackage{color,soul} % highlighting
\soulregister\ref{7} % make soul work with references
\soulregister\cite{7} % make soul work with citations

\usepackage{lineno}
\modulolinenumbers[5]
\usepackage[colorlinks,citecolor=blue]{hyperref}

\usepackage{natbib}

\cvprfinalcopy % *** Uncomment this line for the final submission

\usepackage{enumitem}
\setlist{nolistsep,leftmargin=*}

\newcommand{\msdspleen}{\textit{MSD-Spleen}}
\newcommand{\mospleen}{\textit{MO-Spleen}}
\newcommand{\moliver}{\textit{MO-Liver}}
\newcommand{\mopancreas}{\textit{MO-Pancreas}}
\newcommand{\molkidney}{\textit{MO-L.Kidney}}
\newcommand{\mogallbladder}{\textit{MO-Gallbladder}}

\begin{document}
\title{Going to Extremes: Weakly Supervised Medical Image Segmentation}

%%
%% The "author" command and its associated commands are used to define the authors and their affiliations.
%Holger R. Roth, Dong Yang, Ziyue Xu, Xiaosong Wang, Daguang Xu
\author{Holger R. Roth, Dong Yang, Ziyue Xu, Xiaosong Wang, Daguang Xu \\
\\
NVIDIA, Bethesda, USA \\
\\
{\tt\small Contact: hroth@nvidia.com, daguangx@nvidia.com}}

\maketitle

\begin{abstract}
\noindent Medical image annotation is a major hurdle for developing precise and robust machine learning models. Annotation is expensive, time-consuming, and often requires expert knowledge, particularly in the medical field. Here, we suggest using minimal user interaction in the form of extreme point clicks to train a segmentation model which, in effect, can be used to speed up medical image annotation. An initial segmentation is generated based on the extreme points utilizing the random walker algorithm. This initial segmentation is then used as a noisy supervision signal to train a fully convolutional network that can segment the organ of interest, based on the provided user clicks. Through experimentation on several medical imaging datasets, we show that the predictions of the network can be refined using several rounds of training with the prediction from the same weakly annotated data. Further improvements are shown utilizing the clicked points within a custom-designed loss and attention mechanism. Our approach has the potential to speed up the process of generating new training datasets for the development of new machine learning and deep learning-based models for, but not exclusively, medical image analysis.
\end{abstract}

%\linenumbers
%\sloppypar

%%%%%%%%% BODY TEXT
\section{Introduction}
\label{sec:intro}
\noindent A major bottleneck for the development of novel machine learning (ML) based models is the annotation of datasets that are useful to train such models. This is especially true for healthcare applications, where annotation typically needs to be performed by experts with clinical domain knowledge. This bottleneck inhibits our ability to integrate ML-based models into clinical workflows and in order to increase their productivity. At the same time, there is a growing demand for ML methods to improve clinical image analysis workflows, driven by the growing number of medical images taken in routine clinical practice.

In particular, volumetric analysis has shown several advantages over 2D measurements for clinical applications \citep{devaraj2017nodulevolume}, which in turn, further increases the amount of data (a typical CT scan contains hundreds of slices) needing to be annotated to train accurate 3D models. 
Apart from acquiring accurate measurements, volumetric segmentation is widely desirable for visualization, 3D printing, radiomics, radiation treatment planning, image-guided surgery, and registration.
Despite the increasing need for 3D volumetric training data to train accurate and efficient ML models for medical imaging, the majority of annotation tools available today are constrained to performing the annotation in multiplanar reformatted views. 
The annotator needs to either use a virtual paint brush or draw boundaries around organs of interest, often on a slice-by-slice basis \citep{itksnap}. Classical techniques like 3D region growing or interpolation can speed up the annotation process by starting from seed points or sparsely annotated slices but its usability is often limited to certain types of structures. Some tools allow the user to skip certain regions of the image by using interpolation between slices or cross-sectional views can be helpful, but they often ignore the underlying image information. Hence, these approaches cannot always generalize to the varied use cases in medical imaging.

In this work, we propose to use only minimal user interaction in the form of extreme point clicks at the boundary of the object or organ of interest in order to train a deep learning (DL) based segmentation model. The proposed approach integrates an iterative training and refinement scheme to gradually improve the models' performance. Starting from user-defined extreme points along each dimension of a 3D medical image, an initial segmentation is produced based on the random walker (RW) algorithm \citep{grady2006random}. This segmentation is then used as a noisy supervisory signal to train a fully convolutional network (FCN) that can segment the organ of interest-based on the provided user clicks. Furthermore, we propose several variations on the deep learning setup to make full use of the extreme point information provided by the user. For example, we integrate the point information into a novel point-based loss function and combine it with an attention mechanism to further guide the segmentations. Through large-scale experimentation, we show that the network’s predictions can be iteratively refined using several rounds of training and prediction. Always using the same weakly annotated point data as our only manually provided supervision signal.
%%%%%%%%%%%%%%%%%%%%%%%%%%%%%%%%%%%%%%%%%%%%%%%%%%%%%%%%%%%%%%%%%
%%%%%%%%%%%%%%%%%%%%%%%%%%%%%%%%%%%%%%%%%%%%%%%%%%%%%%%%%%%%%%%%%
\subsection{Related work}
\paragraph{Segmentation networks}
Fully convolutional networks (FCNs) \citep{long2015fully} have established themselves as the state-of-the-art methods for medical image segmentation in recent years \citep{ronneberger2015u,milletari2016v,cciccek20163d,liu20183d,myronenko20183d}. However, a major drawback is that they are very data-hungry, limiting their application in healthcare where data annotation is very expensive. To reduce the cost of labeling, semi-automated/interactive and weakly supervised methods have been proposed in the literature \citep{guo2018review,tajbakhsh2019embracing}.

\paragraph{Interactive segmentation}
The integration of semi-automated approaches has been an active area of development \citep{an2017accuracy}, typically utilizing classical methods such as graph cut \citep{boykov2006graph}, random walks \citep{grady2006random}, active shape models \citep{van2003interactive,schwarz20073d}, and others \citep{dougherty2011medical}. Machine learning methods have also been considered as a viable way for interactive algorithms. In \citet{wang2016slic}, an online random forest is used in combination with conditional random fields and 4D graph cuts to segment, in a minimally interactive framework, the human placenta in fetal MRI scans. Recently, building on advances in deep learning, several new methods have been proposed. 
One popular form of interaction is user-drawn scribbles. In \citet{amrehn2017ui}, a user can iteratively add scribble hints as seed points to improve the segmentation result given by an FCN. In \citet{wang2018deepigeos}, the DeepIGeoS algorithm leverages geodesic distance transforms and scribbles to allow interactive segmentation. An alternative method \citep{wang2018interactive} uses image-specific fine-tuning and leverages both bounding boxes and scribble-based interaction. \citet{can2018learning} proposes to use scribbles with random walks \citep{grady2006random} and FCN predictions to achieve semi-automated segmentation. Scribbles are also used to generate pixel-level maximum category likelihood via propagation to their neighborhood in \citep{DiasWACV19}. 
Instead of scribbles, point clicks is another widely practiced interaction. In \citet{sakinis2019interactive}, the authors utilize the clicks as Gaussian kernels and put them in a separate input channel to an FCN to model user interactions via seed-point placing. \citet{KhanMICCAI19} extends the Gaussian kernel idea to a confidence map derived from extreme points that quantitatively encodes some priors. \citet{MajumderCVPR19} transforms the positive and negative clicks into images based on superpixel and object proposals, so that image information can be utilized with clicks to generate a guidance map. 
In addition to scribbles and points, \citet{LingCVPR19} parameterizes the segmentation boundary as polygons/splines, which are further modeled as a graph. Location shifts for each node are then predicted via Graph Convolutional Networks (GCN). 

\paragraph{Weakly-supervised segmentation}
Weak supervision significantly reduced the time needed for user annotation, and therefore is an important research area for DL. One popular idea is to apply classical non-learning-based methods over a DL-generated feature map. For example, in \citet{DiasACCV18}, Monte Carlo region growing is triggered from confidence scores given by a network, and in \citet{CerroneCVPR19}, random walks is performed over learnt edge weights. 
An ``opposite'' idea is to use classical unsupervised methods as initial estimate for further learning process. In \citet{rajchl2017deepcut}, an initial \textit{GrabCut} segmentation is used for this purpose, and segmentation performance is then improved with several rounds of predictions using CNN plus Dense CRF post-processing. Similarly, in \citet{zhang2018self}, segmentation results based on K-means are used to train a deep segmentation network on cystic lung regions. Without proper supervision, such approaches might work well if unsupervised techniques can have good enough initial performance. However, completely unsupervised techniques might fail to generalize to organs where the boundary information is not as clear. One possible way to address this issue is to add a confidence network \citep{NieMICCAI18} to judge the quality of additional information generated, so that unlabeled data can be included to adversarially train the segmentation network. More recently, \citet{kervadec2019constrained} introduced inequality constraints based on target-region size and image tags in the loss function of a CNN in order to train the network for weakly supervised segmentation. 
Instead of information extracted by classical methods, weakly-supervised or self-learning can also make use of measurements readily available, or use non-experts' judgements. One example is the measurements acquired during evaluation of the RECIST criteria \citep{cai2018accurate} in the hospital picture archiving and communication system (PACS). However, such measurements are typically constraint to 2D and might miss adequate constraints for more complex three-dimensional shapes. Non-expert annotations can be acquired by utilizing crowd-sourcing platform, \citet{rajchl2016learning} distributes super-pixel weak annotation tasks and collects such annotations from a crowd of non-expert raters, and further use them as weak-supervision for network training. 

\subsection{Contributions}
\noindent This work follows our preliminary study presented in \citep{roth2019weakly} which investigated a 3D extension of \citep{maninis2017deep} in a weakly supervised setting and building on random walker initialization from scribbles. In this work, we extend this approach and add the following contributions:
\begin{itemize}
    \item We utilize a modern network architecture shown to be very efficient for medical image segmentation tasks, namely the architecture proposed in \citet{myronenko20183d} and integrate the attention mechanism proposed by \citet{oktay2018attention}. 
    \item We make proper use of the point channel information not just at the input level of the network, but throughout the network, namely in the new attention gates.
    \item We furthermore propose a novel loss function that integrates the extreme point locations to encourage the boundary of our model's predictions to align with the clicked points.
    \item We extend the experimentation to a new multi-organ dataset that shows the generalizability of our approach.
\end{itemize}

%%%%%%%%%%%%%%%%%%%%%%%%%%%%%%%%%%%%%%%%%%%%%%%%%%%%%%%%%%%%%%%%%
\begin{figure}[t]
\begin{center}
    \includegraphics[width=1.0\columnwidth]{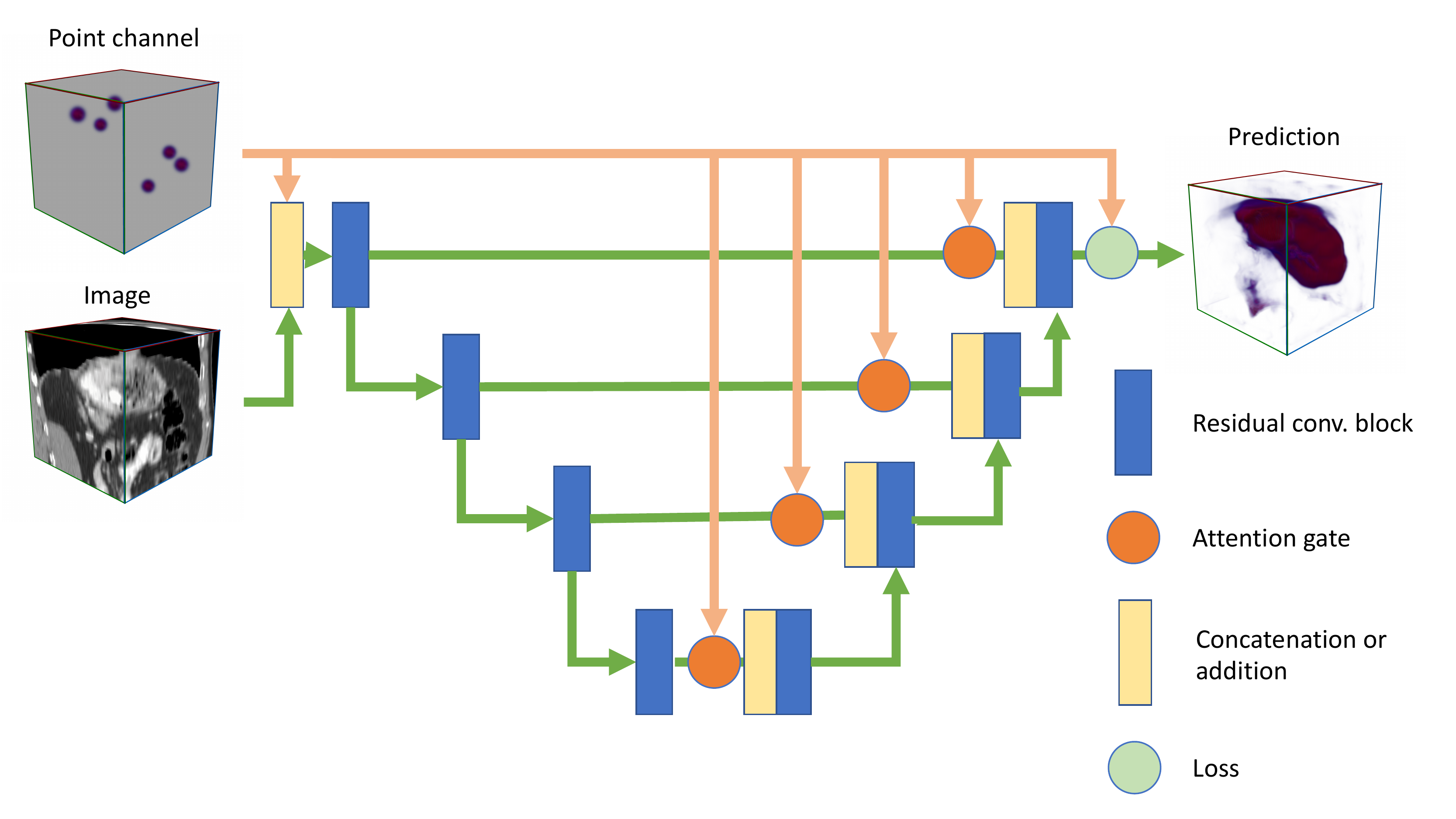}
\end{center}
\caption{A high-level overview of our proposed network architecture. The network receives both a \textit{image channel} input and a \textit{point channel} input that represents the user-provided extreme points. The point channel is then used throughout the network to further guide the segmentation training, i.e. as an additional input to attention gates and in the loss function.}
\label{fig:network}
\end{figure}
%%%%%%%%%%%%%%%%%%%%%%%%%%%%%%%%%%%%%%%%%%%%%%%%%%%%%%%%%%%%%%%%%
%
\section{Method}
\label{sec:method}
\noindent The starting point for our framework are a set of user-provided clicks on the extreme points $\{e\}$ that lie on the surface of the organ of interest. We follow the approach of \citet{maninis2017deep} and assume the users to provide only the extreme points along each image dimension in a three-dimensional medical image. This information is then utilized at several places within the network and during our iterative training scheme. The overall proposed algorithm for weakly supervised segmentation from extreme points can be divided into the steps which are detailed below:
\begin{enumerate}
    \small
    \item Extreme point selection
    \item Initial segmentation from scribbles via random walker (RW) algorithm
    \item Segmentation via deep fully convolutional network (FCN), where we explore several variations on the training scheme
    \begin{enumerate}
        \item Without RW and Dice loss
        \item With RW but without the extra point channel and Dice loss
        \item With RW and Dice loss
        \item With RW and Dice loss and point loss
        \item With RW and Dice loss and point loss and attention
        \item With RW and Dice loss and point loss and point attention
    \end{enumerate}
    \item Regularization using random walker algorithm
\end{enumerate}
Steps 2, 3, and 4 will be iterated until convergence. Here, convergence is defined based on a hold out validation set.
\subsection{Step 1: Extreme point selection}
\noindent Defining extreme points $\{e\}$ on the surface of the organ will allow the extraction of a bounding box around the organ of interest. Additional padding is typically useful to allow the network to learn some contextual information around the organ of interest.

Bounding box selection significantly reduces the image content to be analyzed and simplifies the machine learning problem, as previous work on cascaded approaches showed \citep{roth2018spatial}. 
The computer vision literature has extensively studied bounding boxes and extreme points on objects \citep{maninis2017deep}. They give some advantages over the technical drawbacks of bounding box selection in which the user often has to pick the corners of bounding boxes outside of the object of interest. 
This is particularly difficult to do for three-dimensional objects where users typically have to traverse three multi-planar reformatted views (axial, coronal, sagittal) to accomplish the task. Recent studies also demonstrated the time savings achieved with extreme point selection instead of conventional bounding box selection \citep{maninis2017deep,papadopoulos2017extreme}. 
At the same time, extreme points can provide the segmentation model with additional information which can be seen in our experimental section, Table
\ref{table:results} where we compare various ways of integrating the extreme point information into the model training. 
They lie on the surface of the object. In the basic approach, we can model them together with the image intensities as an additional input channel. This extra channel $G($\{e\}$)$ includes 3D Gaussians centered on each user clicked point location $e$. This approach is similar to \citet{maninis2017deep} but we have extended this approach to problems with 3D medical imaging. At the same time, we can utilize the point information to guide the loss function towards making predictions whose boundary aligns with the point locations (see \ref{sec:point_loss}) or use it as an additional signal that can be used to guide model attention mechanisms (see \ref{sec:point_attention}).

Figure \ref{fig:network} illustrates the different ways of how the extreme point information can be used by our proposed network architecture. We ask the user to click on six extreme points that describe the largest extent of the organ. Here, six click locations are shown after conversion to Gaussians in the extra input channel to the network, loss, and attention gates. These points are then used to compute a bounding box $B$ automatically, including some padding $p$. In this study, we extract the extreme points automatically during training from a given ground truth mask. In order to simulate user interaction, we add some Gaussian noise to the $x, y, z$ point locations at each DL training iteration as in \citet{maninis2017deep}.

After cropping the image based on $B$, we resize each bounding box region to a constant size $S=s_x \times s_y \times s_z$. In all our experiments we set $s_x = s_y = s_z = 128$ and choose $p$=20 mm which can include enough contextual information for typical applications of clinical CT scanning (see Section \ref{sec:experiments}).
%%%%%%%%%%%%%%%%%%%%%%%%%%%%%%%%%%%%%%%%%%%%%%%%%%%%%%%%%%%%%%%%%%%%%%%%%%%%%%%%%%%%%%%%%%%%%%%%
%%% OLD FIGURE!!!!!!!!
%%%%%%%%%%%%%%%%%%%%
%\begin{figure*}[htbp!]
%	\centering
%	\begin{tabular}{cccc}
%	    %\subfloat[(a)]{\rotatebox{0}{\adjincludegraphics[valign=c,width=0.32\textwidth]{figs/liver_14_2D_input_selcection}}} &
%		\subfloat[(a)]{\adjincludegraphics[valign=c,height=1.4cm]{figs/liver_14_2D_input_selcection}} &
%		\hfill
%		\subfloat[(b)]{\adjincludegraphics[valign=c,height=1.4cm]{figs/liver_14_3D_screenshot_points_w}} &  
%		\hfill
%		\subfloat[(c)]{\adjincludegraphics[valign=c,height=1.0cm]{figs/liver_14_seeds}} &		
%		\hfill
%		\subfloat[(d)]{\adjincludegraphics[valign=c,height=1.4cm]{figs/liver_14_3D_screenshot_pred_annotation}}
%	\end{tabular}
%	\caption{Our weakly supervised segmentation framework. (a) The user selects extreme points that define the organ of interest (here the liver) in 3D space. (b) Extreme points are modeled as Gaussians in an extra image channel which is fed to a 3D segmentation model. (c) Foreground scribbles are generated automatically to initialize random walker (the ground truth surface is shown in red for reference). (d) Model returns the segmentation results.
%	\label{fig:dextr3D}}
%\end{figure*}
%%%%%%%%%%%%%%%%%%%%%%%%%%%%%%%%%%%%%%%%%%%%%%%%%%%%%%%%%%%%%%%%%%%%%%%%%%%%%%%%%%%%%%%%%%%%%%%%
%%%%%%%%%%%%%%%%%%%%%%%%%%%%%%%%%%%%%%%%%%%%%%%%%%%%%%%%%%%%%%%%%%%%%%%%%%%%%%%%%%%%%%%%%%%%%%%%
\begin{figure*}[htbp!]
    \newcommand{\figheight}{1.5cm}
	\centering
	\begin{tabular}{ccccc}
	    %\subfloat[(a)]{\rotatebox{0}{\adjincludegraphics[valign=c,width=0.32\textwidth]{figs/liver_14_2D_input_selcection}}} &
		\subfloat[(a)]{\adjincludegraphics[valign=c,height=\figheight]{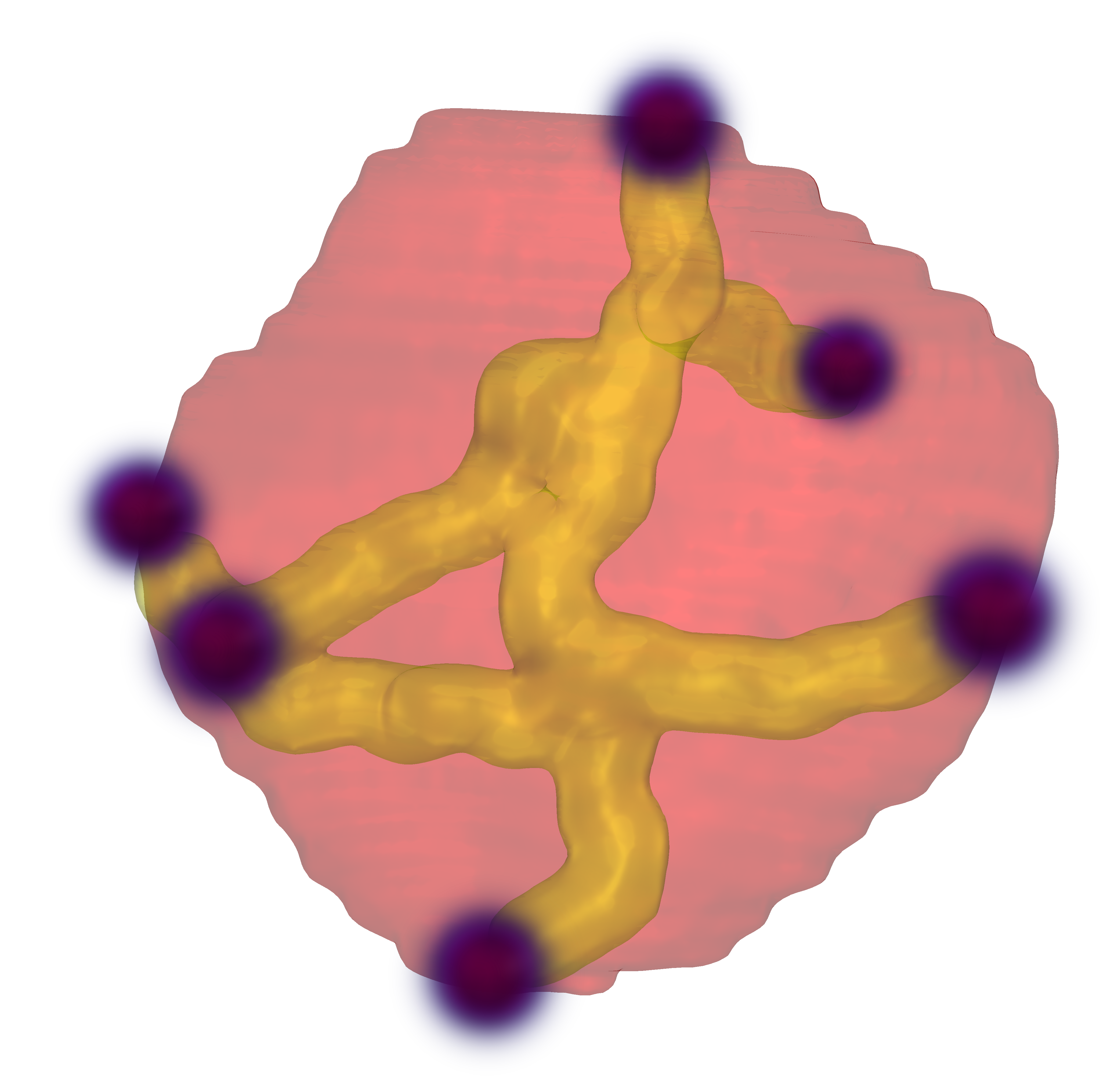}} &
		\hfill
		\subfloat[(b)]{\adjincludegraphics[valign=c,height=\figheight]{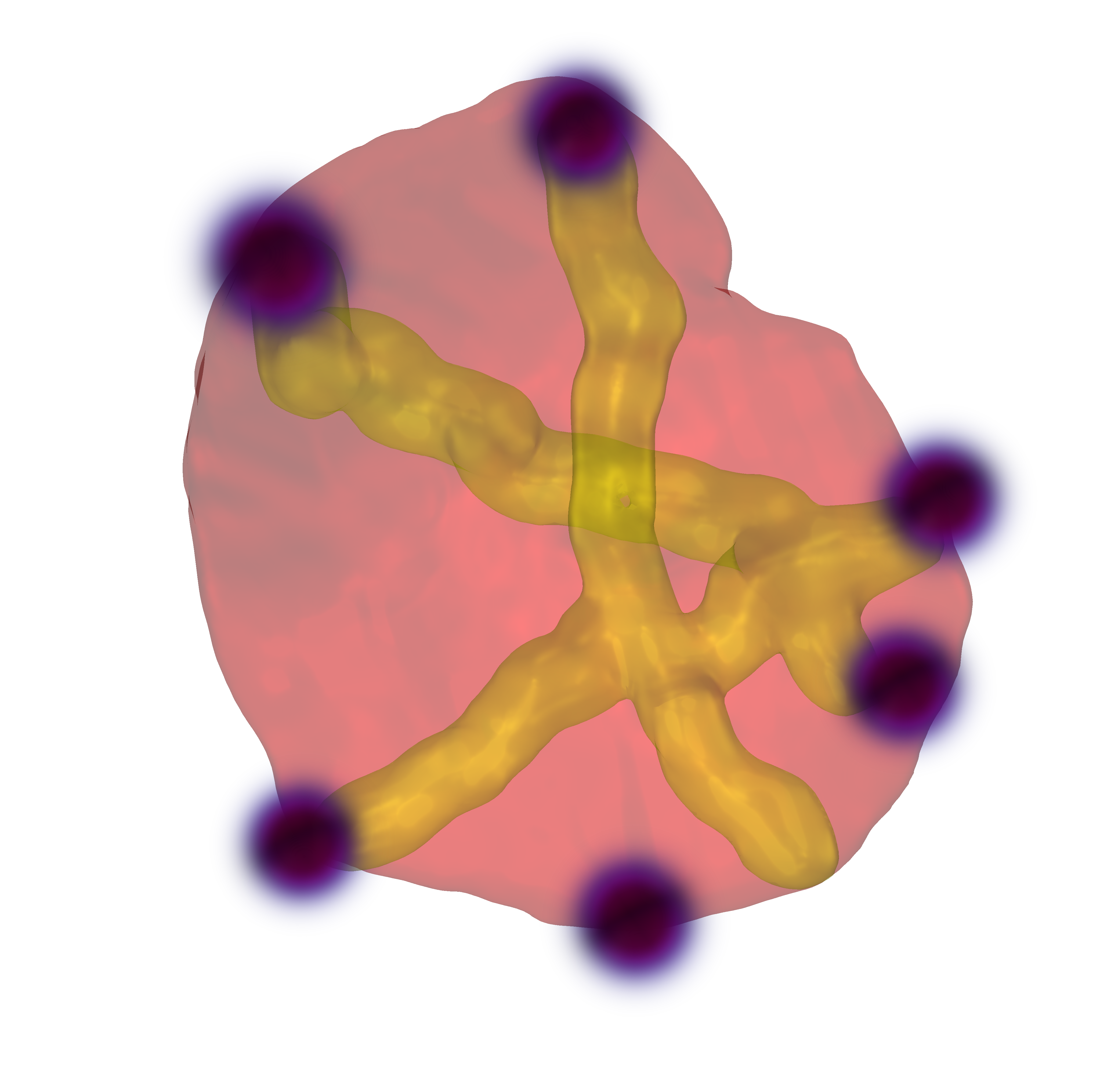}} &  
		\hfill
		\subfloat[(c)]{\adjincludegraphics[valign=c,height=\figheight]{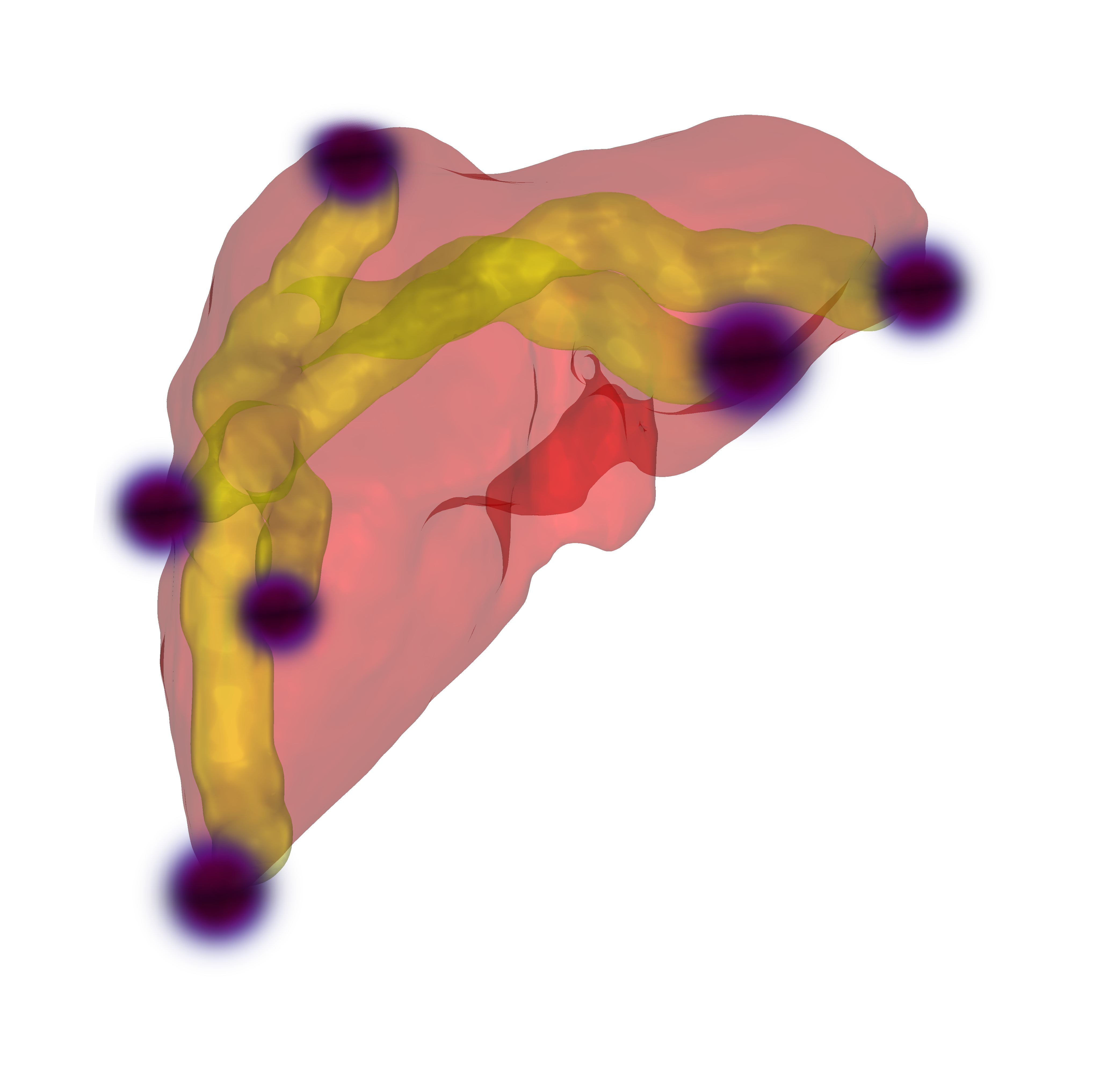}} &		
		\hfill
		\subfloat[(d)]{\adjincludegraphics[valign=c,height=\figheight]{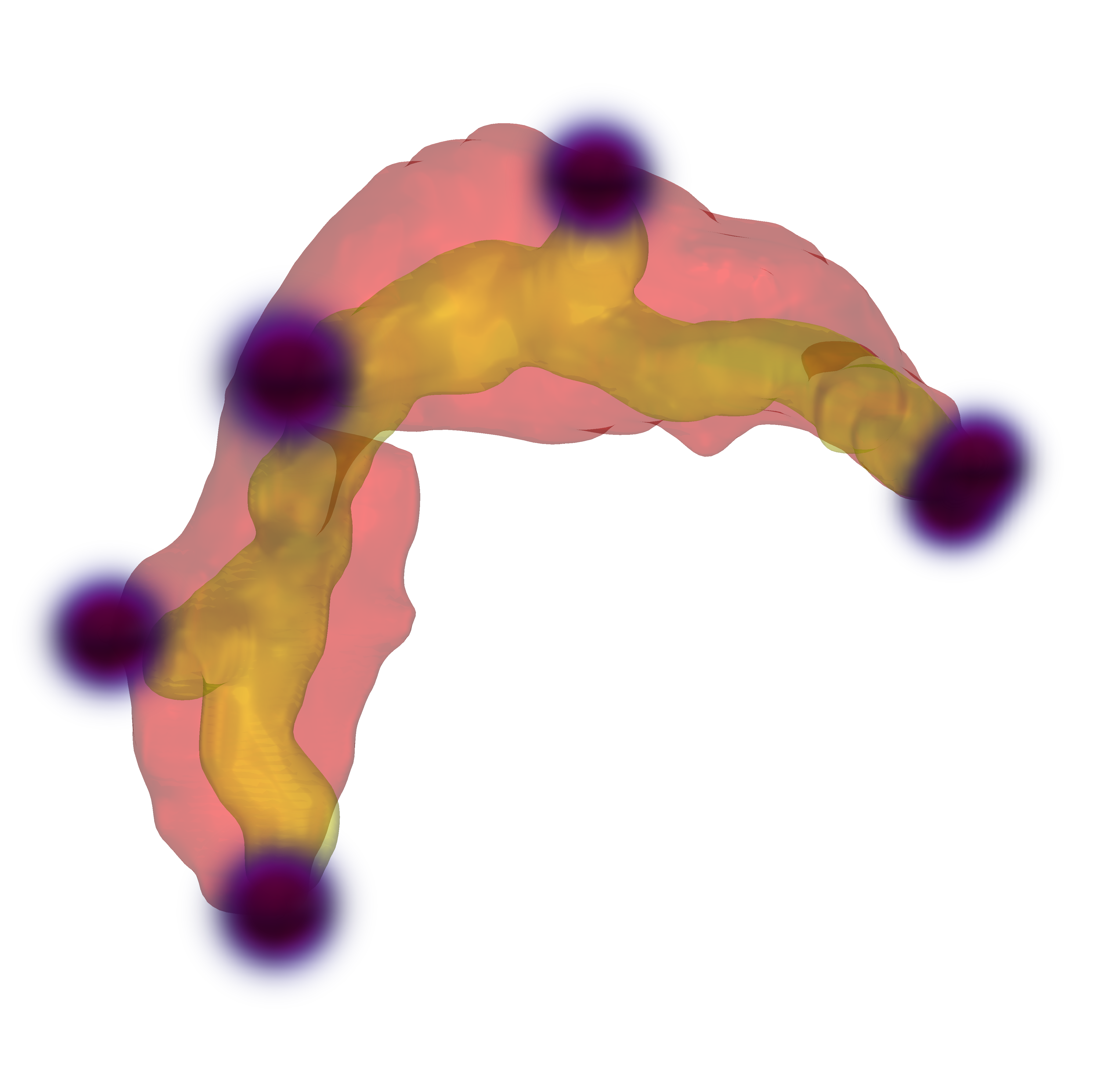}} &
		\hfill
		\subfloat[(e)]{\adjincludegraphics[valign=c,height=\figheight]{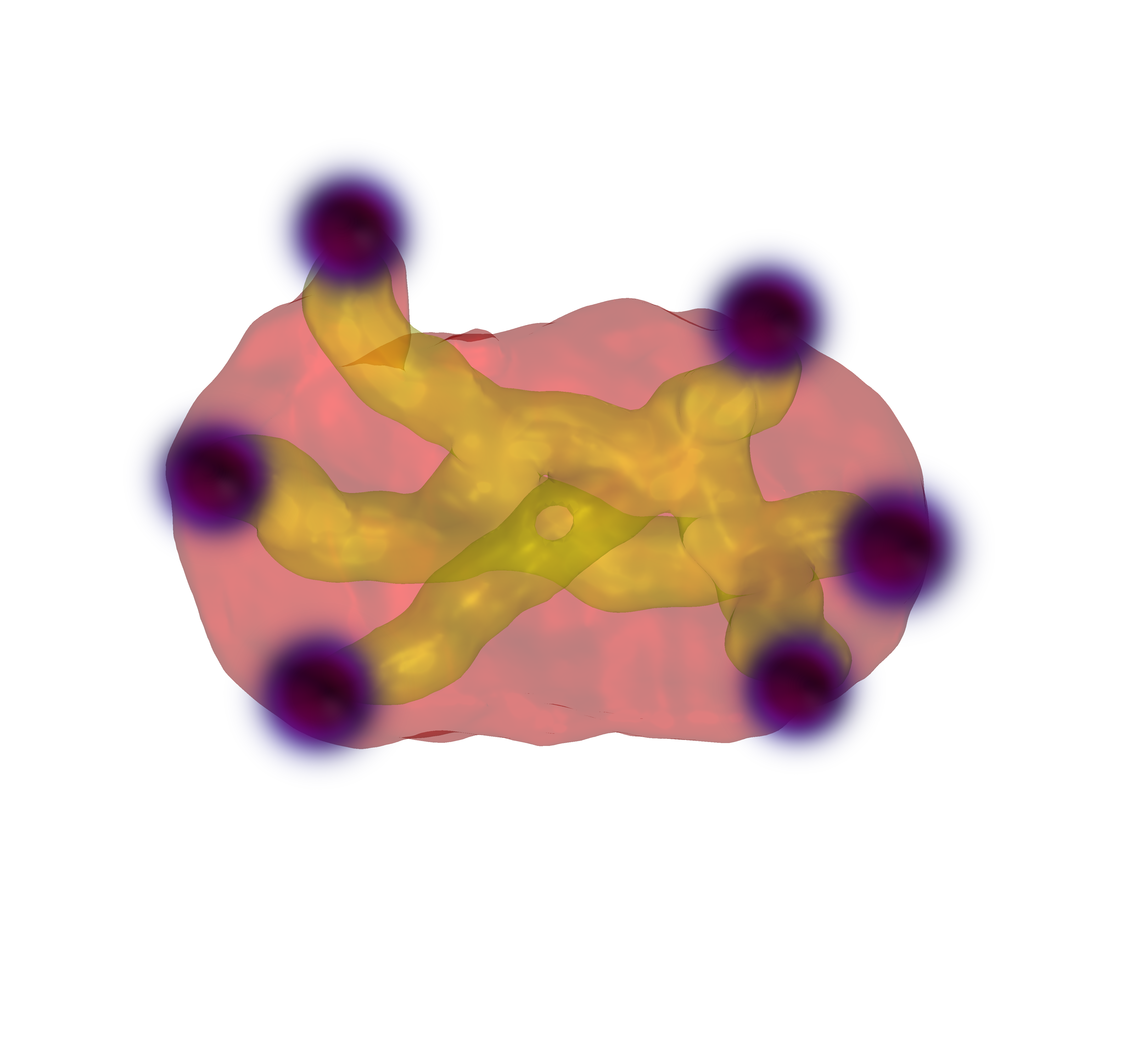}}		
	\end{tabular}
	\caption{Examples of automatically created foreground ``scribbles'' (yellow) from extreme point clicks, modelled as 3D Gaussians in our network learning. We use a geodesic shortest path algorithm to compute a scribble based on the image information alone that connects two opposing extreme points across one of the three image dimensions. (a)-(e) are showing examples from \msdspleen, \mospleen, \moliver, \mopancreas, \molkidney, and \mogallbladder, respectively. The surface rendering show the ground truth segmentations for reference in red. Best viewed in color.
	\label{fig:scribbles}}
\end{figure*}
%%%%%%%%%%%%%%%%%%%%%%%%%%%%%%%%%%%%%%%%%%%%%%%%%%%%%%%%%%%%%%%%%%%%%%%%%%%%%%%%%%%%%%%%%%%%%%%%

\subsection{Step 2: Initial segmentation from scribbles via random walker algorithm}
\noindent In this step, we turn the generated scribbles into a probability map $\hat{Y}$ that can act as a pseudo-dense  or ``noisy'' label map that can supervise a 3D deep network to learn the segmentation task. To achieve this goal, we select a set of foreground and background scribbles based on the initial set of extreme points $\{e\}$ that serve as the input seeds for the random walker algorithm \citep{grady2006random}. 

The shortest path between each extreme point pair along each image axis is computed via the Dijkstra algorithm \citep{dijkstra1959note}. Here, we model the distance between neighboring voxels by their gradient magnitude
%{\scriptsize $D = \sqrt{ \left(\frac{\partial f}{\partial x}\right)^2 + \left(\frac{\partial f}{\partial y}\right)^2 + \left(\frac{\partial f}{\partial z}\right)^2}$}.
\begin{equation}
    D = \sqrt{ \left(\frac{\partial I}{\partial x}\right)^2 + \left(\frac{\partial I}{\partial y}\right)^2 + \left(\frac{\partial I}{\partial z}\right)^2},
    \label{eq:dijkstra}
\end{equation}
where $I$ denotes the image intensity.
The resulting path can be seen as an approximation of the geodesic distance between the two extreme points in each dimension \citep{wang2018deepigeos} with respect to the content of the image. Figure \ref{fig:scribbles} displays the foreground scribbles to be used for the random walker algorithm used as input seeds and shows the ground truth surface information for reference. Note that this ground truth is not used to computed the scribbles (apart from simulating the extreme points). To increase the number of foreground seeds for the random walker, each path will also be dilated with a 3D ball structure element of radius $r_\mathrm{foreground}=2$. The background seeds are estimated as the dilated and inverted version of the input scribbles. The amount of dilation needed for successful initialization depends on the size of the organ of interest. We typically dilate the scribbles with a ball structure element of radius $r_\mathrm{background}=30$ which achieves good initial seeds for organs such as spleen and liver (see Fig. \ref{fig:results}).

%%%%%%%%%%%%%%%%%%%%%%%%%%%%%%%%%%%%%%%%%%%%%%%%%%%%%%%%%%%%%%%%%%%%%%%%%%%%%%%%%%%%%%%%%%%%%%%%
%\begin{figure*}[htbp!]
%	\centering
%	\begin{tabular}{cc}
%		\subfloat[(a)]{\adjincludegraphics[valign=c,height=2cm]{figs/six_point_scribbles}} &
%		\hfill
%		\subfloat[(b)]{\adjincludegraphics[valign=c,height=2cm]{figs/seven_point_scribbles}}
%	%\end{tabular}
	%\caption{Scribbles derived from extreme point clicks. In (a), we use only the six extreme points to compute shortest path connections between points, while in (b) an extra (7th) point is assumed to be clicked in the rough organ center. The ground truth surface is shown in green for reference. It can be seen that the seventh point constraints the scribbles to run mostly through the organ interior, while six points might result in scribbles running outside the organ boundaries for ``concave'' organs like the spleen.
%	\label{fig:extreme_scribbles}}
%\end{figure*}
%%%%%%%%%%%%%%%%%%%%%%%%%%%%%%%%%%%%%%%%%%%%%%%%%%%%%%%%%%%%%%%%%%%%%%%%%%%%%%%%%%%%%%%%%%%%%%%%
\paragraph{Random walker} Next, the random walker algorithm \citep{grady2006random} is used to produce an initial prediction map $\hat{Y}$ based on the background $s_0$ and foreground $s_1$ scribbles mentioned above. 
The random walker basically solves the diffusion equation between voxels by turning the scribbles $S = {s_0, s_1}$ into a source and sink. The 3D volume here is defined as a $G(E, V)$ graph with $e \in E$ edges and $v \in V$ vertices. Each edge between two vertices of $v_i$ and $v_j$ is referred to as $e_{ij}$ and a weight of $w_{ij}$ can be assigned based on gradients of the image intensities. In addition, $d_i = \sum{w_{ij}}$ defines the degree of a given vertex.
In order to get a probability $p(\omega|x)$ of whether each vertex $v_i$ belongs to the foreground $\omega_1$, we solve the diffusion equation. $L$ is the Laplacian matrix of the weighted image graph $G$ with each element of the matrix defined as:
\begin{equation}
    L_{ij} = 
    \begin{cases}
        d_i,& \text{if } i = j, \\
        -w_{ij},& \text{if } i \text{ and } j \text{ are adjacent voxels}, \\
        0,& \text{otherwise}
    \end{cases}
\end{equation}
The weights between adjacent voxels can be defined as $w_{ij}=e^{-\beta|z_j-z_i|^2}$. This will make the diffusion between similar voxel intensities $z_i$ and $z_j$ easier and hence allow them to be assigned to the same class. Here, $\beta$ is a tunable hyperparameter that controls the amount of diffusion. We keep $\beta=130$ in all our experiments.
By separating the voxels marked by scribbles $S$ and unmarked voxels, the Laplacian matrix $L$ can be decomposed into blocks.
\begin{equation}
L = \left[ {\begin{array}{cc}
    L_M & B \\
    B^T & L_U \\
    \end{array} } \right]    
\end{equation}
Here, $M$ corresponds to voxels marked by scribbles $S$ and $U$ to unmarked voxels.

This can be formulated as a system of equations which can be analytically solved as:
\begin{equation}
L_U X = - B^T M, 
\label{equ:random_walker}
\end{equation}
where $M$ is made of elements $m_j^\omega$ which are 1 for marked voxels of $s_\omega$ for the given class $\omega$, and 0 otherwise. Solving Equ. \ref{equ:random_walker}, results in a probability for each voxel $p(\omega|x) = x^\omega_i$, resulting in our pseudo label $\hat{Y}$
%%%%%%%%%%%%%%%%%%%%%%%%%%%%%%%%%%%%%%%%%%%%%%%%%%%%%%%%%%%%%%%%%%%%%%%%%%%%%%%%%%%%%%%%%%%%%%%%
%% OLD FIGURE!!!!!!
%%%%%%%%%%%%%%%%%%%
%\begin{figure*}[htbp!]
%	\centering
%	\begin{tabular}{cc}
%		\subfloat[(a)]{\adjincludegraphics[valign=c,height=2cm]{figs/six_point_scribbles}} &
%		\hfill
%		\subfloat[(b)]{\adjincludegraphics[valign=c,height=2cm]{figs/seven_point_scribbles}}
%	\end{tabular}
%	\caption{Random walker result. (a) image, (b) prediction.
%	\label{fig:extreme_scribbles}}
%\end{figure*}
%%%%%%%%%%%%%%%%%%%%%%%%%%%%%%%%%%%%%%%%%%%%%%%%%%%%%%%%%%%%%%%%%%%%%%%%%%%%%%%%%%%%%%%%%%%%%%%%
\subsection{Step 3: Segmentation via deep fully convolutional network}
\noindent Next, we can train a fully convolutional neural network to segment the given foreground class with $P(X) = f(X)$ with pairs of $X$ and pseudo labels $\hat{Y}$. 
Our preferred network architecture follows the encoder-decoder network proposed in \citet{myronenko20183d} (without the VAE part), using 3D convolutions throughout the network.
\paragraph{Encoder}
The encoder uses residual blocks \citep{he2016deep}, where each block consists of two convolutions with normalization and ReLU, followed by additive skip connection. Here, we use~\textit{group normalization} (GN) \citep{wu2018group}, which typically shows better performance than batch normalization \citep{ioffe2015batch} when batch size is small (in our case batch size 4). We adopt a standard FCN approach to slowly decrease the number of image dimensions by 2 and simultaneously increase the number of features by 2 as in \citet{ronneberger2015u}. For downsizing, we use strided convolutions with a stride of 2. All conversions are $3 \times 3 \times 3$ with an initial filter number equal to 8 in the input layer of the network.
\paragraph{Decoder}
The design of the decoder is identical to the one of the encoder, but with a single residual block per each spatial level of the network. Each level of decoders starts with up-sampling that involves reducing the number of features by a factor of 2 (using $1\times 1\times 1 $convolutions) and doubling the spatial dimension using trilinear up-sampling. This is followed by adding or concatenating the features from the equivalent spatial level encoder. In this study we use addition due to the lower memory consumption of the resulting network. At the end of the decoder, the features have the same spatial size as the original image and the number of features equal to the size of the initial input function. This is followed by $1 \times 1 \times 1 $conversion into one output channel followed by a final \textit{sigmoid} activation as we are assuming the binary segmentation case in this work.

\paragraph{Attention}
\label{sec:point_attention}
We follow the approach of \citet{oktay2018attention} to implement attention gates in the decoder part of our segmentation network. The attention gates help the model to focus on the structures of interest. Attention gates can encourage the model to suppress regions that are irrelevant to the segmentation task and highlight the regions of interest most relevant to the segmentation task (see figure \ref{fig:network}).

\begin{figure}[htbp]
  \newcommand{\figheight}{1.25cm}
	\centering
	\begin{tabular}{ccc}
	    %%%%%%%%%%%%%%%%%%%%%%%%%%%%%%%%%%%%%%%%%%%%%%%%%%%%%%%%%%%%%%%%%%%%%%%%%%
		\subfloat[(a)]{\adjincludegraphics[valign=c,height=\figheight]{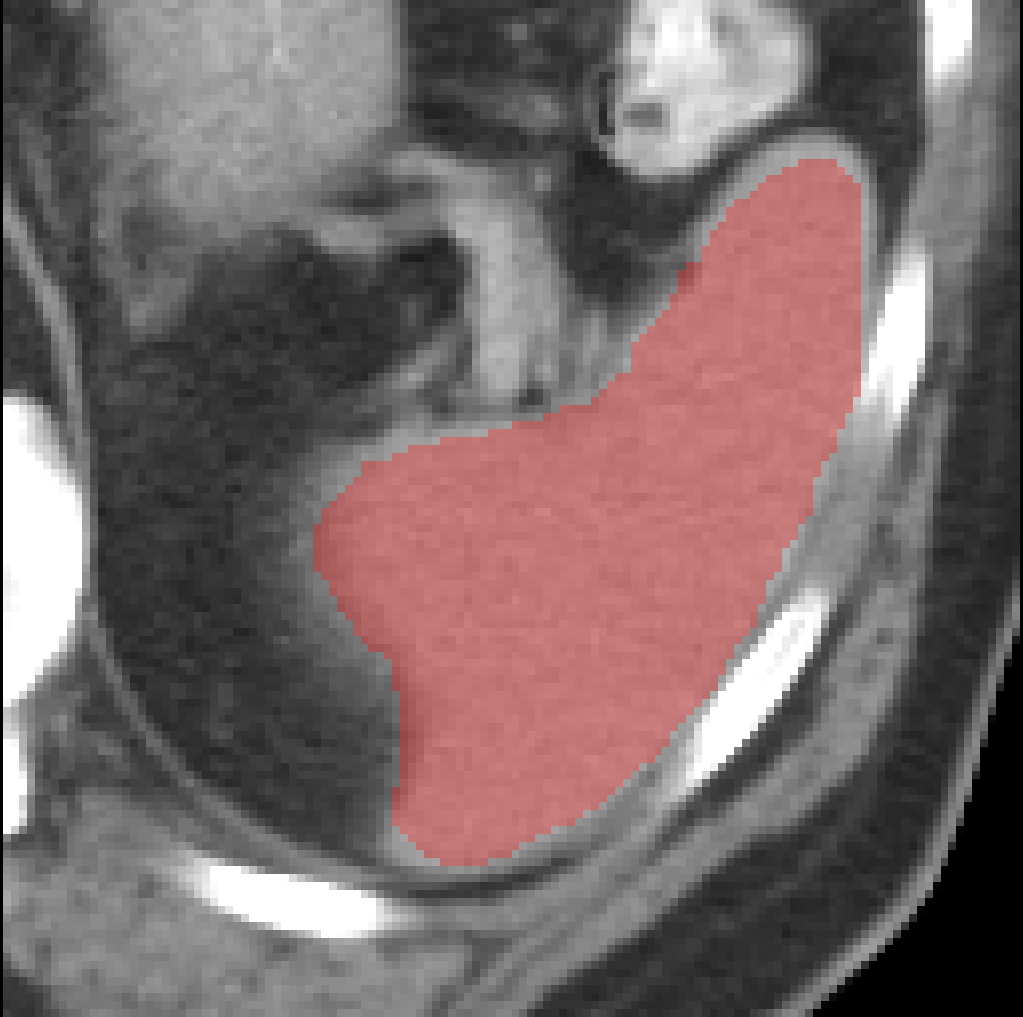}} &
	    \hfill
		\subfloat[(b)]{\adjincludegraphics[valign=c,height=\figheight]{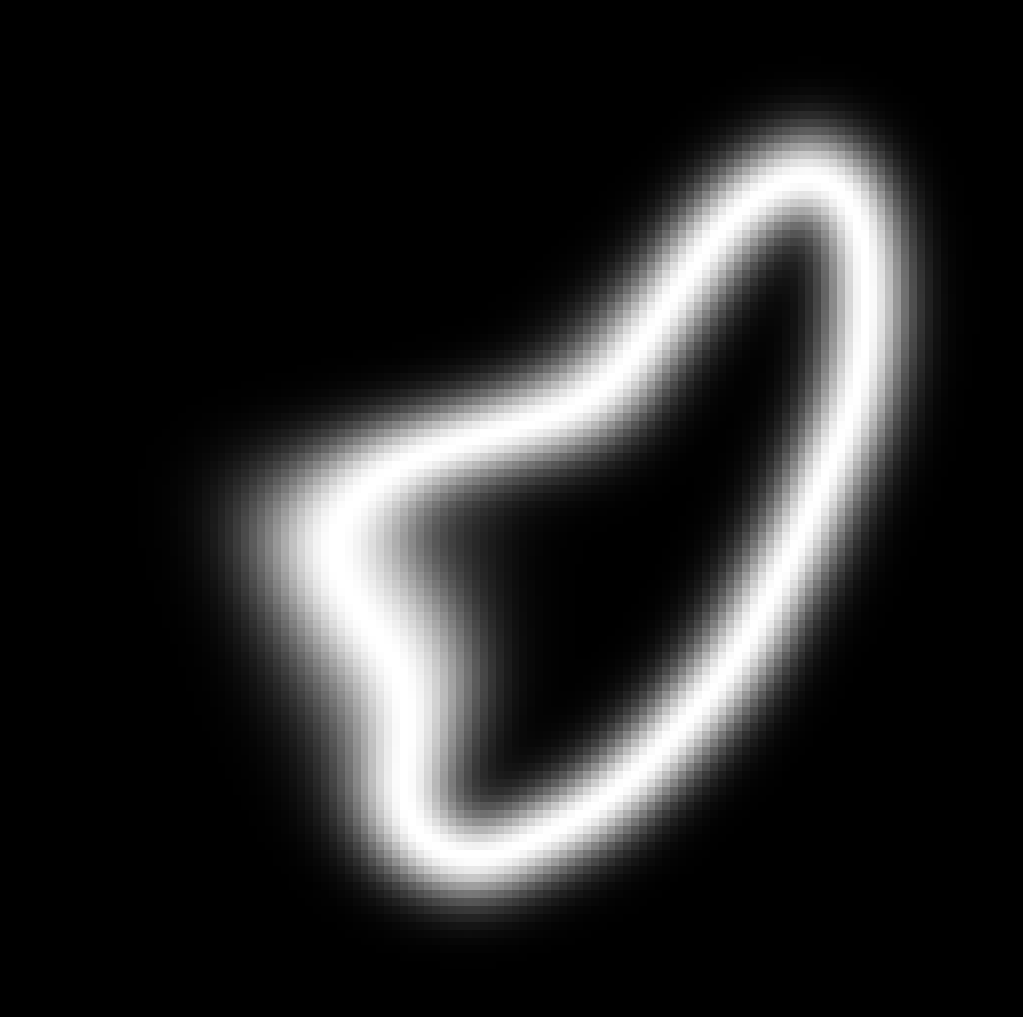}} &
		\hfill
		\subfloat[(c)]{\adjincludegraphics[valign=c,height=\figheight]{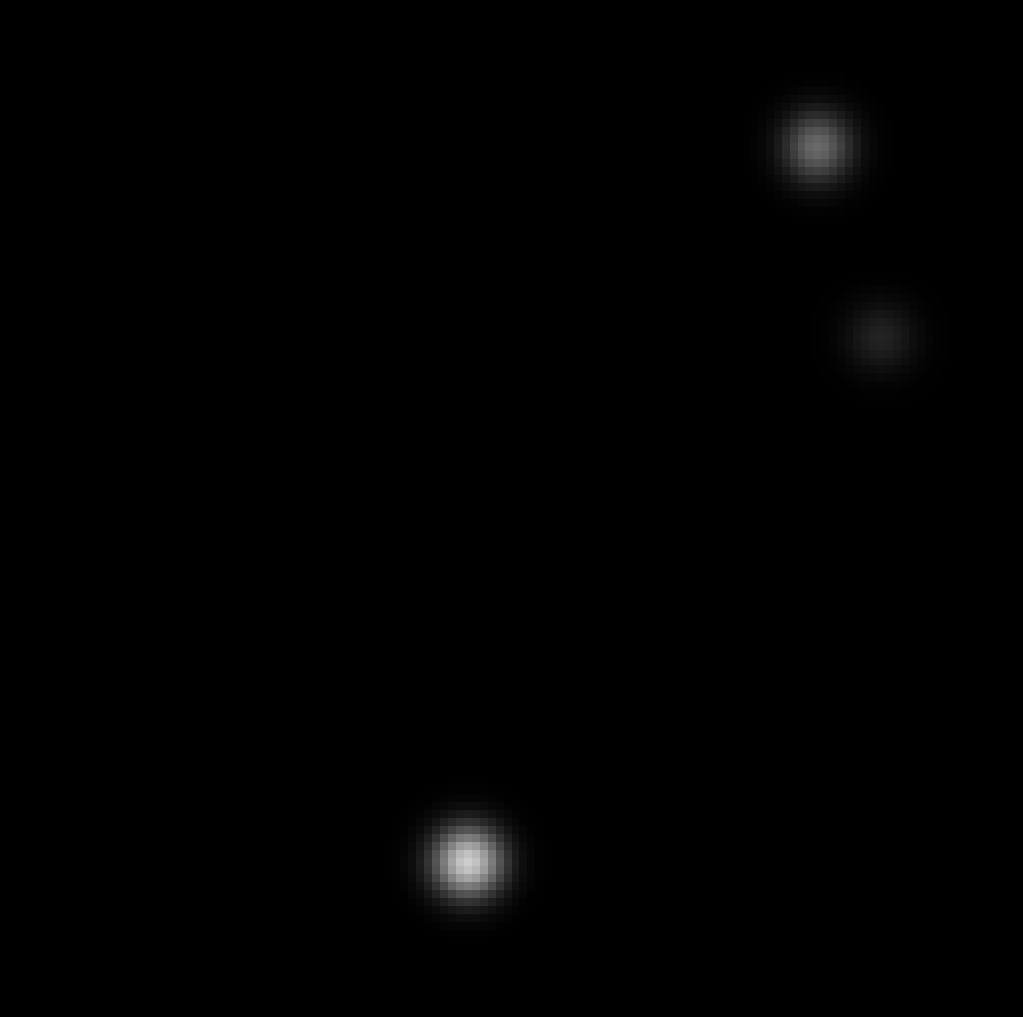}} 
		%%%%%%%%%%%%%%%%%%%%%%%%%%%%%%%%%%%%%%%%%%%%%%%%%%%%%%%%%%%%%%%%%%%%%%%%%%		
	\end{tabular}
	\caption{Visualization of the boundary enhancement map computed in Equ. 6-8 in the paper. In this example, we show (a) the ground truth overlaid on the image, (b) the boundary enhancement map $b(P)$, and (c) the point channel $G(\{e\})$ on the corresponding axial slice of the 3D volume used in the computation of the point loss $L_\mathrm{points}$ (see Equ. 9). The loss is minimized if the prediction's boundary $b(P)$ aligns with the center of each clicked extreme point $e$ in $G(\{e\})$. Note, that during training, we compute the boundary on the model's prediction $P$ but here we show it computed on the ground truth for illustration purpose. 
	\label{fig:point_loss}}
\end{figure}

The attention gate can be further augmented by the point channel information available from extreme point selection. We propose to add the extreme point channels $G($\{e\}$)$ at each level of the decoder to further guide the network to learn the relevant information. In practice, we downsample the initial input point channel to match the resolution of each decoder level and concatenate it with the gating features from the encoder path of the network in each attention gate.
%
%TODO: add a figure here to describe attention
%
\paragraph{Dice loss}
\label{sec:dice_loss}
The Dice loss \citep{milletari2016v} is a popular objective function for segmentation tasks in medical imaging. Its properties allow it to automatically scale to unbalanced labeling problems. At the same time, it also naturally adapts without any changes to the original formulation to learn from probability maps:
\begin{equation}
    \mathcal{L}_{Dice} = 1 - \frac{2\sum_{i=1}^{N}y_i\hat{y}_i}{\sum_{i=1}^{N} y_i^2 + \sum_{i=1}^{N} \hat{y}_i^2}
    \label{eq:dice_loss}
\end{equation}
Here, $y_i$ is the predicted probability from our network $f(X)$ and $\hat{y_i}$ is the weak label probability from our pseudo label map $\hat{Y}$ at voxel $i$.
\paragraph{Point loss}
\label{sec:point_loss}
The extreme points selected by the user for weak annotation cannot only be used for generating initial scribbles but also in an additional loss function during the training of our deep neural network.
We add an additional constraint to the deep learning training making use of the extreme points the user has already selected. This new loss $L_\mathrm{points}$ penalizes the distance between the boundary of our model’s predicted segmentation mask $P = f(X)$ and the location of the extreme points.
%First, we use $B$ convolution operations to enhance the boundary $b(P)$ of the prediction:
%\begin{align}
%    b_0(P) &= conv_B\left(...\left(conv_3\left(conv_2\left(conv_1\left(P\right)\right)\right)\right)...\right) \\
%    b_1(P) &= \left(b_0(P) - 0.5\right)^2 \\
%    b(P)   &= e^{-b_1(P)}
%    \label{eq:boundary}
%\end{align}
%Here, the convolutional kernel in each $conv$ operation is set to be constant $n \times n \times n$ kernel with each element being $1/{n^3}$. $B$ should be adjusted depending on the size of input image and the extent of organ of interest. In our setting, we use $B=25$ to achieve a good boundary enhancement at the scale of the images and targeted organs.
% \begin{equation}
%    L_\mathrm{points} = -\frac{1}{N}\sum_{N}\mathcal{G}\left(P(X)\right) \cdot G(\{e\}),
%    \label{eq:point_loss}
% \end{equation}
To compute it, we apply a Gaussian filter $\mathcal{G(\cdot)}$ to our models prediction $P(X)$ which can be easily implemented using standard 3D convolutional operations with a constant $n \times n \times n$ kernel with each element being $1/{n^3}$.
The resulting point distance loss between the filtered prediction $\mathcal{G}\left(P(X)\right)$ and the extreme points channel $G(\{e\})$ (which includes a Gaussian kernel placed over each extreme point) is therefore
\begin{equation}
    L_\mathrm{points} = -\frac{1}{N}\sum_{i=1}^{N}g_i \mathrm{g}_i,
    \label{eq:point_loss}
\end{equation}
where ${N}$ are the number of voxels $i$ in the image and $g_i \in \mathcal{G}\left(P(X)\right)$ and $\mathrm{g}_i \in G(\{e\})$, respectively. The point loss computation is illustrated in Fig. \ref{fig:point_loss}.
This results in a new total loss used for training:
\begin{equation}
    L = L_\mathrm{Dice}  + \alpha L_\mathrm{points}.
    \label{eq:joined_loss}
\end{equation}
Here, $\alpha$ is hyperparameter weight that controls the influence of the point distance loss. 
%
%\vspace{-5cm}
%%%%%%%%%%%%%%%%%%%%%%%%%%%%%%%%%%%%%%%%%%%%%%%%%%%%%%%%%%%%%%%%%%%%%%%%%%%%%%%%%%%%%%%%%%%%%%%%
%%%%%%%%%%%%%%%%%%%%%%%%%%%%%%%%%%%%%%%%%%%%%%%%%%%%%%%%%%%%%%%%%
\paragraph{Point loss implementation}
\noindent We implement the Gaussian filter $\mathcal{G}(\cdot)$ using a set of standard 3D convolutions.

First, we use $B$ convolution operations to enhance the boundary of the prediction $P(X)$:
\begin{align}
    \mathcal{G}_0(P(X)) &= conv_B\left(...\left(conv_3\left(conv_2\left(conv_1\left(P(X)\right)\right)\right)\right)...\right) \nonumber\\
    \mathcal{G}_1(P(X)) &= \left(\mathcal{G}_0(P(X)) - 0.5\right)^2  \nonumber\\
    \mathcal{G}(P(X))   &= e^{-\mathcal{G}_1(P(X))} 
    \label{eq:boundary}
\end{align}
Here, the convolutional kernel in each $conv$ operation is set to be constant $n \times n \times n$ kernel with each element being $1/{n^3}$. $B$ should be adjusted depending on the size of input image and the extent of organ of interest. In our setting, we use $B=25$ to achieve a good boundary enhancement at the scale of the images and targeted organs.

The resulting point distance loss between the filtered prediction $\mathcal{G}\left(P(X)\right)$ and the extreme points channel $G(\{e\})$ (which includes a Gaussian kernel placed over each extreme point) is therefore as in Equ. \ref{eq:point_loss}.

%%%%%%%%%%%%%%%%%%%%%%%%%%%%%%%%%%%%%%%%%%%%%%%%%%%%%%%%%%%%%%%%%
\subsection{Step 4: Regularization using random walker algorithm}
\noindent We could stop learning after the above segmentation network $f(X)$ is trained on the $\hat{Y}$ pseudo labels for the first time. Nevertheless, we note that an additional regularization step by an additional random walker segmentation as mentioned above may be of great benefit to the convergence of our weakly-supervised segmentation approach. This approach is close in spirit to \citet{rajchl2017deepcut}, where a DenseCRF is used as the post-processing step during iterative refinement.

To increase the amount of regularization that the random walker can give to the predictions $P(X)$ of the network, we define an area of uncertainty $\mathcal{U}(P(X))$. The foreground and background in the prediction map can be defined as $P(X)>=0.5$ and $P(X)<0.5$, respectively. Here, we chose a ball structure element of radius $r_\mathrm{randomwalker}=4$ to erode both the foreground and background regions in all our segmentation tasks to compute $\mathcal{U}$ which in turn is acting as the unmarked voxels in the random walker algorithm. This allows the random walker to generate new predictions around the foreground object's boundary that differ from previous 3D network's predictions and, in turn, help the next deep learning training iteration to learn new features from the same set of training images and to not get stuck in a poor local minimum. Besides, we find that our weakly supervised segmentation framework becomes unstable without this step and does not converge as easily to a satisfactory result (see Figure \ref{fig:convergence}).

\section{Experiments \& Results}
\label{sec:experiments}
%%%%%%%%%%%%%%%%%%%%%%%%%%%%%%%%%%%%%%%%%%%%%%%%%%%%%%%%%%%%%%%%%%%%%%%%%%%%%%%%%%%%%%%%%%%%%%%%
\begin{figure*}[htbp]
  \newcommand{\figheight}{1.2cm}
	\centering
	\begin{tabular}{cccccc}
	    %%%%%%%%%%%%%%%%%%%%%%%%%%%%%%%%%%%%%%%%%%%%%%%%%%%%%%%%%%%%%%%%%%%%%%%%%%
		\scriptsize\msdspleen\hfill & \subfloat{\adjincludegraphics[valign=c,height=\figheight]{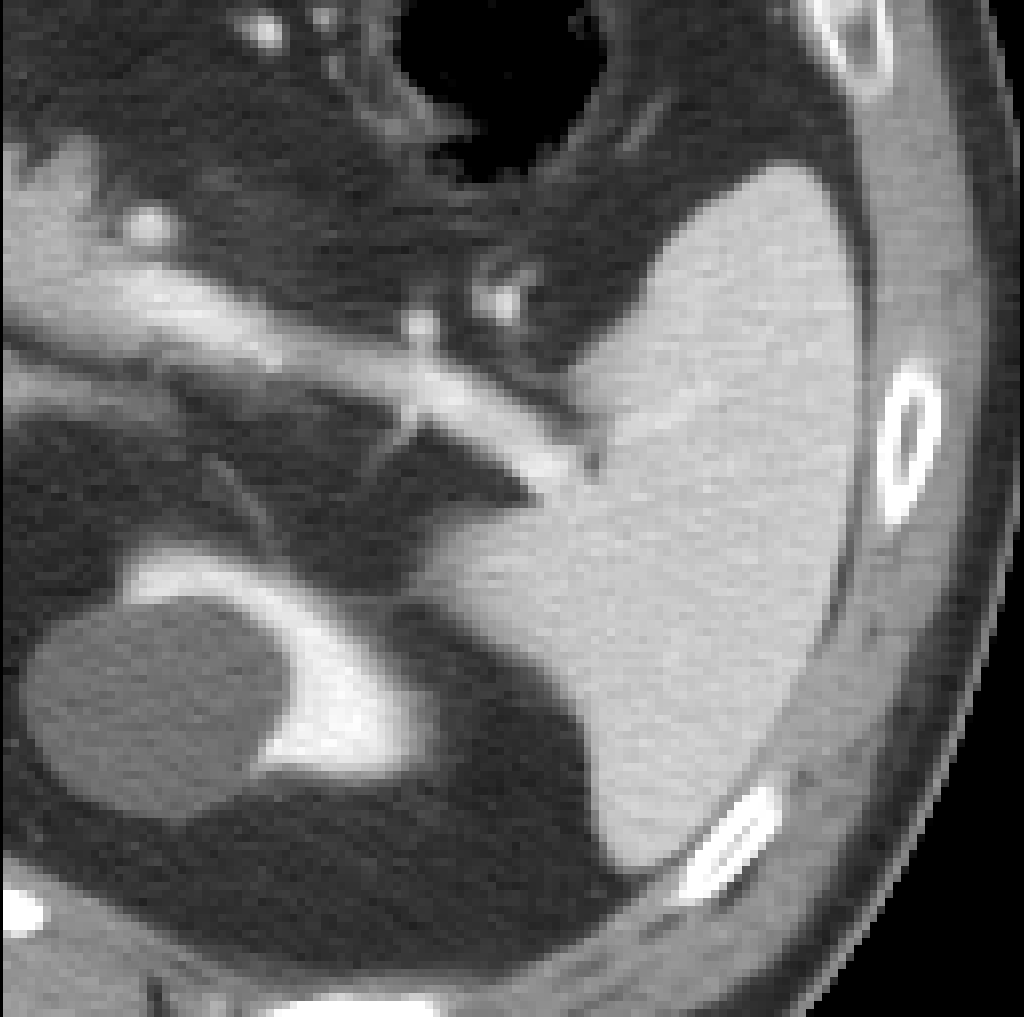}} &
		\hfill
		\subfloat{\adjincludegraphics[valign=c,height=\figheight]{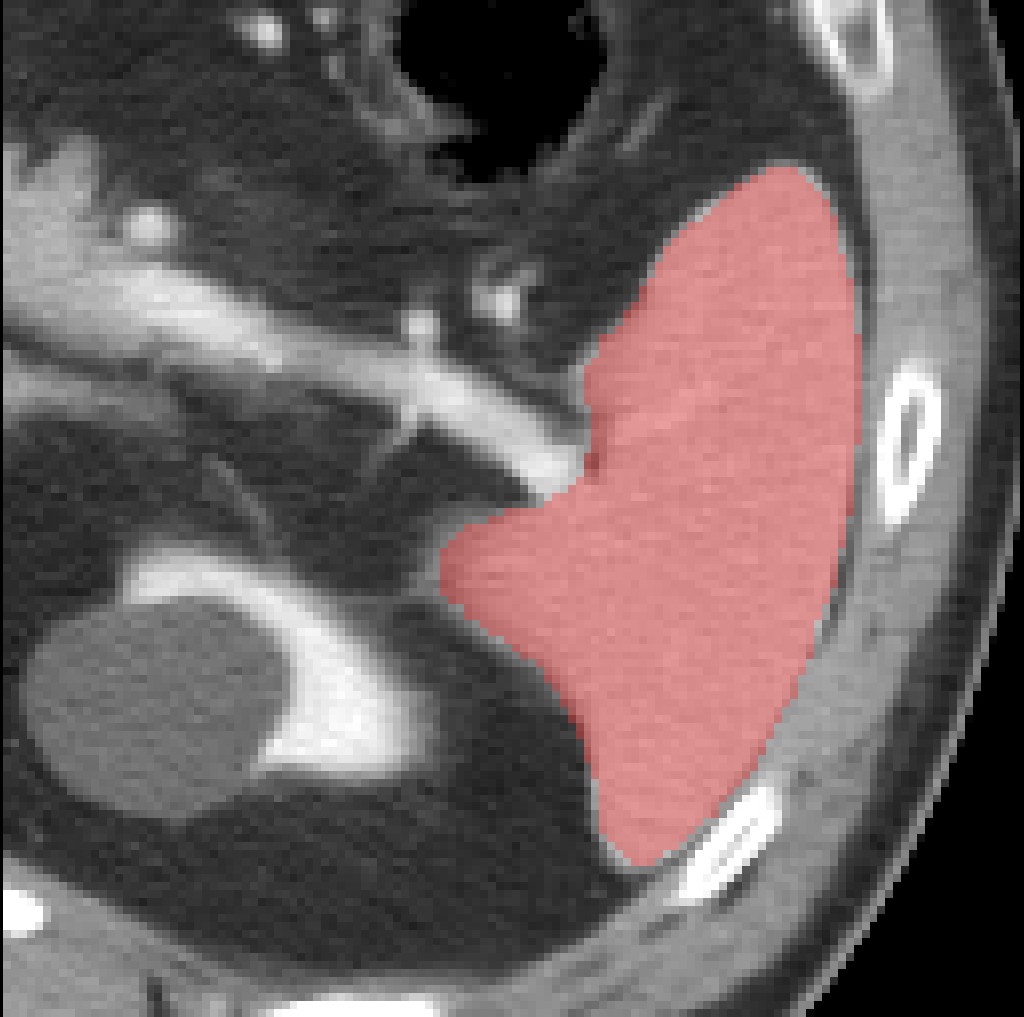}} & 
		\hfill
		\subfloat{\adjincludegraphics[valign=c,height=\figheight]{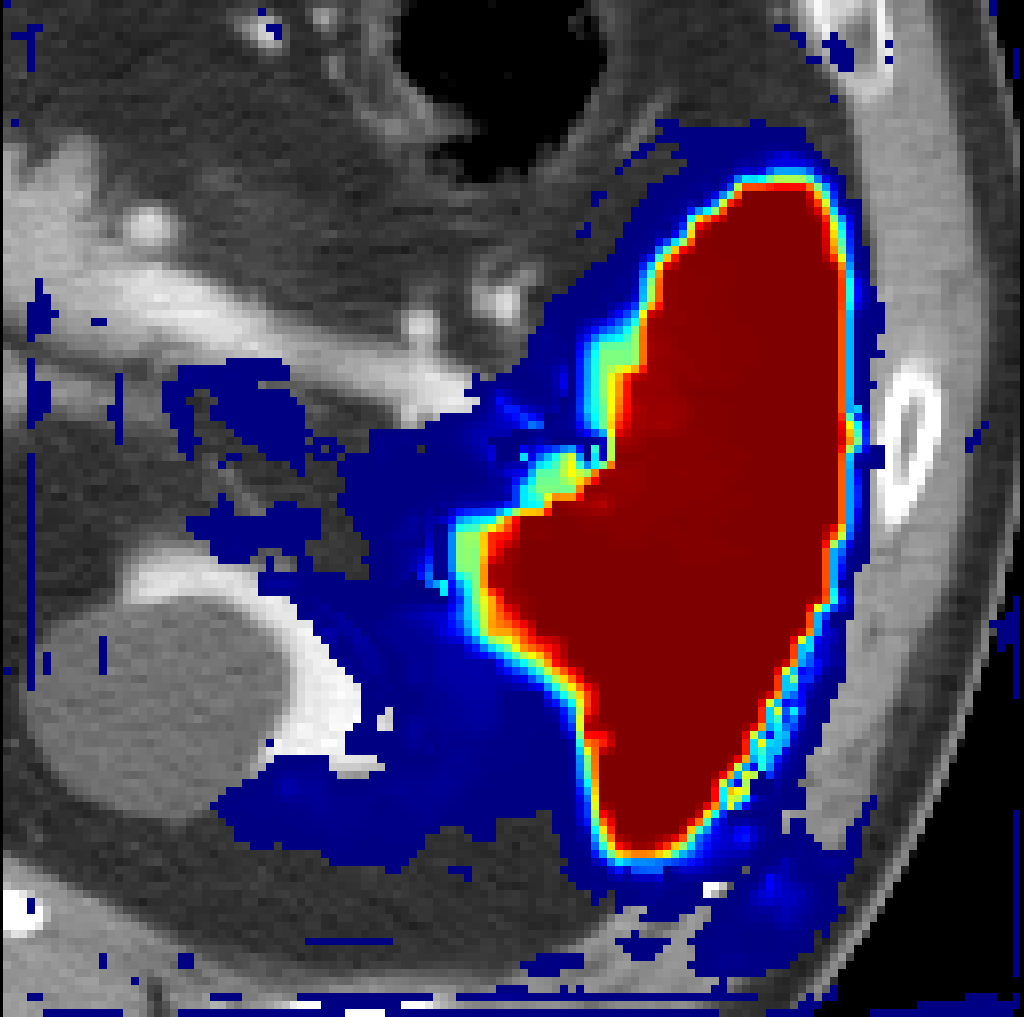}} &
		\hfill
		\subfloat{\adjincludegraphics[valign=c,height=\figheight]{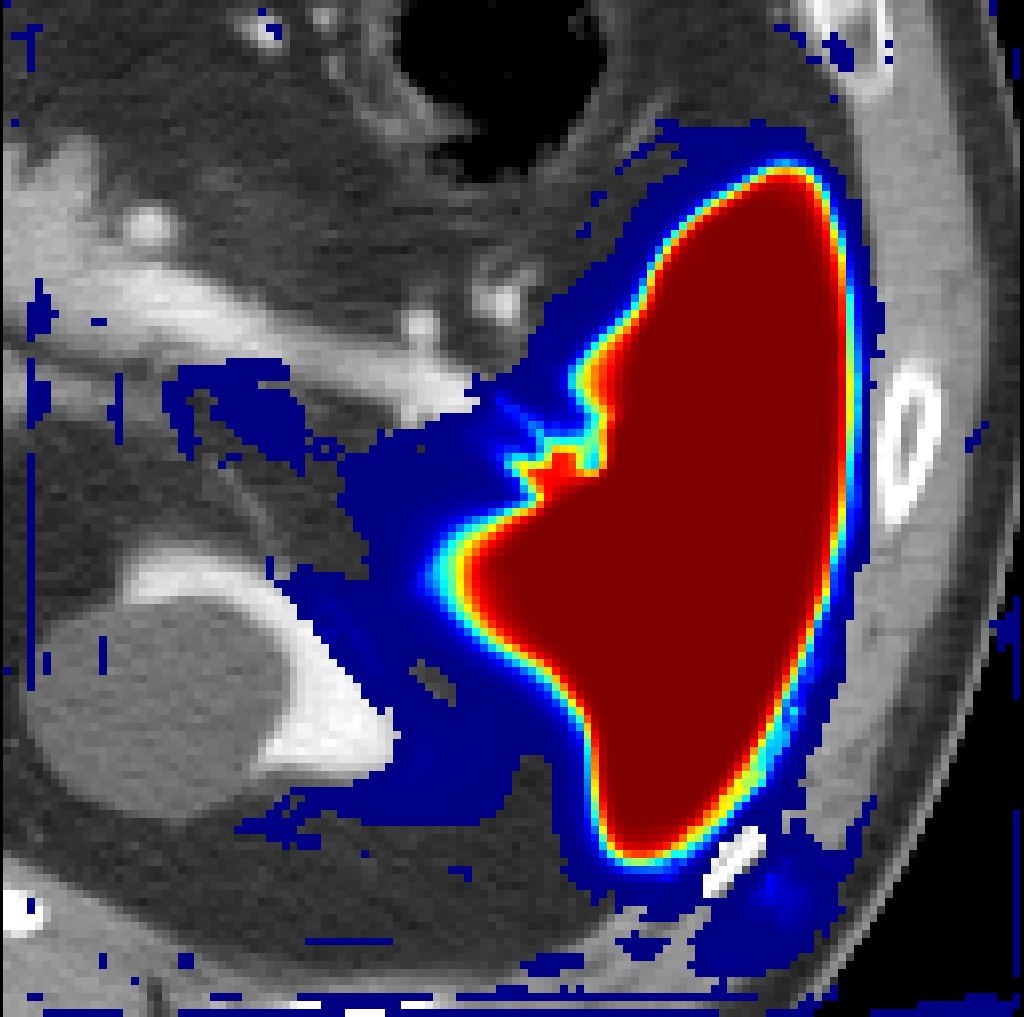}} &
		\hfill
		\subfloat{\adjincludegraphics[valign=c,height=\figheight]{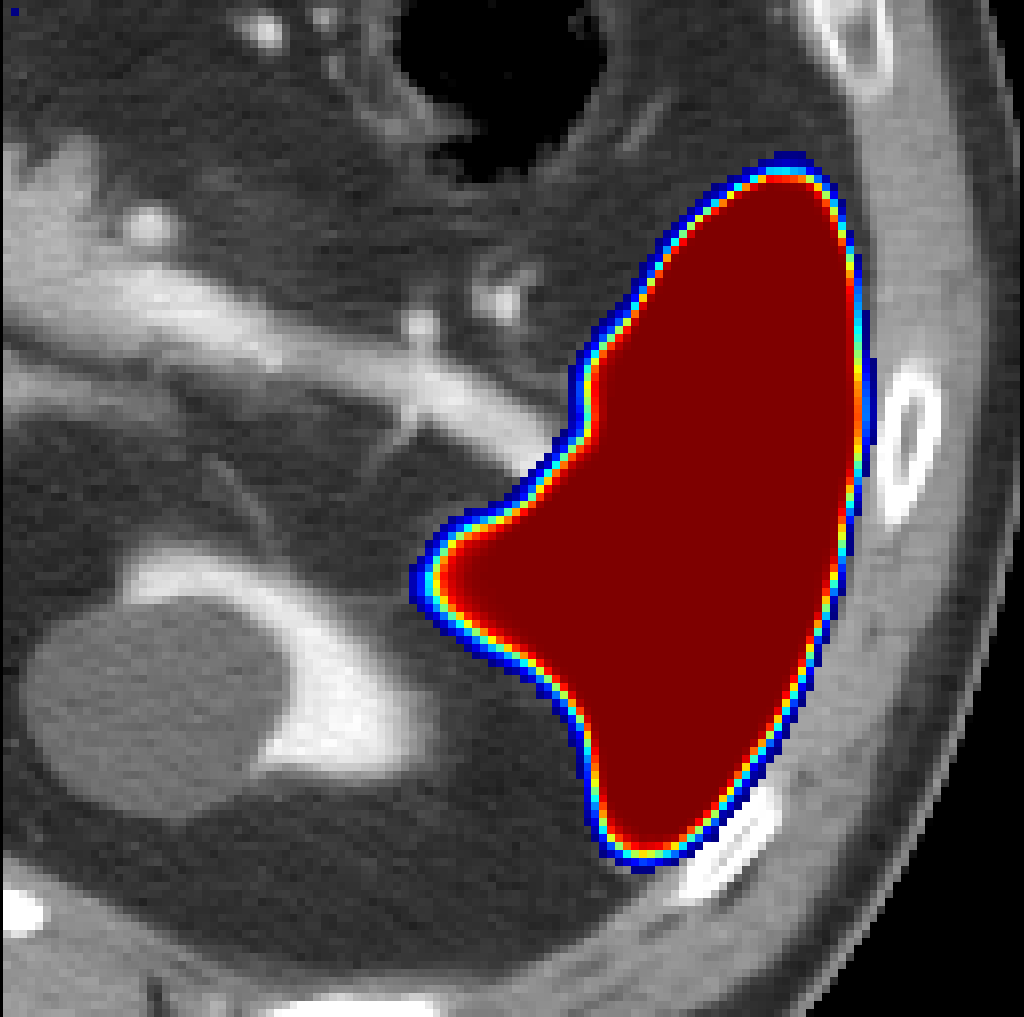}} \\ %UPDATE WITH FULL. SUP. DICE
		%%%%%%%%%%%%%%%%%%%%%%%%%
	    %%%%%%%%%%%%%%%%%%%%%%%%%
	    \scriptsize\mospleen \hfill &
		\subfloat{\adjincludegraphics[valign=c,height=\figheight]{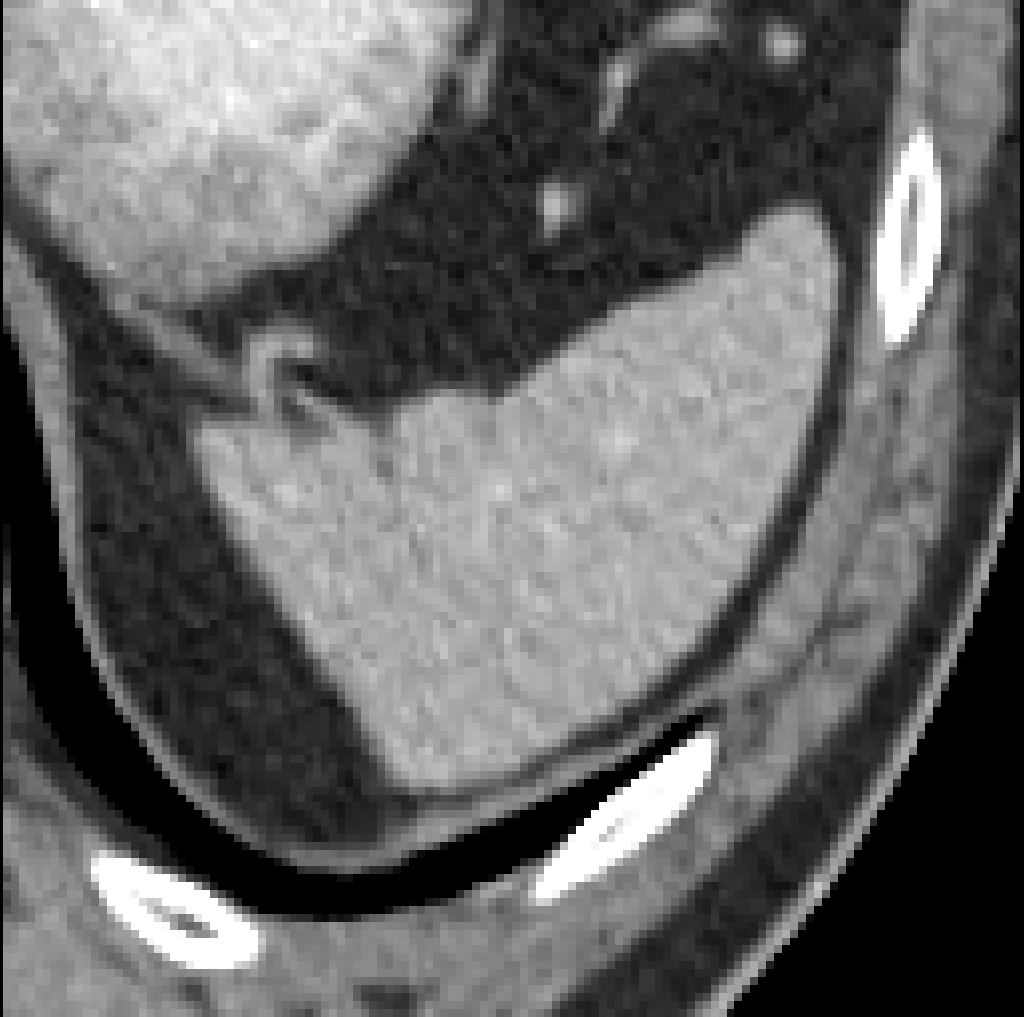}} &
		\hfill
		\subfloat{\adjincludegraphics[valign=c,height=\figheight]{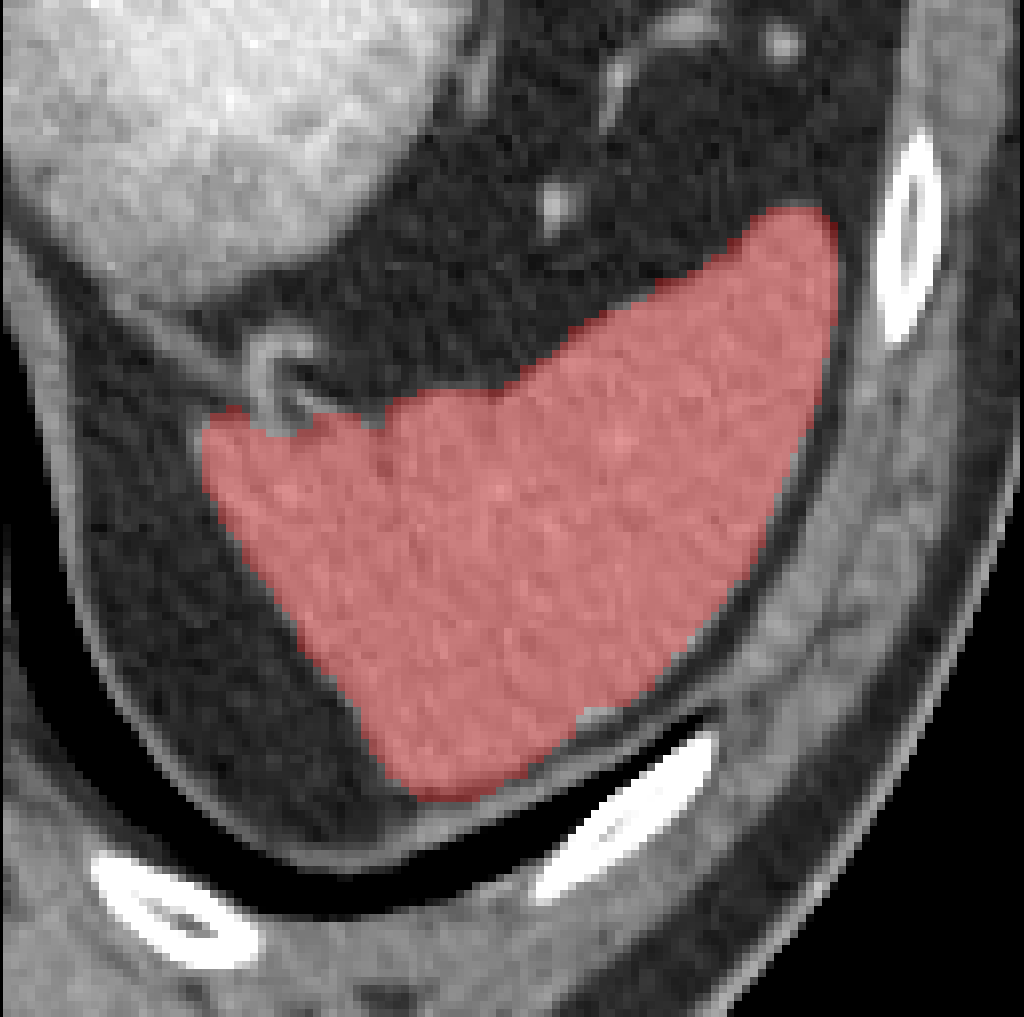}} & 
		\hfill
		\subfloat{\adjincludegraphics[valign=c,height=\figheight]{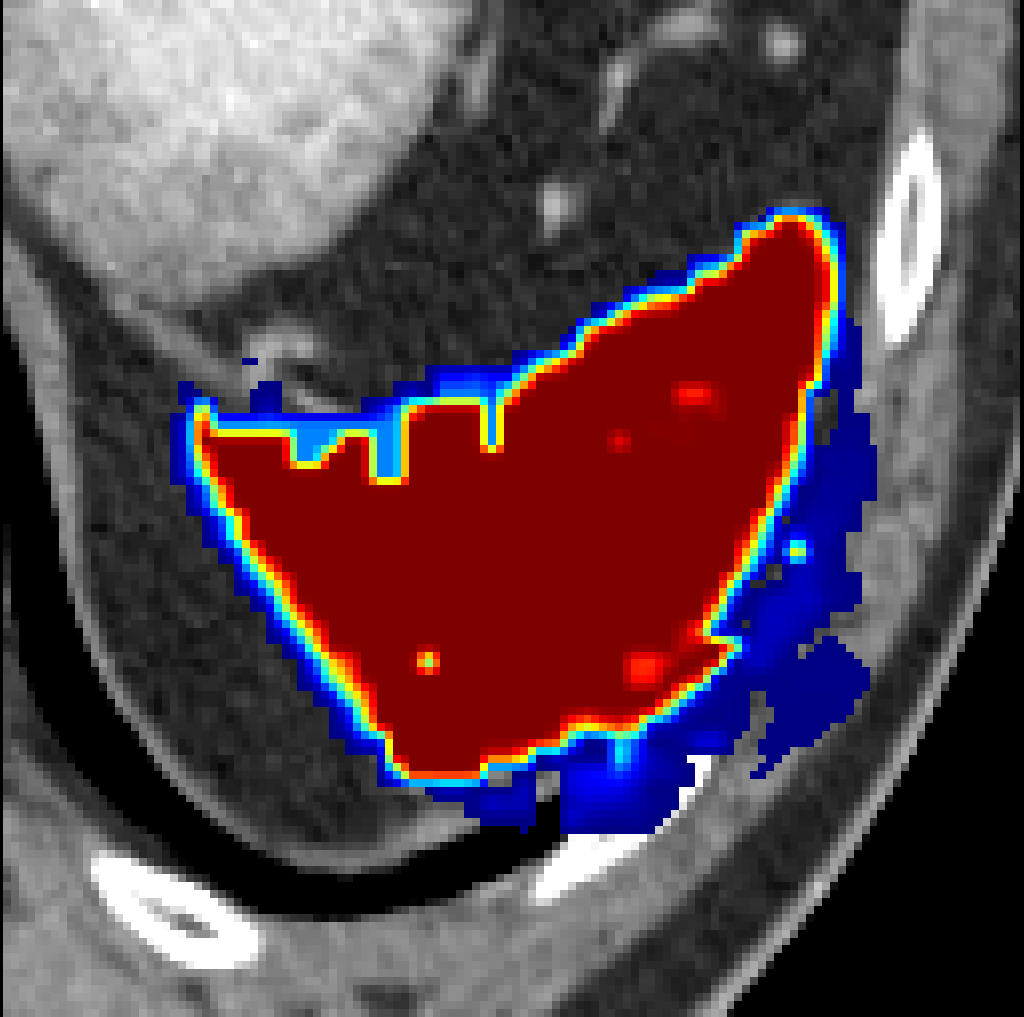}} &
		\hfill
		\subfloat{\adjincludegraphics[valign=c,height=\figheight]{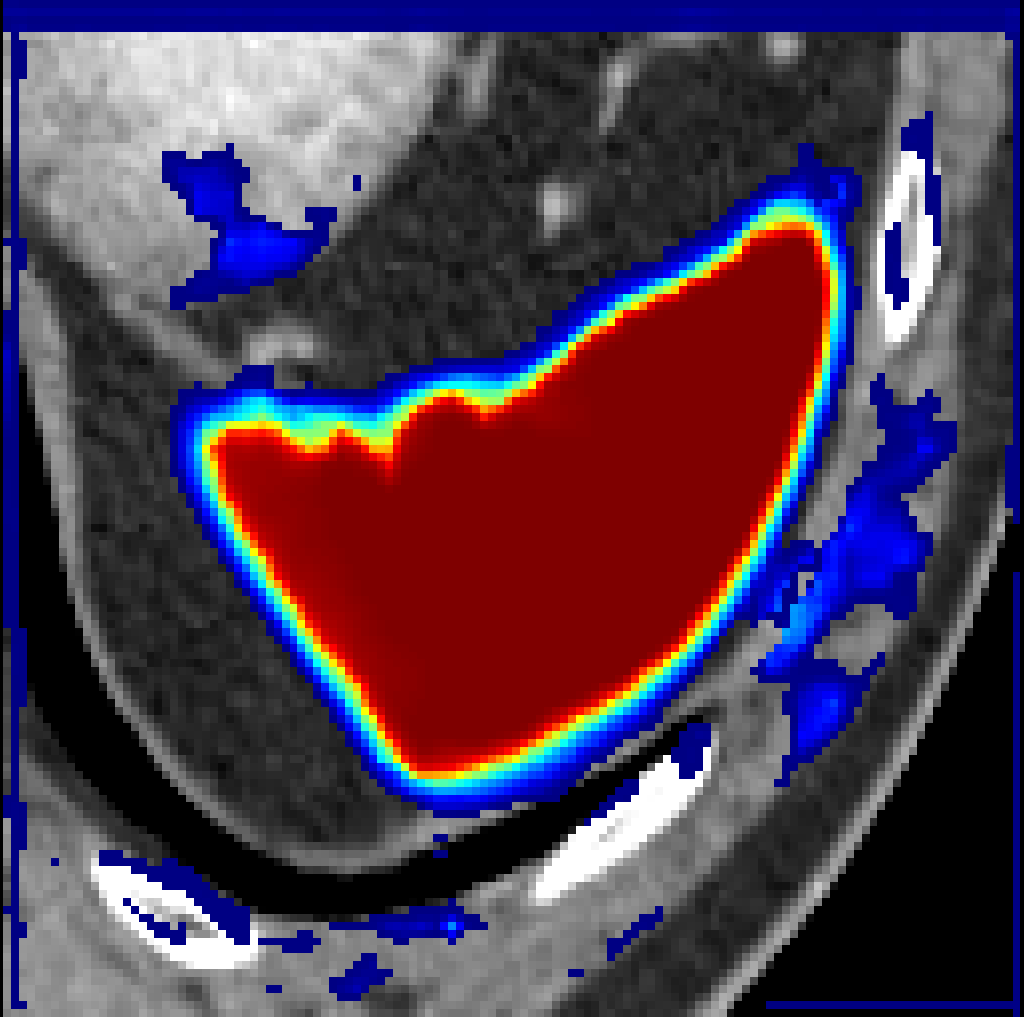}} &
		\hfill
		\subfloat{\adjincludegraphics[valign=c,height=\figheight]{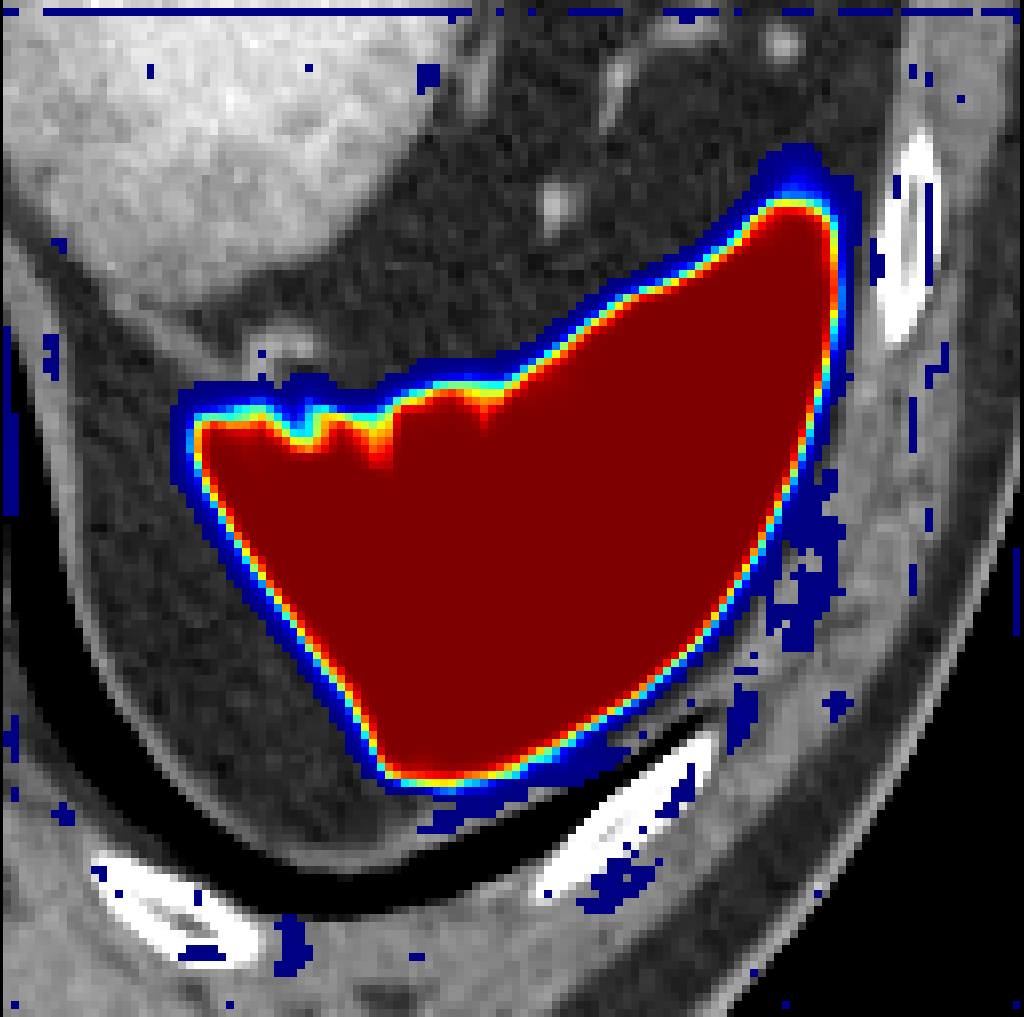}} \\ %UPDATE WITH FULL. SUP. DICE
		%%%%%%%%%%%%%%%%%%%%%%%%%
	    %%%%%%%%%%%%%%%%%%%%%%%%%
	    \scriptsize\moliver \hfill &
		\subfloat{\adjincludegraphics[valign=c,height=\figheight]{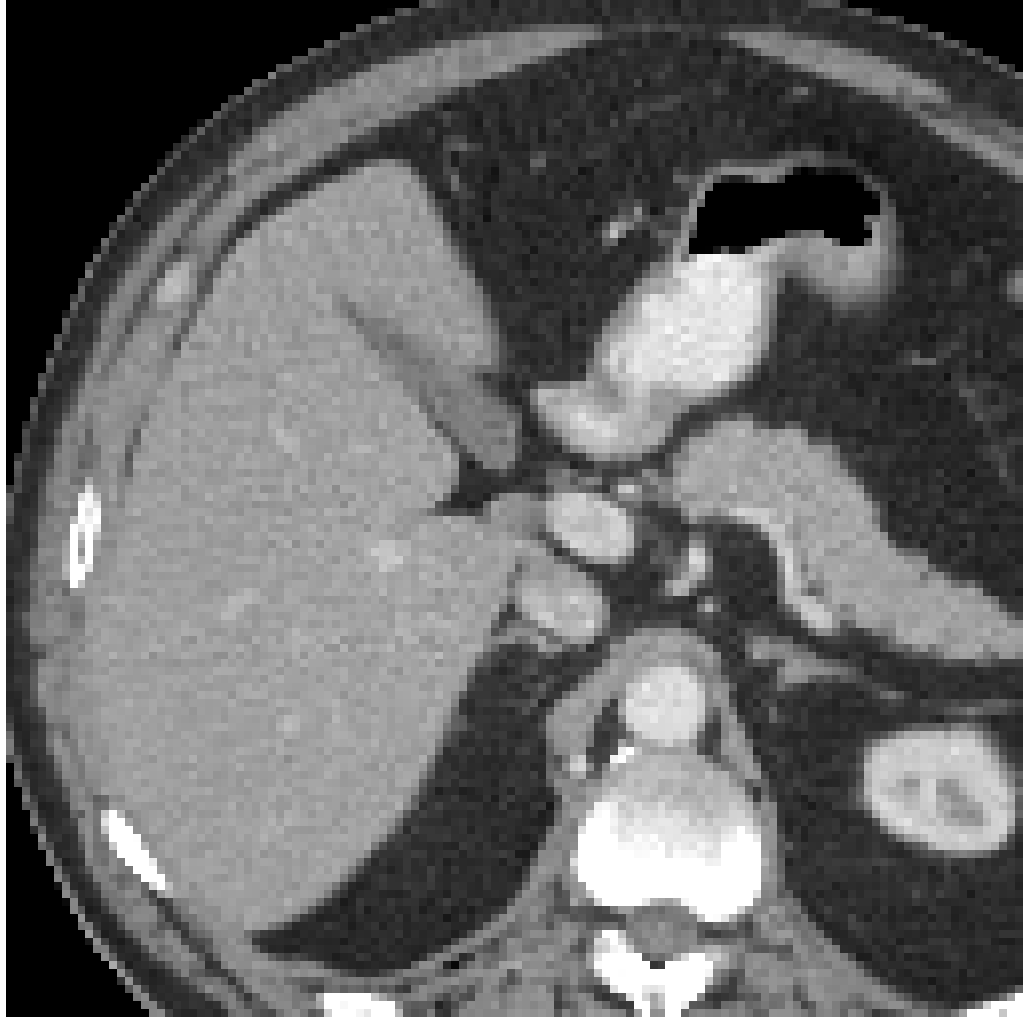}} &
		\hfill
		\subfloat{\adjincludegraphics[valign=c,height=\figheight]{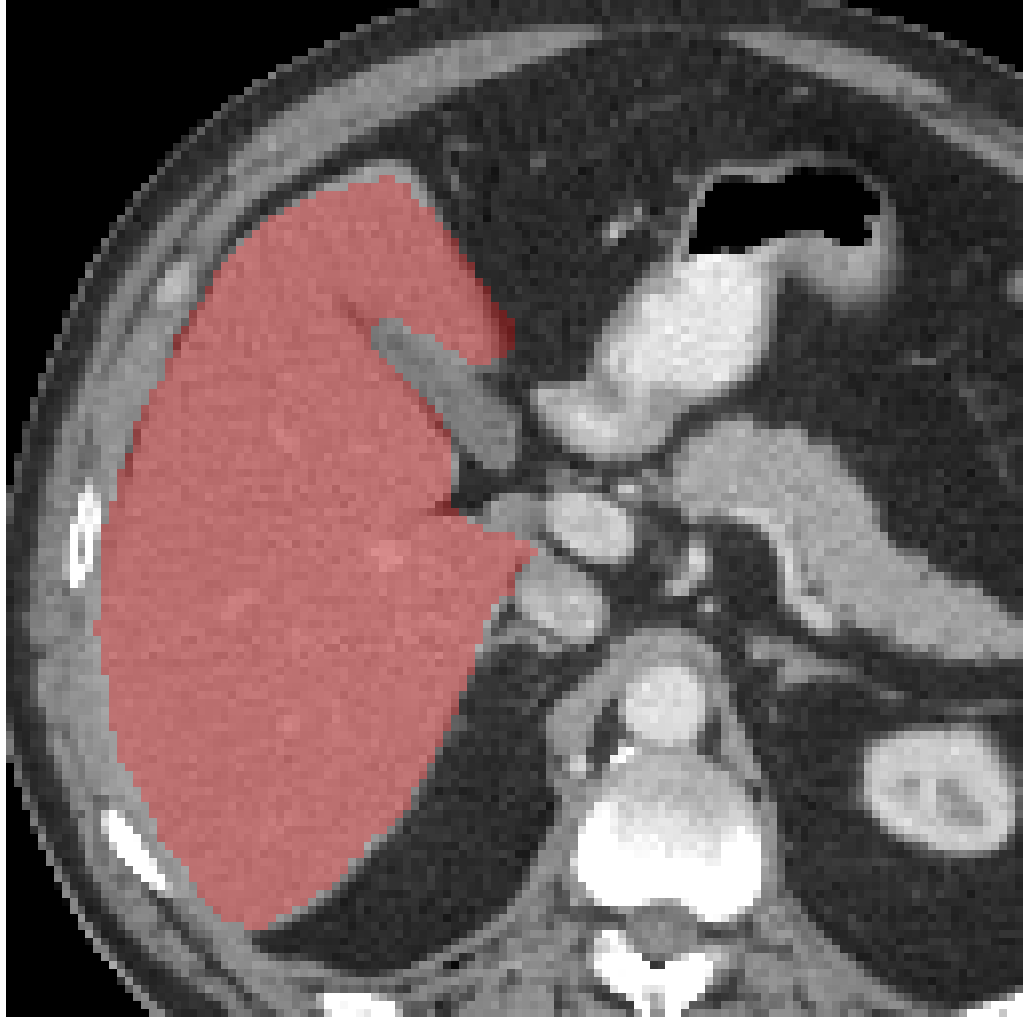}} & 
		\hfill
		\subfloat{\adjincludegraphics[valign=c,height=\figheight]{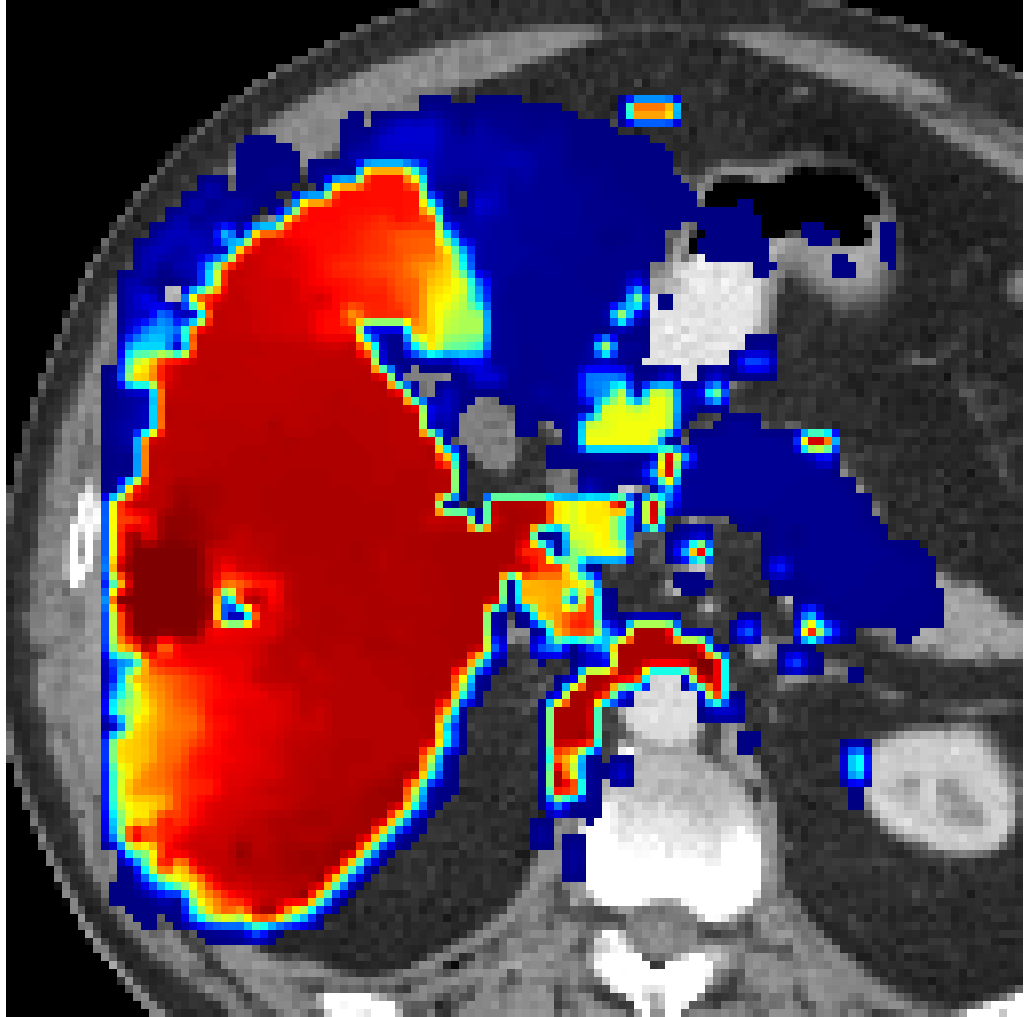}} &
		\hfill
		\subfloat{\adjincludegraphics[valign=c,height=\figheight]{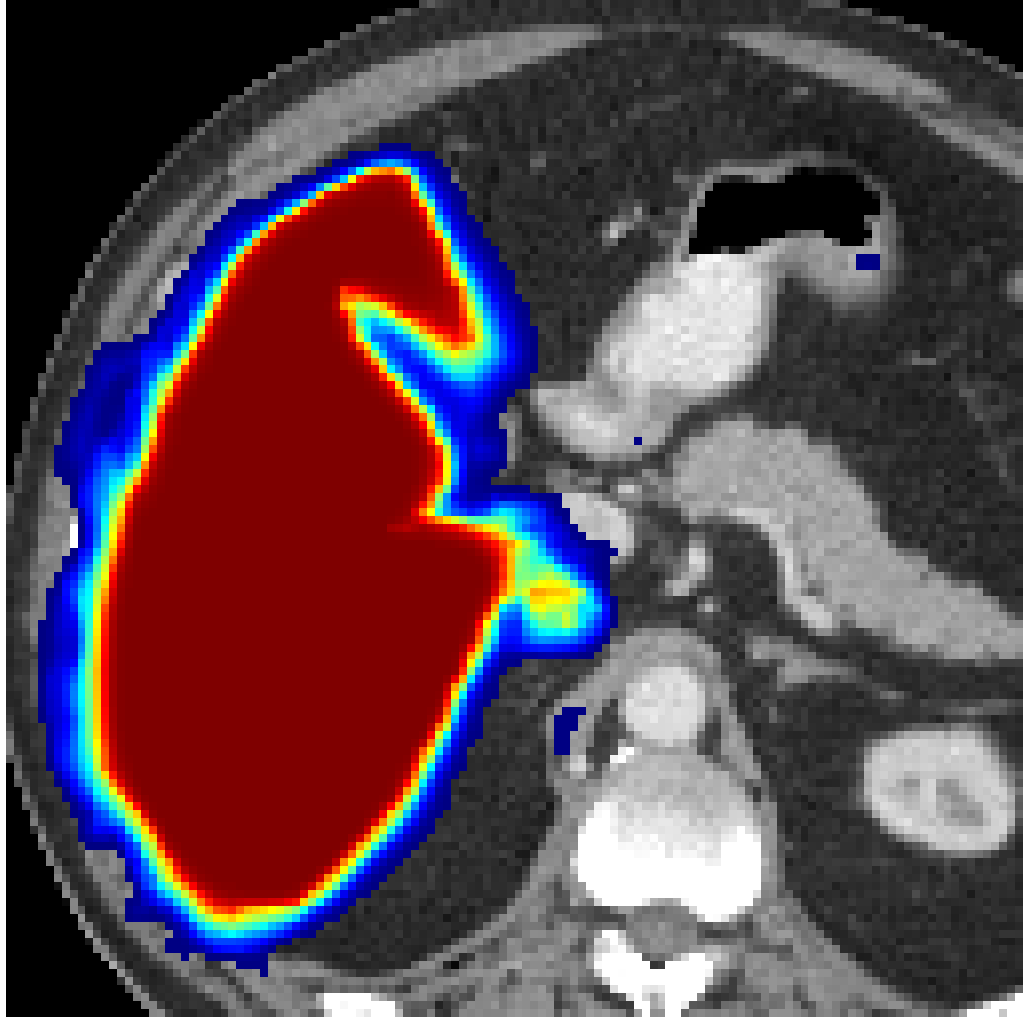}} &
		\hfill
		\subfloat{\adjincludegraphics[valign=c,height=\figheight]{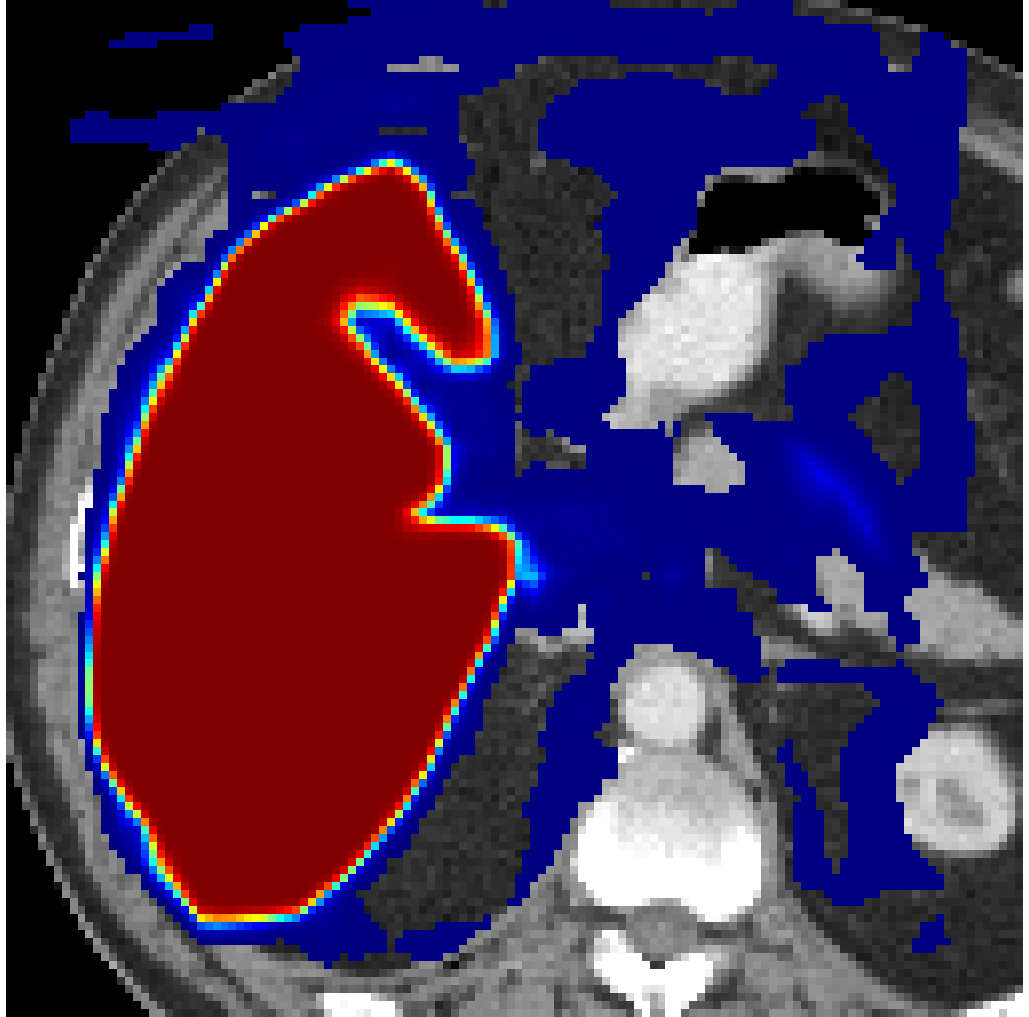}} \\
		%%%%%%%%%%%%%%%%%%%%%%%%%
	    %%%%%%%%%%%%%%%%%%%%%%%%%
	    \scriptsize\mopancreas \hfill &
		\subfloat{\adjincludegraphics[valign=c,height=\figheight]{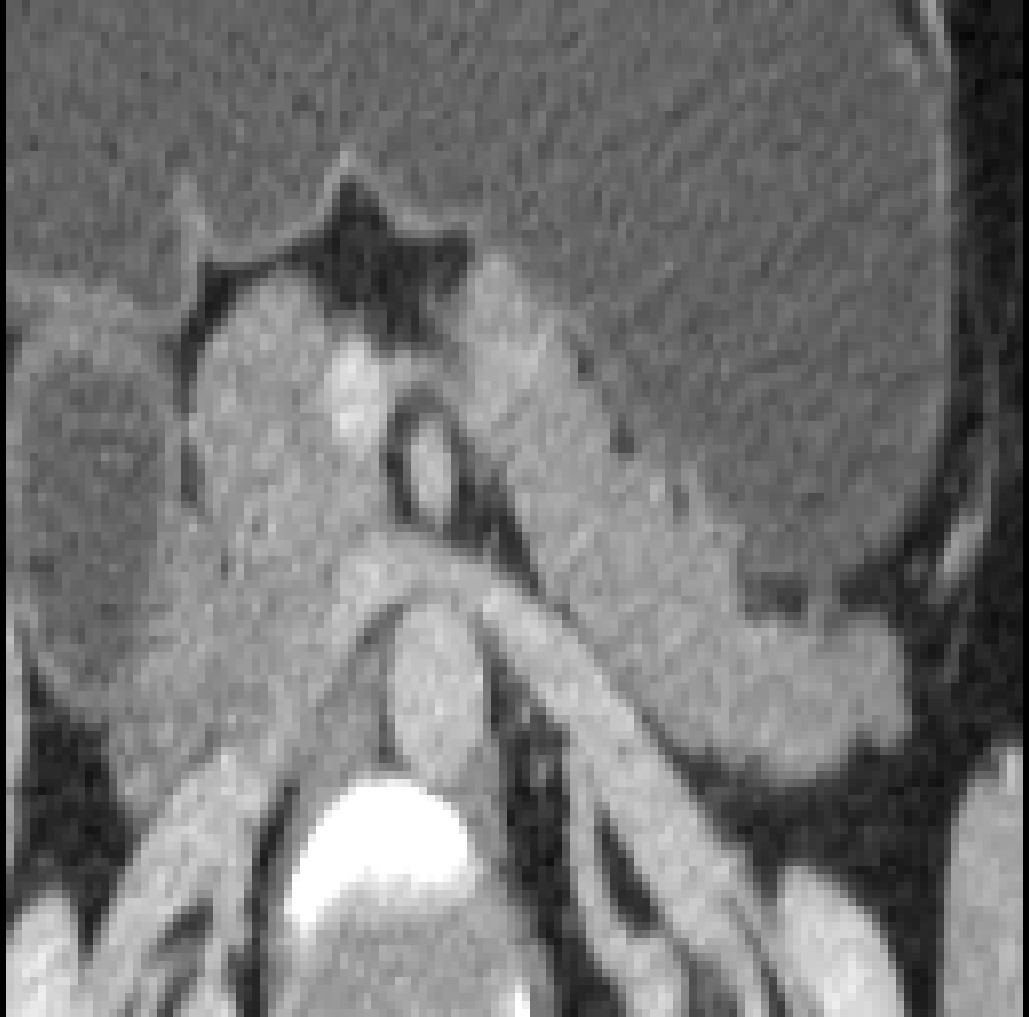}} &
		\hfill
		\subfloat{\adjincludegraphics[valign=c,height=\figheight]{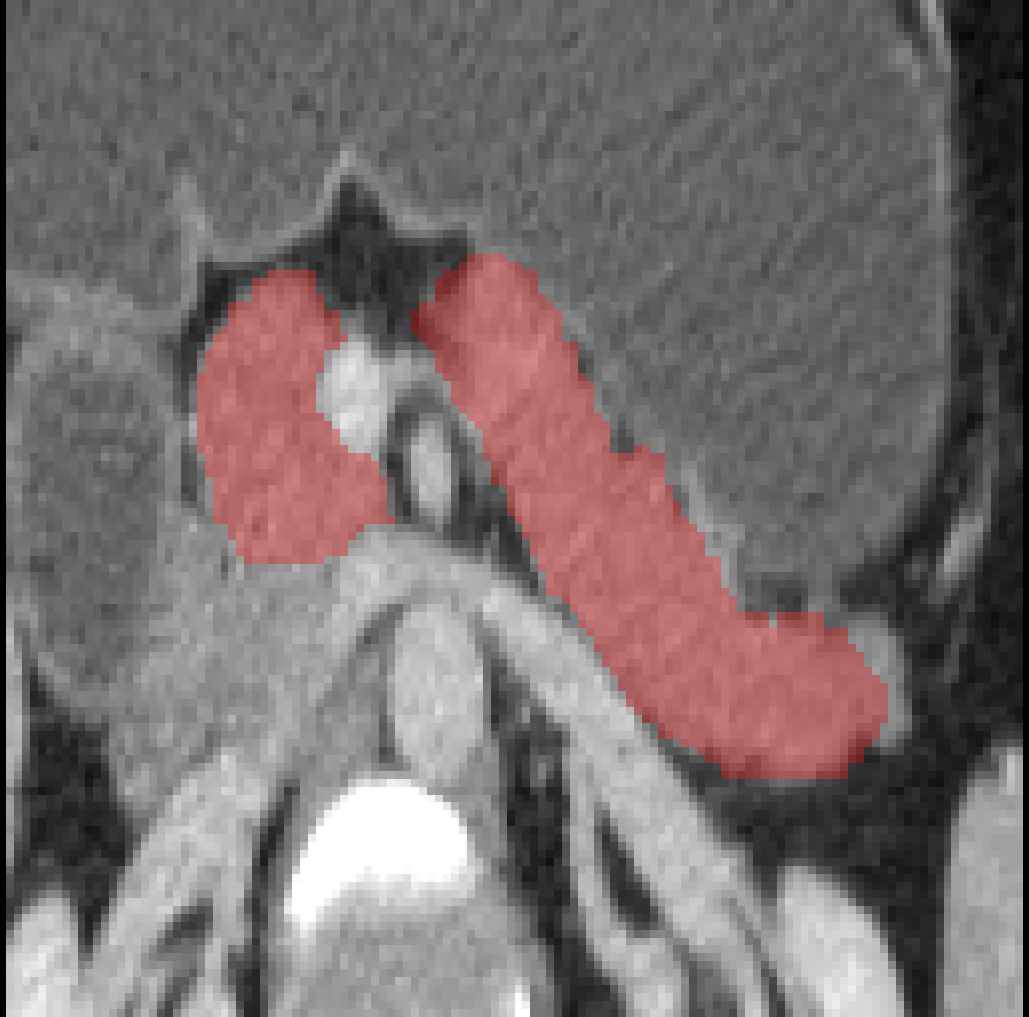}} & 
		\hfill
		\subfloat{\adjincludegraphics[valign=c,height=\figheight]{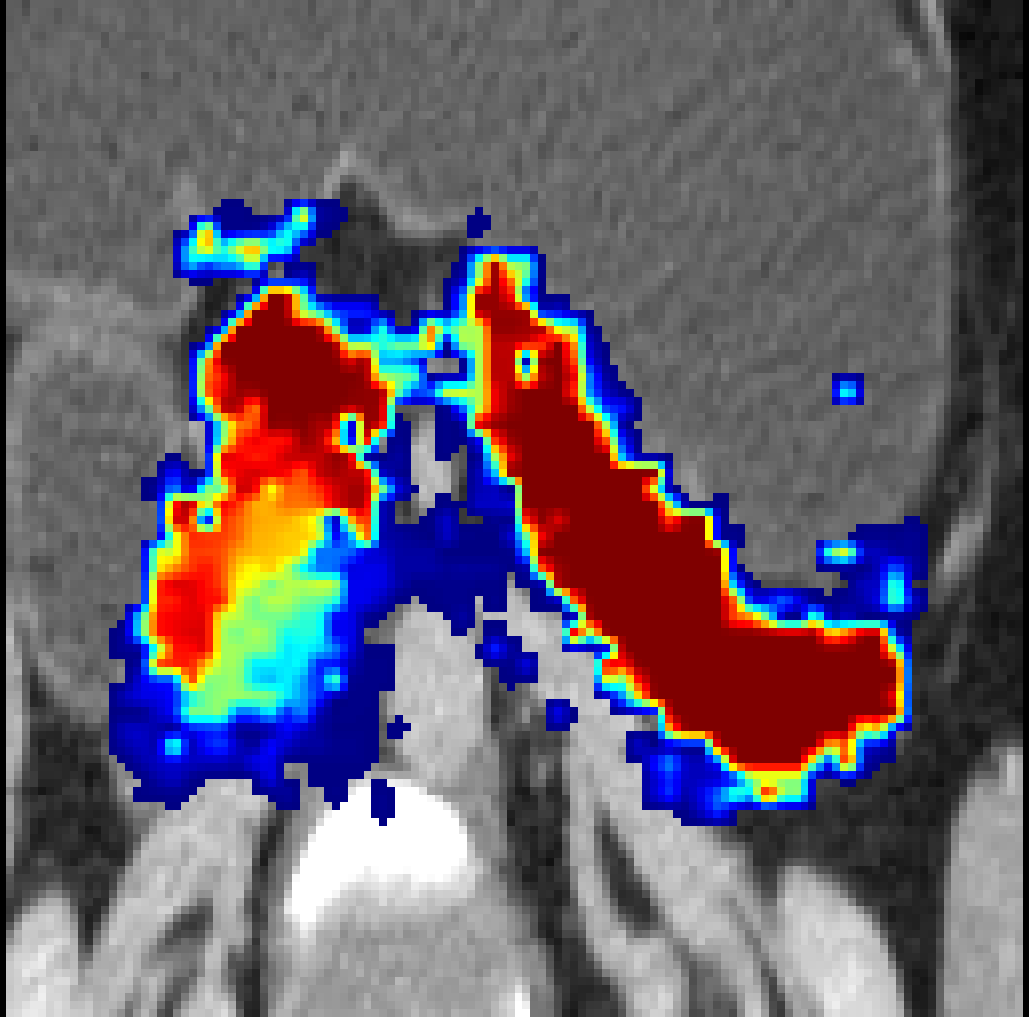}} &
		\hfill
		\subfloat{\adjincludegraphics[valign=c,height=\figheight]{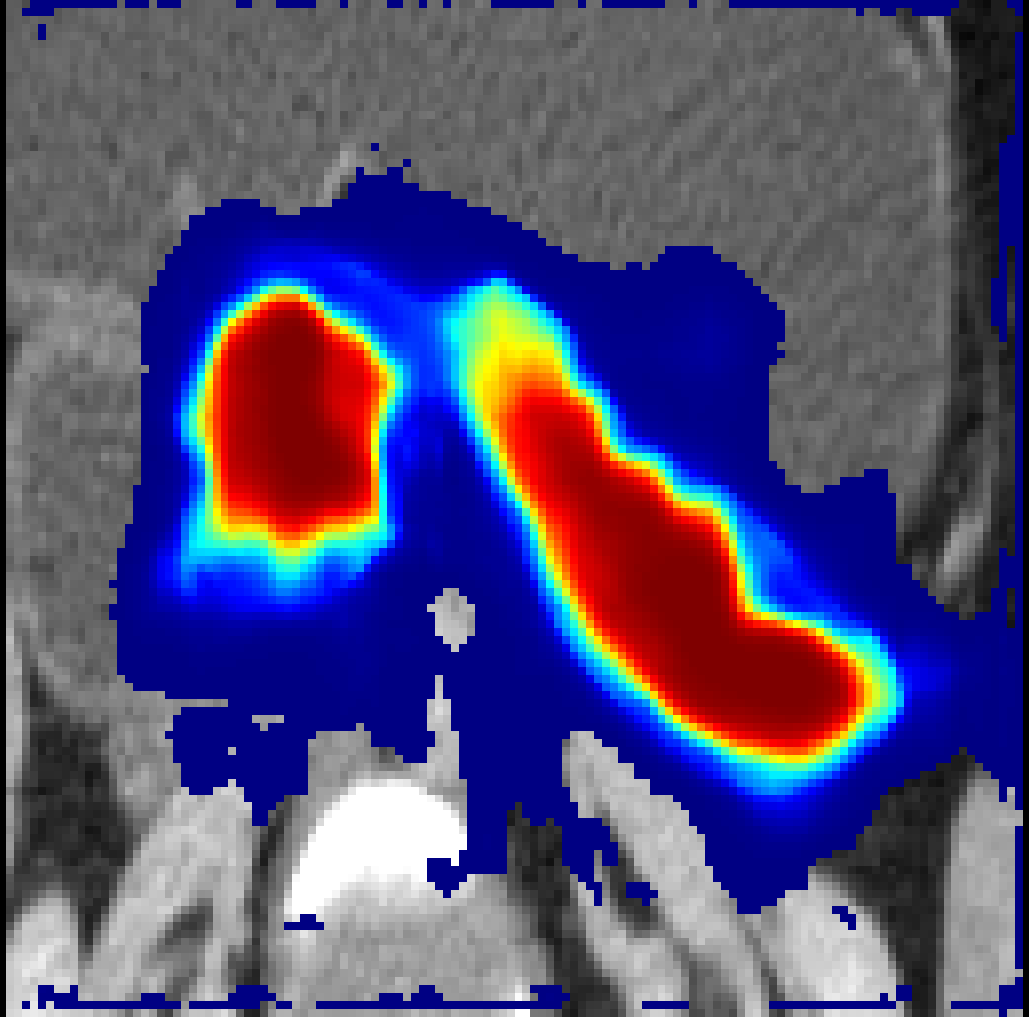}} &
		\hfill
		\subfloat{\adjincludegraphics[valign=c,height=\figheight]{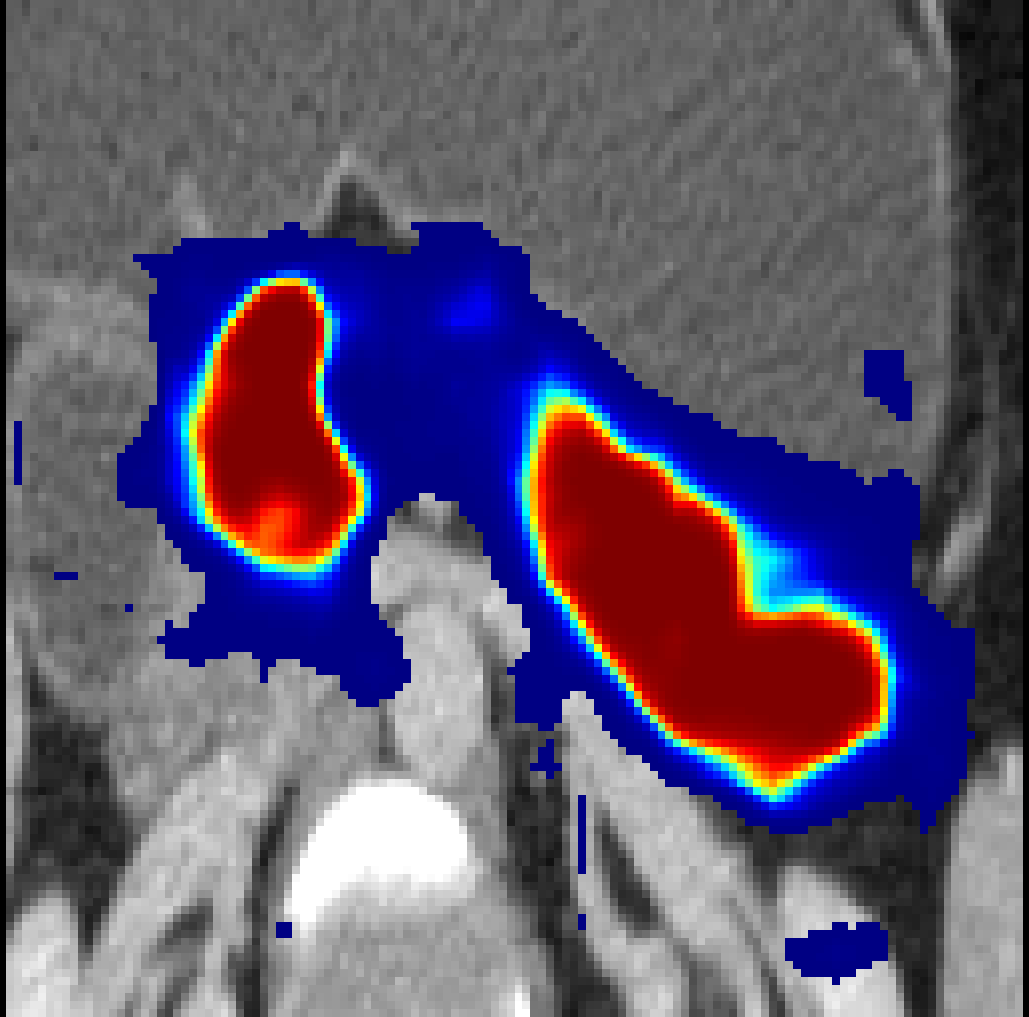}} \\
		%%%%%%%%%%%%%%%%%%%%%%%%%
	    %%%%%%%%%%%%%%%%%%%%%%%%%
	    \scriptsize\molkidney \hfill &
		\subfloat{\adjincludegraphics[valign=c,height=\figheight]{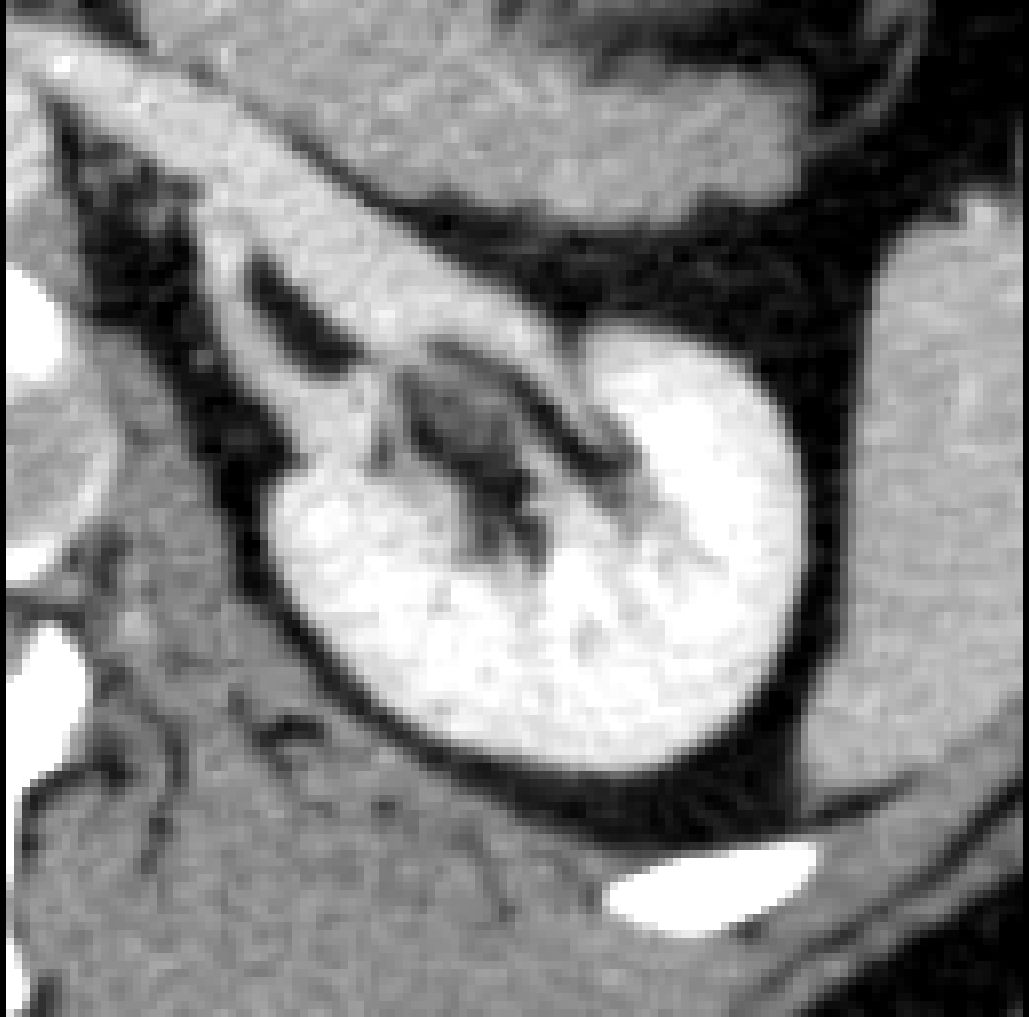}} &
		\hfill
		\subfloat{\adjincludegraphics[valign=c,height=\figheight]{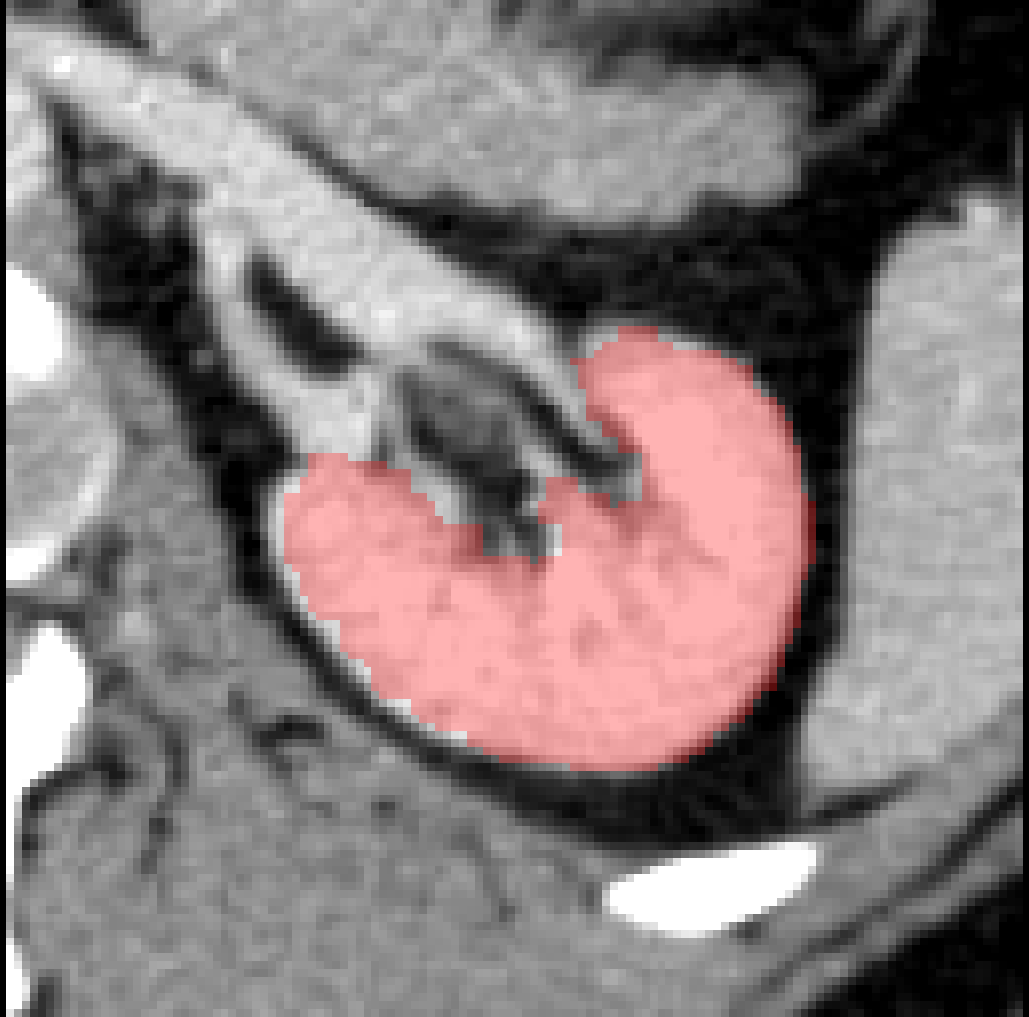}} & 
		\hfill
		\subfloat{\adjincludegraphics[valign=c,height=\figheight]{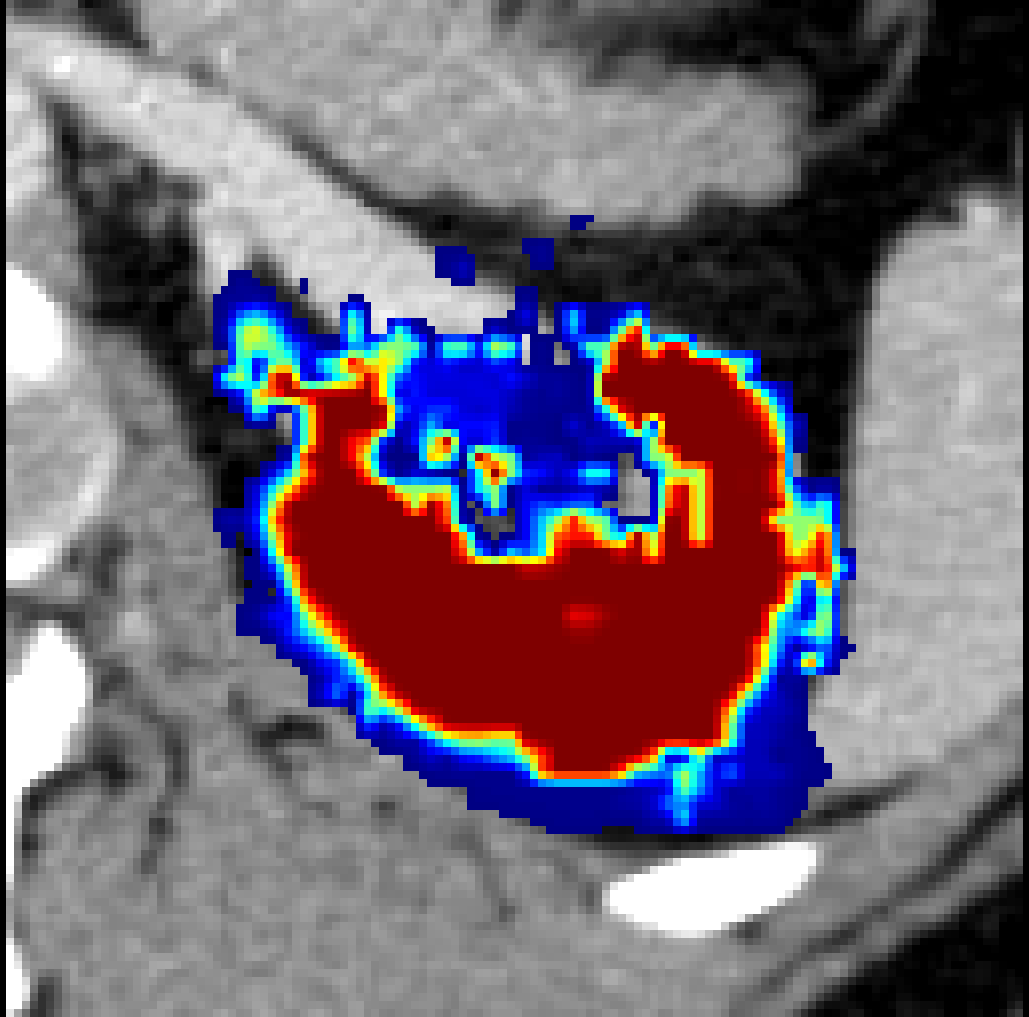}} &
		\hfill
		\subfloat{\adjincludegraphics[valign=c,height=\figheight]{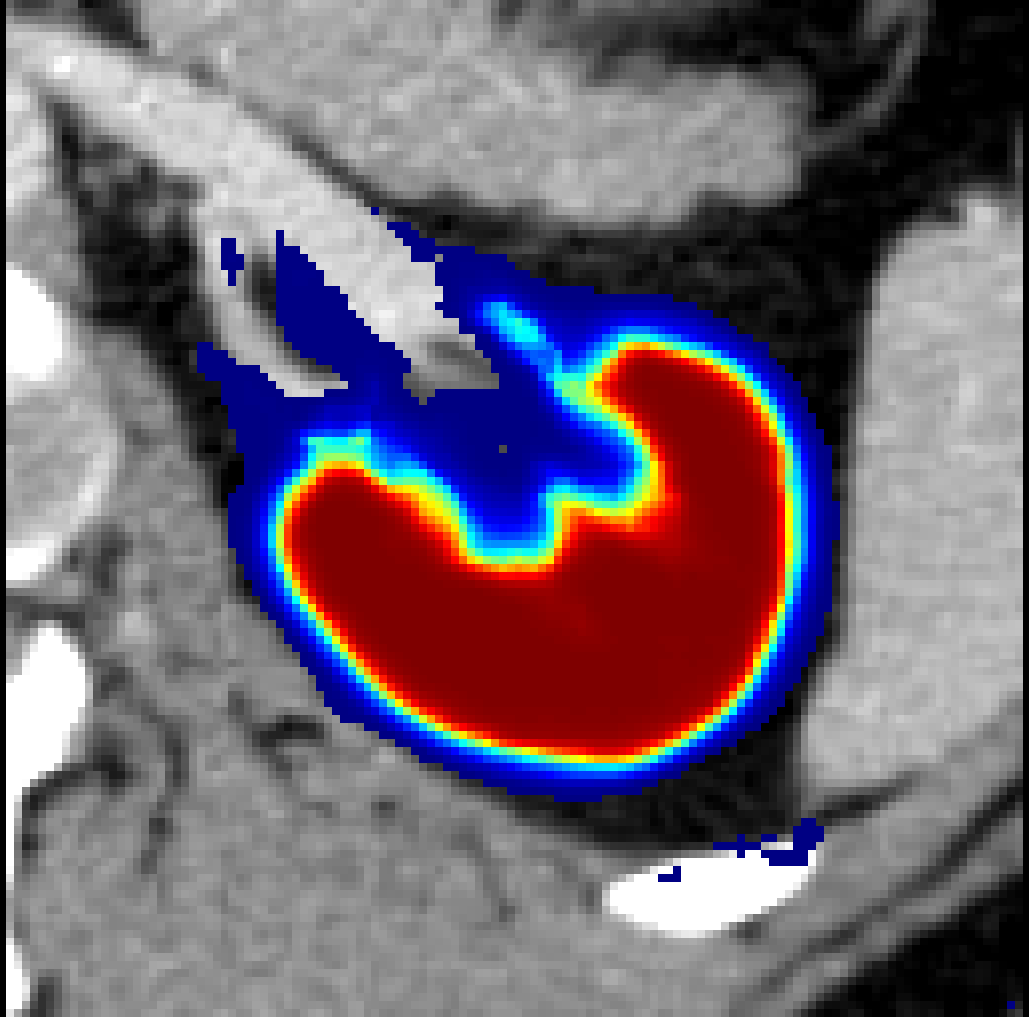}} &
		\hfill
		\subfloat{\adjincludegraphics[valign=c,height=\figheight]{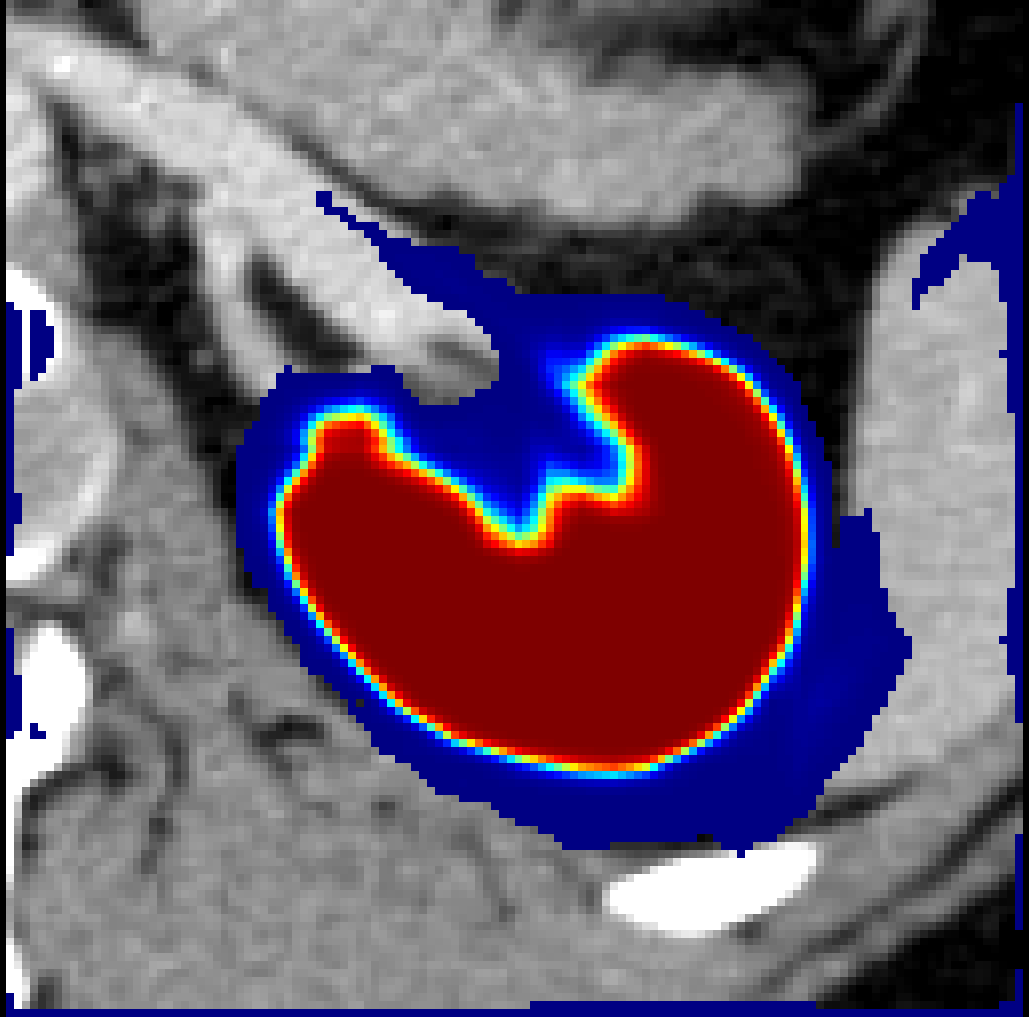}} \\
		%%%%%%%%%%%%%%%%%%%%%%%%%
	    %%%%%%%%%%%%%%%%%%%%%%%%%
	    \scriptsize\mogallbladder \hfill &
		\subfloat[(a)]{\adjincludegraphics[valign=c,height=\figheight]{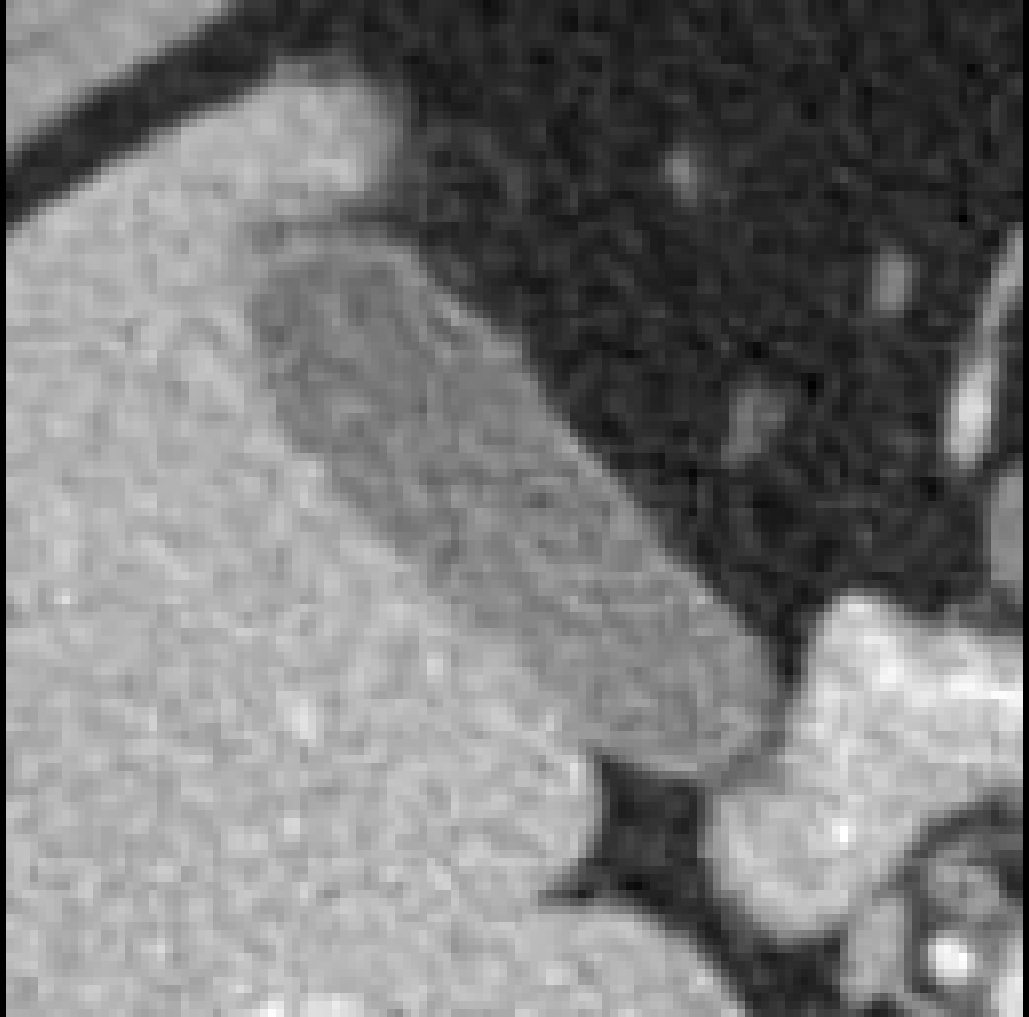}} &
		\hfill
		\subfloat[(b)]{\adjincludegraphics[valign=c,height=\figheight]{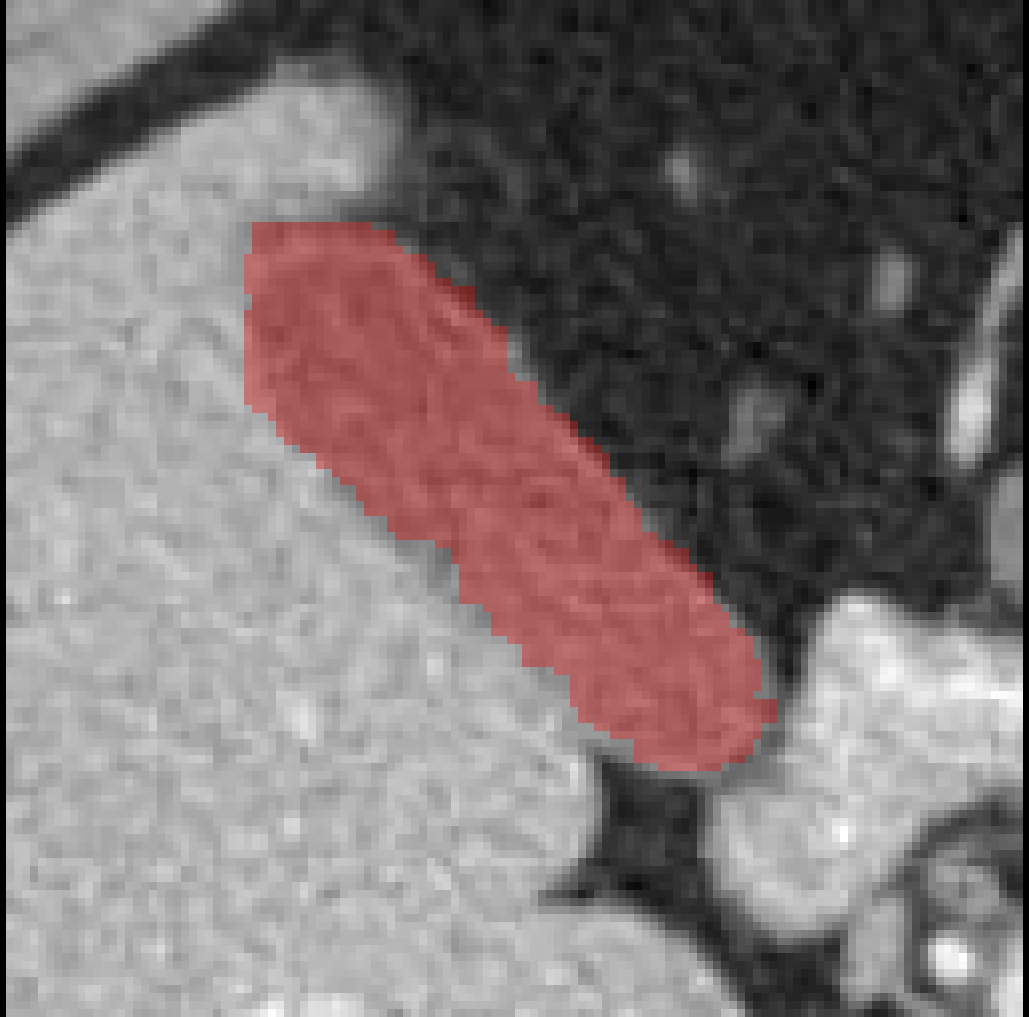}} & 
		\hfill
		\subfloat[(c)]{\adjincludegraphics[valign=c,height=\figheight]{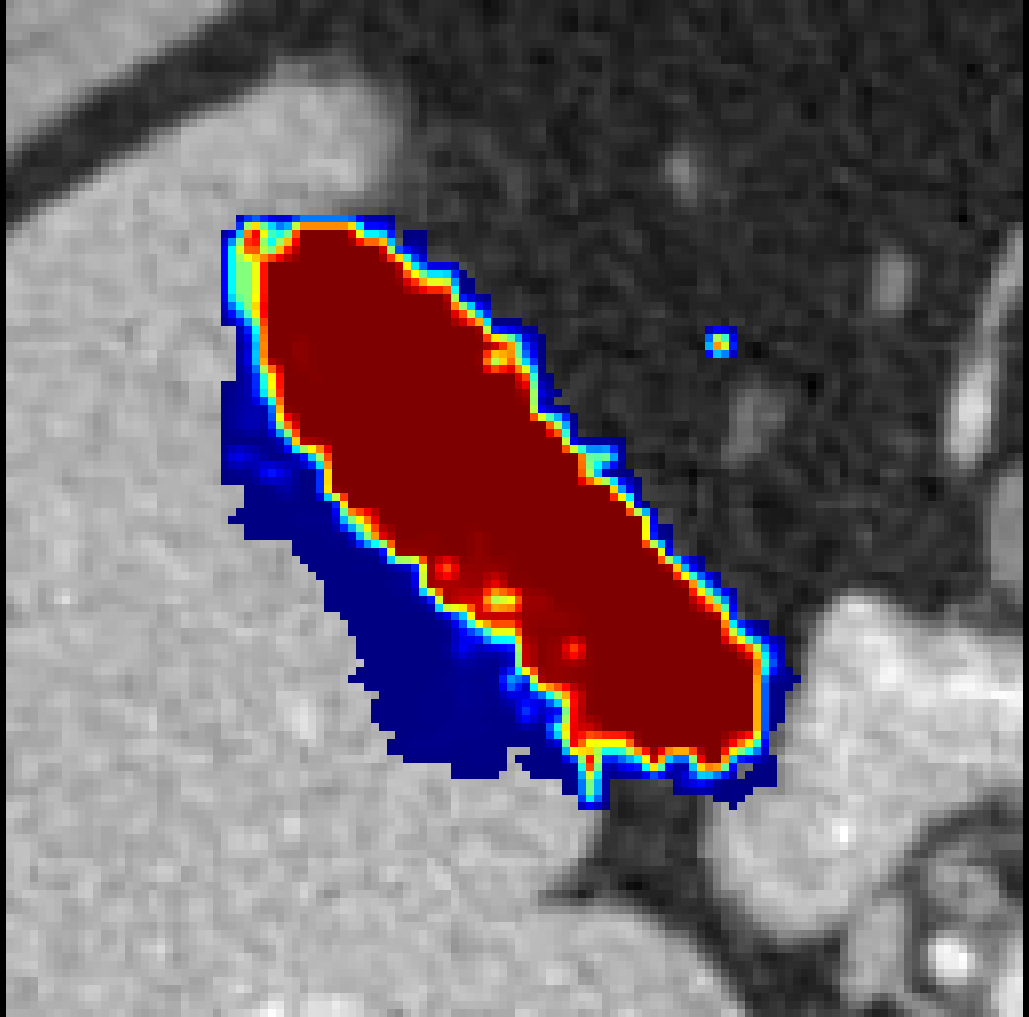}} &
		\hfill
		\subfloat[(d)]{\adjincludegraphics[valign=c,height=\figheight]{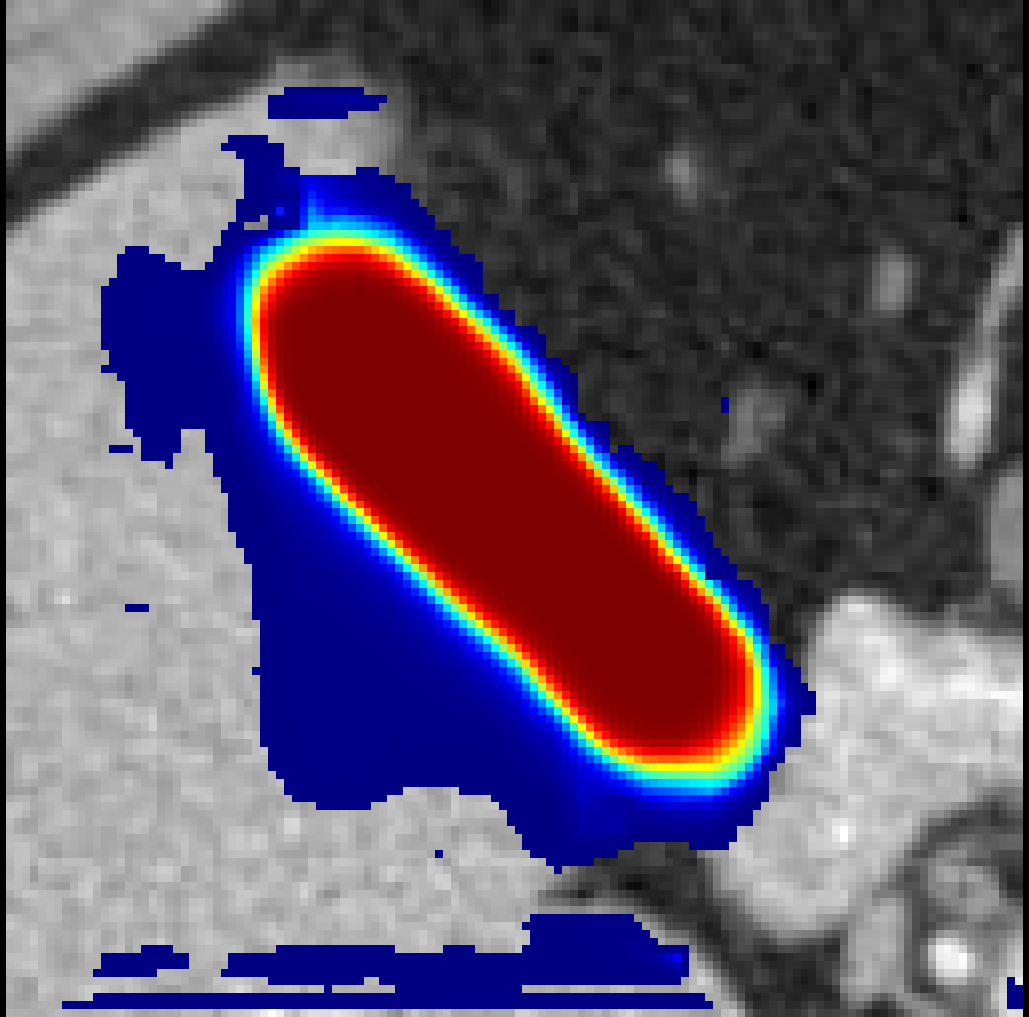}} &
		\hfill
		\subfloat[(e)]{\adjincludegraphics[valign=c,height=\figheight]{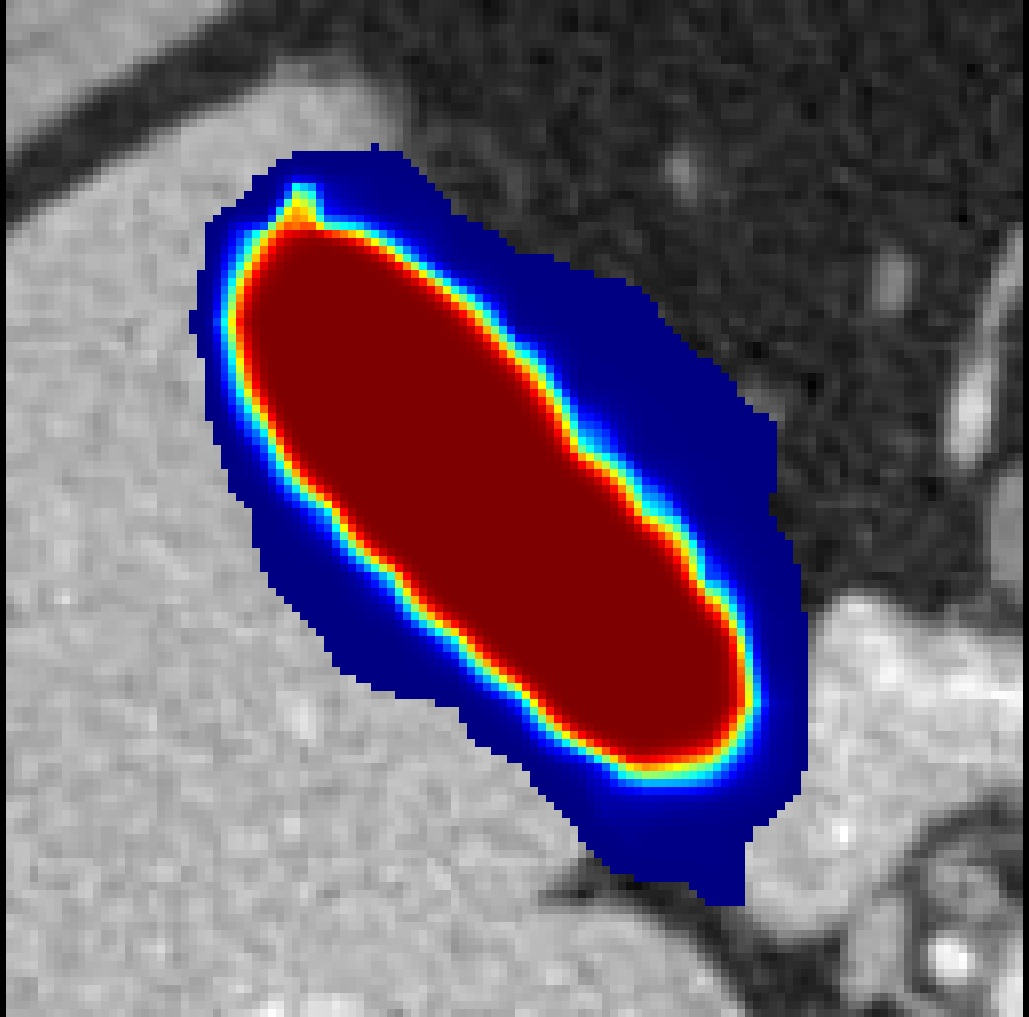}} \\
		%%%%%%%%%%%%%%%%%%%%%%%%%%%%%%%%%%%%%%%%%%%%%%%%%%%%%%%%%%%%%%%%%%%%%%%%%%		
	\end{tabular}
	\caption{Our results on six different segmentation tasks on example cases from the validation set. We show (a) the image after cropping based on extreme points, (b) overlaid (full) ground truth (used for evaluation only), (c) initial random walker prediction, (d) our final segmentation result produced by the weakly supervised segmentation scheme, (e) the fully supervised result for reference. Specifically, we compare example cases for \textit{weak. sup. dextr3D (w RW) Dice + Point loss + Point Attn} and \textit{fully sup. dextr3D Dice loss} for (d) and (e), respectively. The probability maps are scaled between 0 and 1 and we show all non-zero probabilities.
	\label{fig:results}}
\end{figure*}
%%%%%%%%%%%%%%%%%%%%%%%%%%%%%%%%%%%%%%%%%%%%%%%%%%%%%%%%%%%%%%%%%
%%%%%%%%%%%%%%%%%%%%%%%%%%%%%%%%%%%%%%%%%%%%%%%%%%%%%%%%%%%%%%%%%%%%%%%%%%%%%%%%%%%%%%%%%%%%%%%%
\begin{figure*}[htbp]
  \newcommand{\figheight}{1.8cm}
	\centering
	\begin{tabular}{llcrr}
	    %%%%%%%%%%%%%%%%%%%%%%%%%%%%%%%%%%%%%%%%%%%%%%%%%%%%%%%%%%%%%%%%%%%%%%%%%%
		\subfloat{\adjincludegraphics[valign=c,height=\figheight]{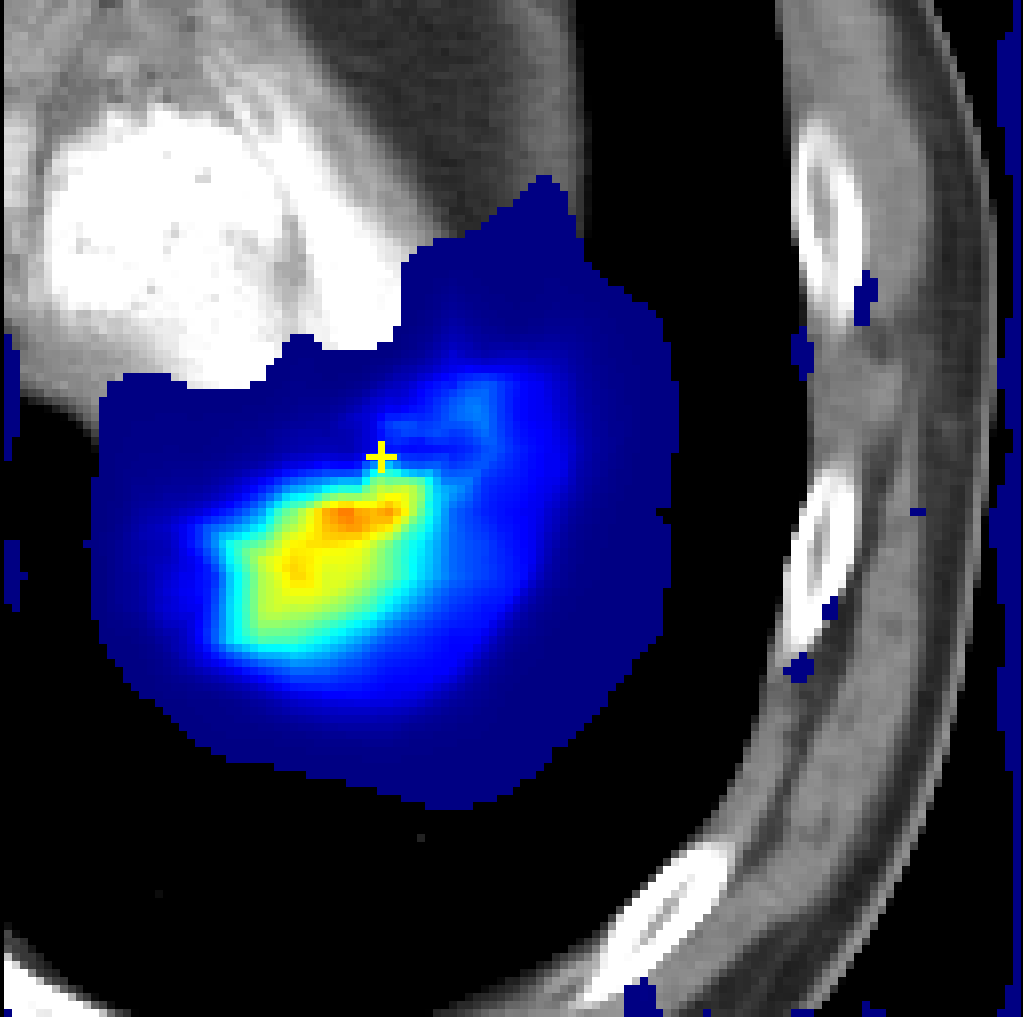}} &
		\hfill
		\subfloat{\adjincludegraphics[valign=c,height=\figheight]{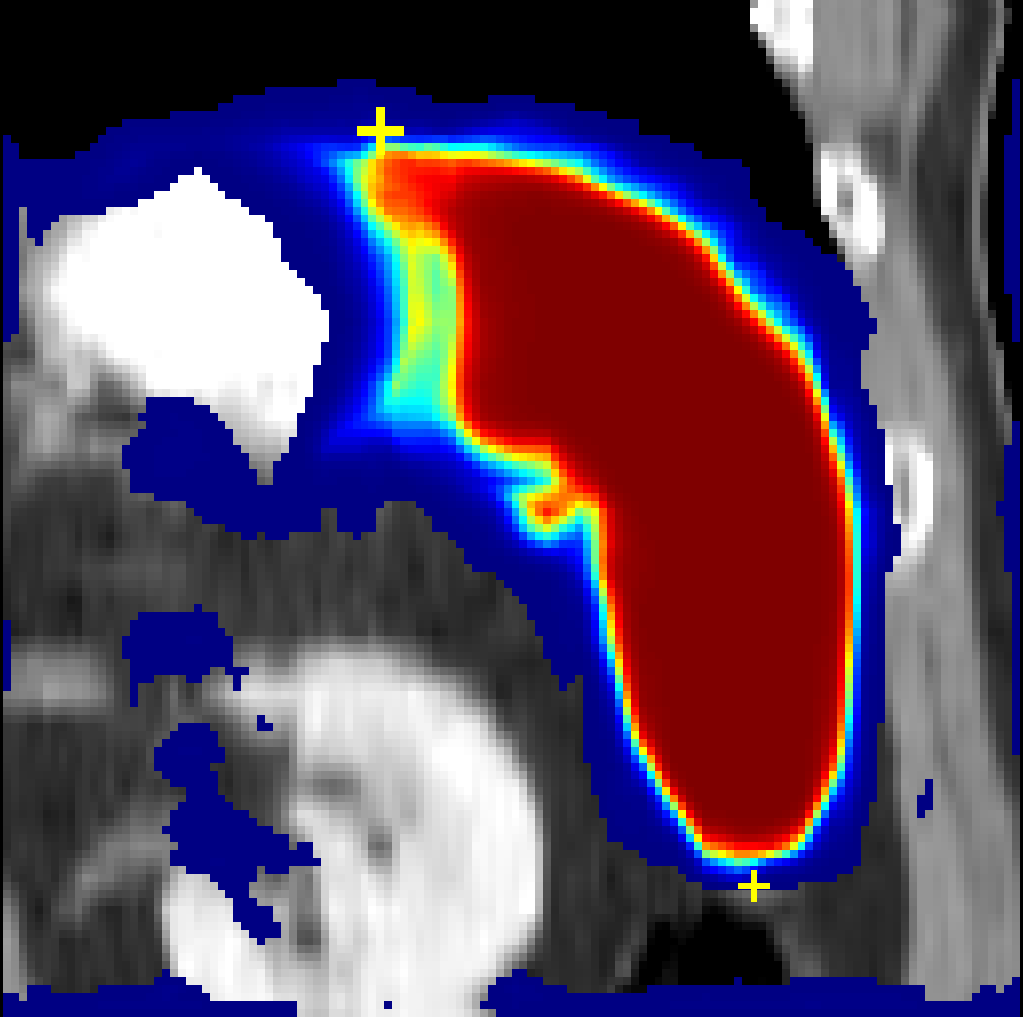}} &
		\hspace{3cm}&
		\hfill
		\subfloat{\adjincludegraphics[valign=c,height=\figheight]{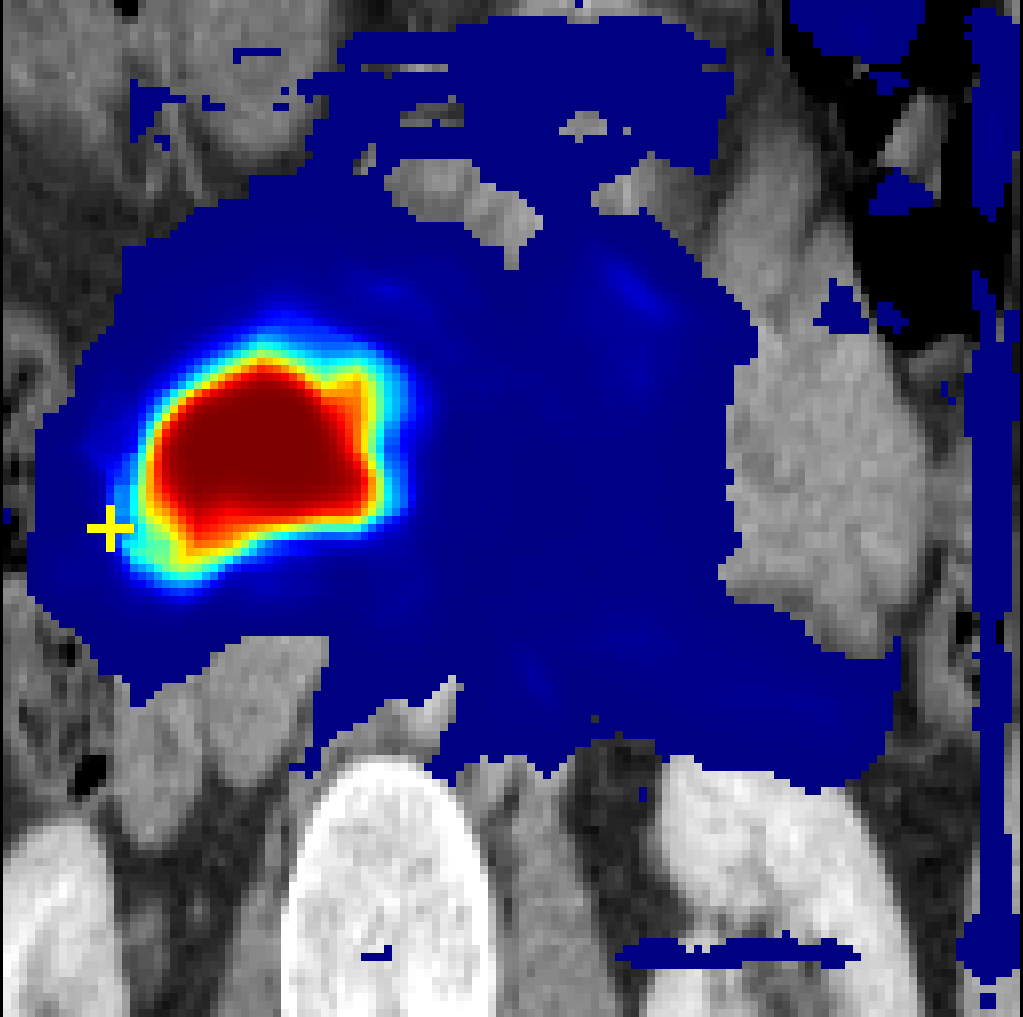}} &
		\hfill
		\subfloat{\adjincludegraphics[valign=c,height=\figheight]{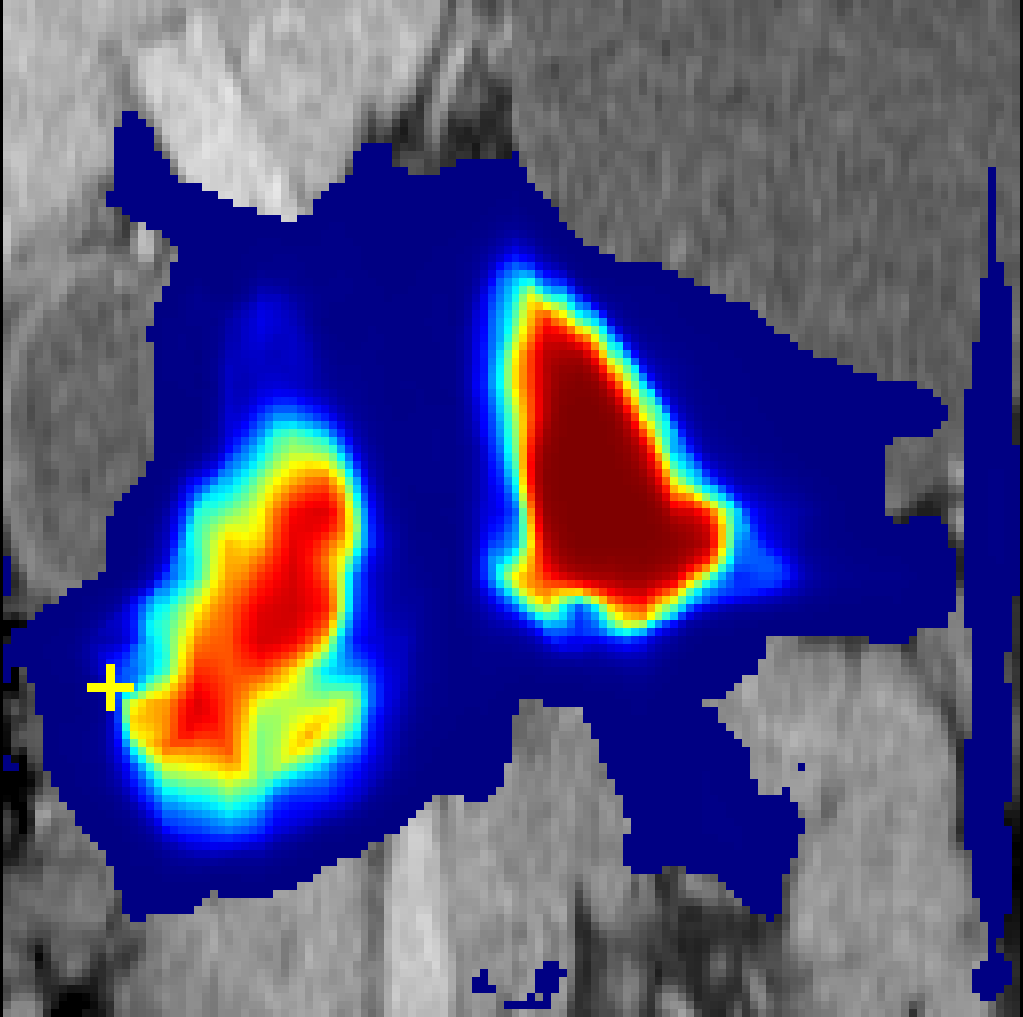}} \\ 
		%%%%%%%%%%%%%%%%%%%%%%%%%
	    %%%%%%%%%%%%%%%%%%%%%%%%%
		\subfloat{\adjincludegraphics[valign=c,height=\figheight]{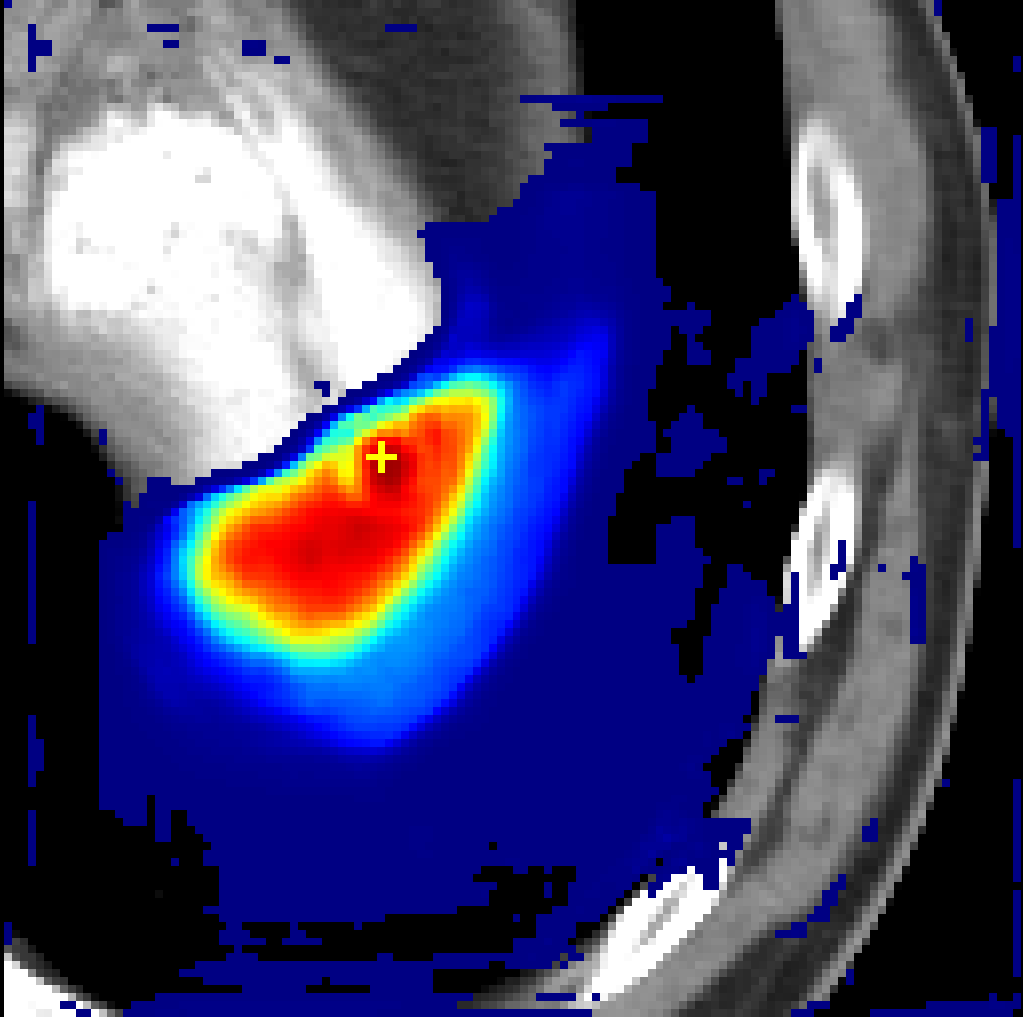}} &
		\hfill
		\subfloat{\adjincludegraphics[valign=c,height=\figheight]{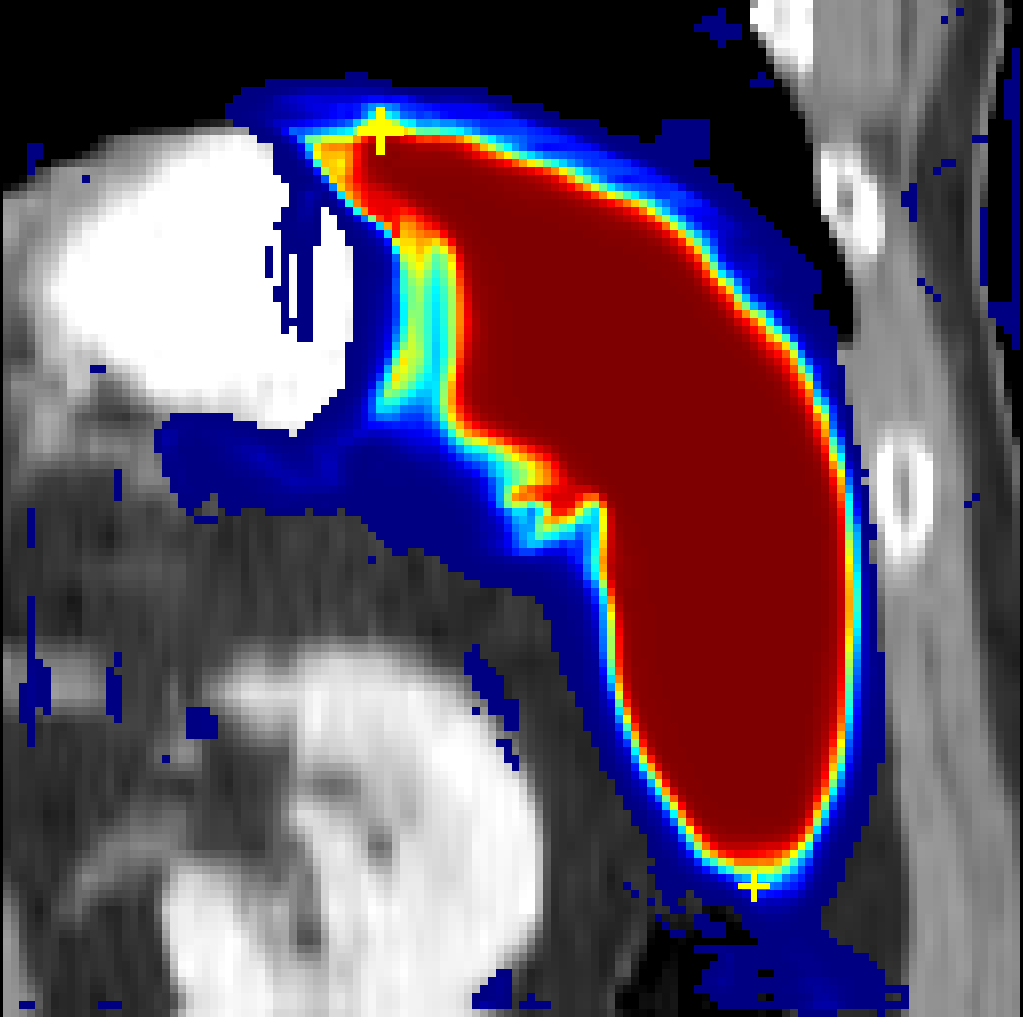}} &
		\hspace{3cm}&
		\hfill
		\subfloat{\adjincludegraphics[valign=c,height=\figheight]{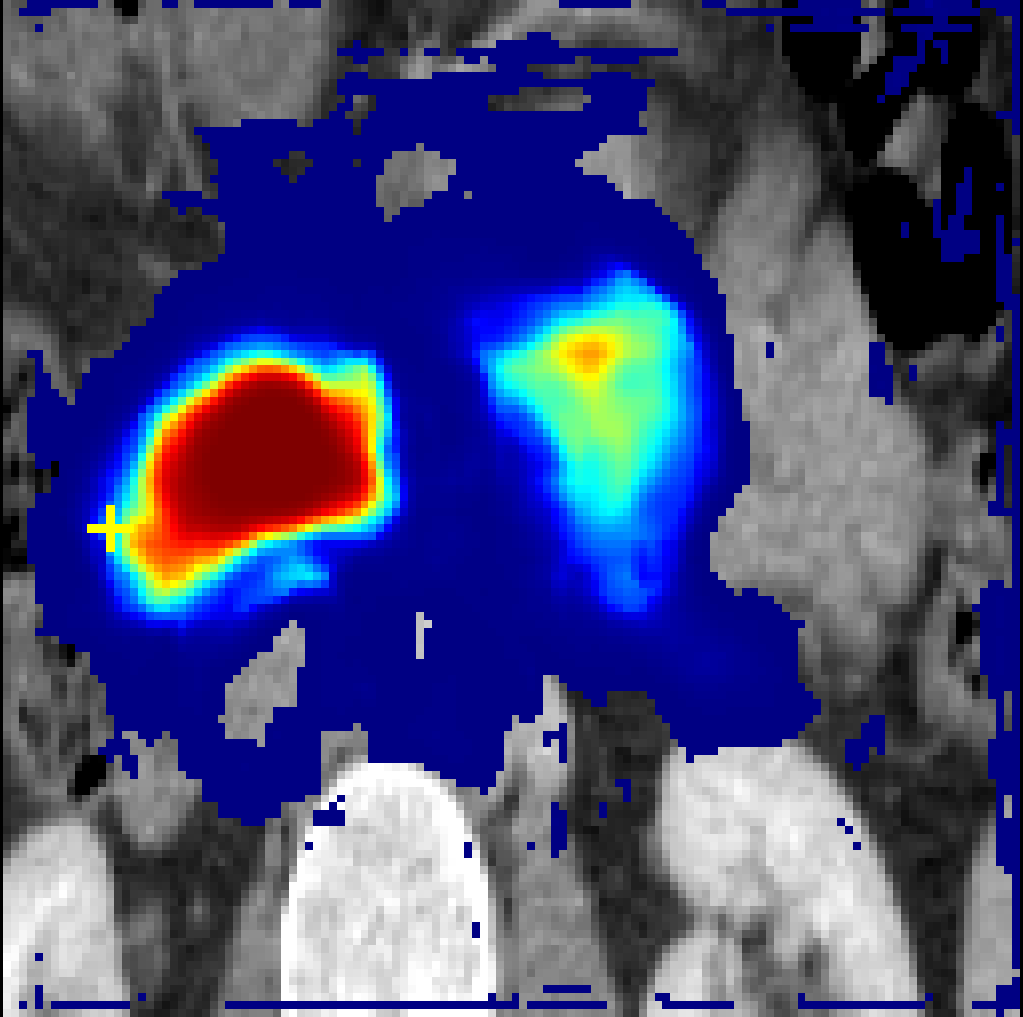}} &
		\hfill
		\subfloat{\adjincludegraphics[valign=c,height=\figheight]{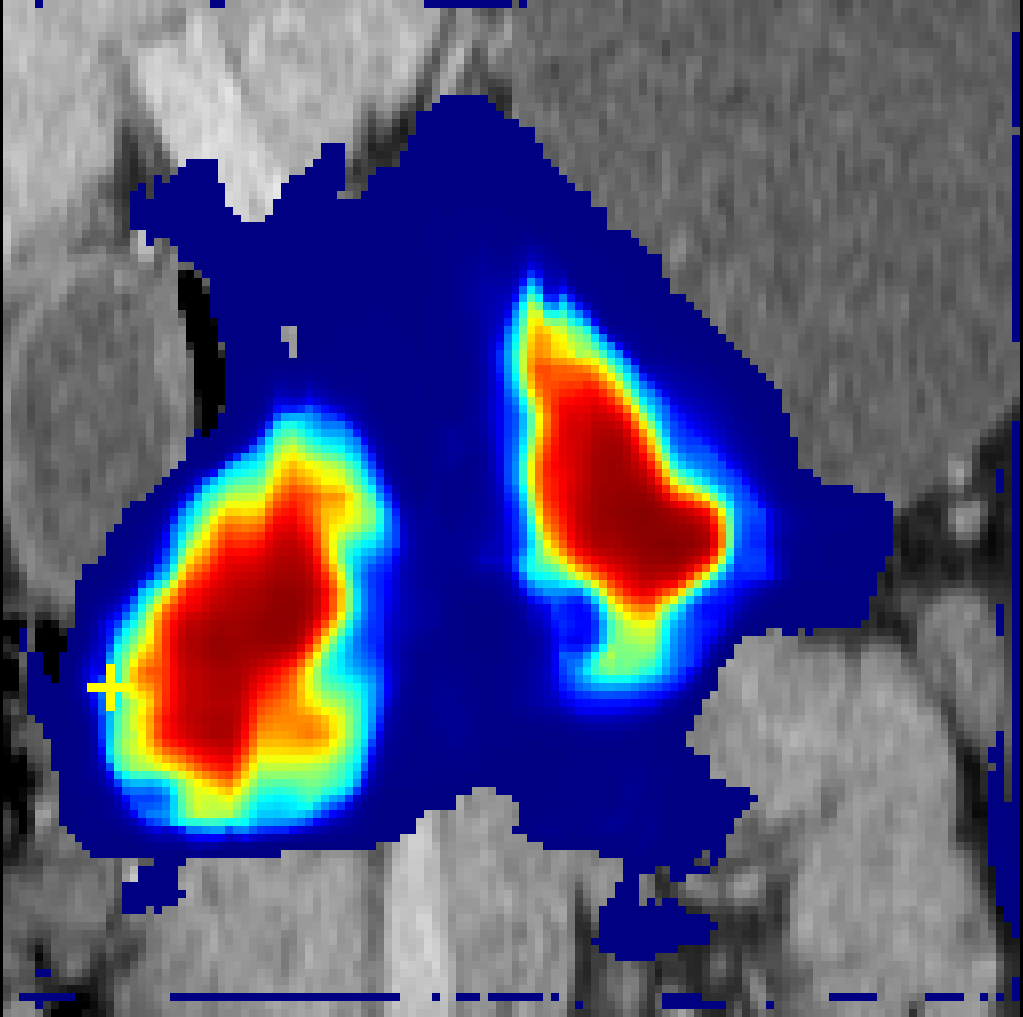}} 
		%%%%%%%%%%%%%%%%%%%%%%%%%%%%%%%%%%%%%%%%%%%%%%%%%%%%%%%%%%%%%%%%%%%%%%%%%%		
	\end{tabular}
	\caption{The impact of adding the point loss and point attention to our weakly supervised models. We show the results of the \textbf{top:} \textit{weak. sup. dextr3D (w RW) Dice}; and \textbf{bottom:} \textit{weak. sup. dextr3D (w RW) Dice + Point loss + Point Attn} settings. Examples from the \msdspleen~(\textbf{left}) and \mopancreas~(\textbf{right}) datasets are shown, respectively. The clicked extreme points are shown by a yellow cross. Best viewed in color. The predictions learned together with the point loss do lie markedly closer to the clicked point locations.
	\label{fig:point_loss_results}}
\end{figure*}
\begin{figure*}[htbp]
	\centering
	\newcommand{\figwidth}{0.5\textwidth}
	\begin{tabular}{cc}
		\subfloat[\scriptsize\moliver]{\includegraphics[valign=c,width=\figwidth]{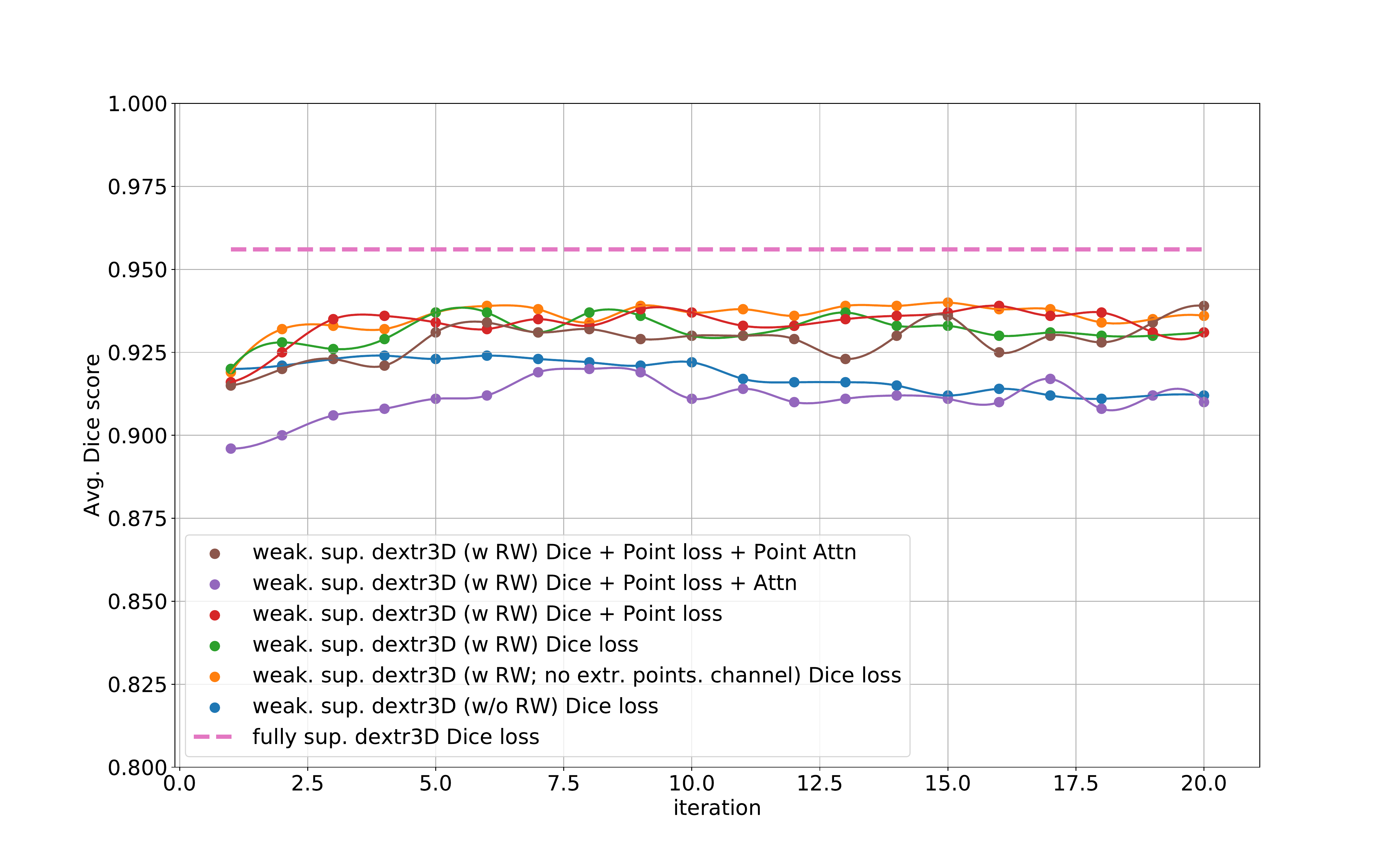}} &
		\subfloat[\scriptsize\mopancreas]{\includegraphics[valign=c,width=\figwidth]{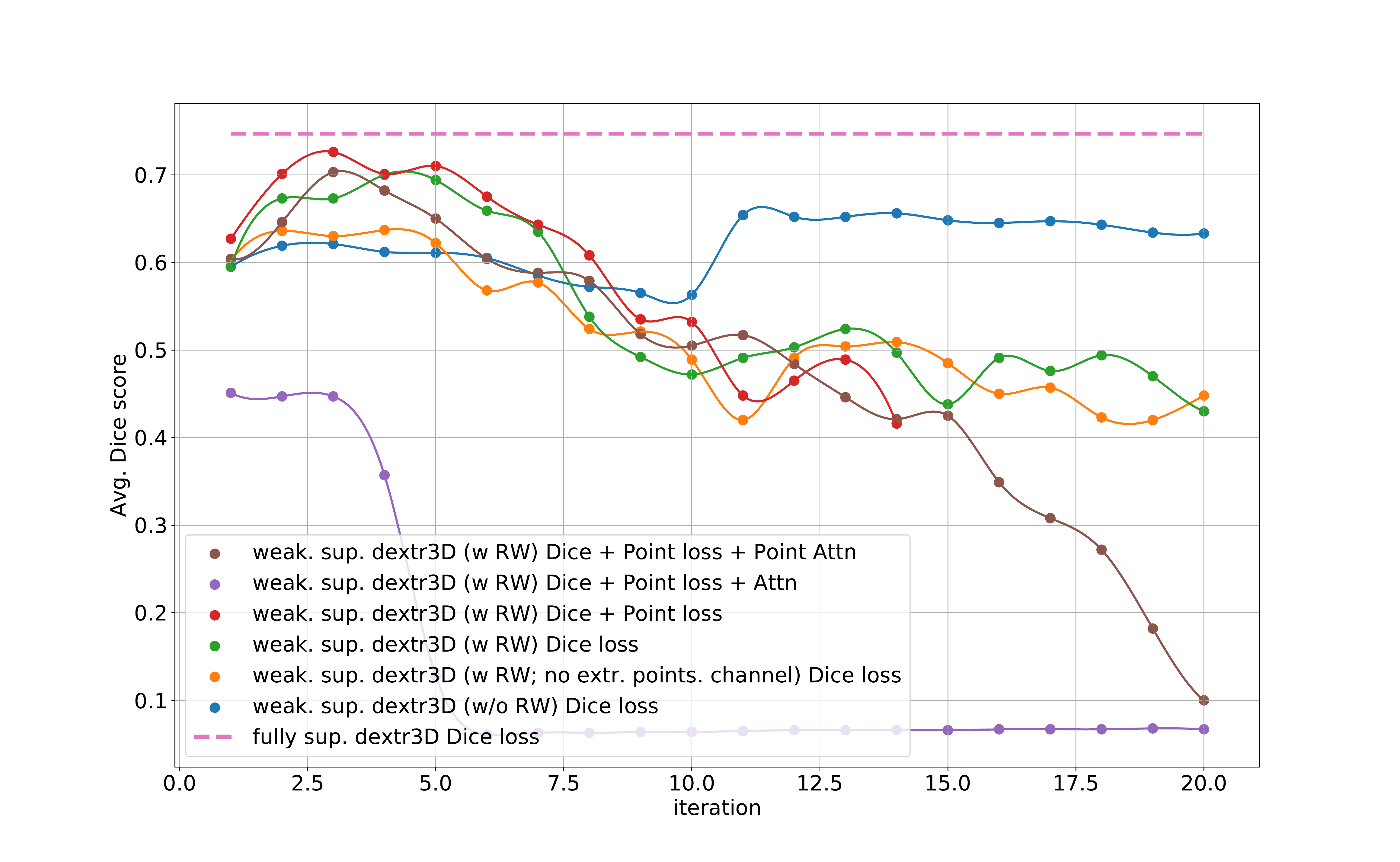}}
	\end{tabular}
	\caption{Weakly supervised training from random walker initialization. For illustration, we only show the \moliver~ and \mopancreas~segmentation tasks with the varying training settings as shown in our ablation study of Table \ref{table:results} at each round of deep network training. While the performance of the \moliver~models generally improves with the number of iterations, it can also be noticed that for \mopancreas~, a poor initialization by the random walker can cause the models to degrade quickly. Notice, that adding the point channel information results in a more stable training behavior.
	\label{fig:convergence}}
\end{figure*}

\begin{table*}[htbp]
%\begin{sidewaystable*}[htbp]
    \caption{Summary of our weakly supervised segmentation results. This table compares the random walker initialization (\textit{rnd. walk. init.}) with
    the weakly supervised training approach using Dice loss (DL) (\textit{weak. sup. dextr3D (w/o RW) DL}), 
    without the extra point channel information as input to the network (\textit{weak. sup. dextr3D (w RW; no extr. points. channel) DL}),
    when using the point channel as input (\textit{weak. sup. dextr3D (w RW) DL}),
    using the proposed point loss (PL) together with DL (\textit{weak. sup. dextr3D (w RW) Dice + PL}),
    integrating the attention mechanism as in \citet{oktay2018attention} (\textit{weak. sup. dextr3D (w RW) Dice + PL + Attn.}),
    attention with point channel information at attention gates (\textit{weak. sup. dextr3D (w RW) Dice + PL + Pt. Attn.}), 
    and fully sup. (\textit{dextr3D DL}) for reference on different datasets.\label{table:results}}
      \centering
      %\footnotesize
      \scriptsize
%\begin{tabular}{|p{20mm}|l|l|l|l|l|l|l|}      
\newcommand{\colwidth}{0.2\columnwidth}
\begin{tabular}{|p{\colwidth}|p{\colwidth}|p{\colwidth}|p{\colwidth}|p{\colwidth}|p{\colwidth}|p{\colwidth}|p{\colwidth}|}
\hline
\textbf{Dice [mean$\pm$std (median)]}& \textbf{\msdspleen}& \textbf{\mospleen}& \textbf{\moliver}& \textbf{\mopancreas}& \textbf{\molkidney}& \textbf{\mogallbladder}\\
\hline\hline
rnd. walk. init.&	0.922$\pm$0.018 (0.922)& 	0.830$\pm$0.144 (0.913)&	0.786$\pm$0.146 (0.847)&	0.458$\pm$0.206 (0.414)&	0.741$\pm$0.137 (0.815)&	0.638$\pm$0.195 (0.619)\\
\hline
weak. sup. (w/o RW) DL&	0.939$\pm$0.011 (0.943)&	0.942$\pm$0.009 (0.939)&	0.924$\pm$0.020 (0.924)&	0.656$\pm$0.089 (0.634)&	0.878$\pm$0.034 (0.893)& 0.678$\pm$0.194 (0.740)\\	
\hline
weak. sup. (w RW; no extr. points. channel)DL&	0.945$\pm$0.012 (0.950)&	0.942$\pm$0.009 (0.937)&	0.940$\pm$0.011 (0.942)&	0.637$\pm$0.166 (0.664)&	0.900$\pm$0.013 (0.899)& 0.677$\pm$0.252 (0.787)\\	
\hline
weak. sup. (w RW)DL&	0.946$\pm$0.011 (0.950)&	0.944$\pm$0.023 (0.945)&	0.937$\pm$0.013 (0.941)&	0.700$\pm$0.068 (0.676)&	0.909$\pm$0.017 (0.907)& 0.701$\pm$0.209 (0.795)\\	
\hline
weak. sup. (w RW)Dice + PL&	0.946$\pm$0.010 (0.949)&	\textbf{0.945$\pm$0.019 (0.947)}&	\textbf{0.939$\pm$0.012 (0.940)}&	\textbf{0.726$\pm$0.080 (0.746)}&	0.906$\pm$0.024 (0.909)& \textbf{0.719$\pm$0.186 (0.789)}\\
\hline
weak. sup. (w RW)Dice + PL + Attn.&	0.945$\pm$0.013 (0.948)&	0.924$\pm$0.053 (0.948)&	0.920$\pm$0.059 (0.943)&	0.451$\pm$0.124 (0.427)&	0.905$\pm$0.023 (0.907)& 0.606$\pm$0.256 (0.632)\\	
\hline
weak. sup. (w RW)Dice + PL + Pt. Attn.&	\textbf{0.948$\pm$0.011 (0.950)}&	\textbf{0.945$\pm$0.021 (0.943)}&	\textbf{0.939$\pm$0.013 (0.939)}&	0.703$\pm$0.077 (0.688)&	\textbf{0.913$\pm$0.013 (0.916)}& 0.702$\pm$0.184 (0.773)\\
\hline
\textbf{fully sup. DL}&	\textbf{{\textit0.958$\pm$0.007 (0.959)}}&	\textbf{{\textit0.954$\pm$0.027 (0.959)}}&	\textbf{{\textit0.956$\pm$0.010 (0.957)}}&	\textbf{\textit{0.747$\pm$0.082 (0.721)}}&	\textbf{\textit{0.942$\pm$0.019 (0.946)}}& \textbf{\textit{0.715$\pm$0.173 (0.791)}}\\	
\hline
\end{tabular}
\end{table*}
%\end{sidewaystable*}

%%%%%%%%%%%%%%%%%%%%%%%%%%%%%%%%%%%%%%%%%%%%%%%%%%%%%%%%%%%%%%%%%
%%%%%%%%%%%%%%%%%%%%%%%%%%%%%%%%%%%%%%%%%%%%%%%%%%%%%%%%%%%%%%%%%
%%%%%%%%%%%%%%%%%%%%%%%%%%%%%%%%%%%%%%%%%%%%%%%%%%%%%%%%%%%%%%%%%
%%%%%%%%%%%%%%%%%%%%%%%%%%%%%%%%%%%%%%%%%%%%%%%%%%%%%%%%%%%%%%%%%
\paragraph{Datasets} 
We utilize the training datasets (as they include ground truth annotations) from public resources, specifically, from the multi-organ (MO) segmentation study in \citet{gibson2018automatic}\footnote{\url{https://zenodo.org/record/1169361\#.XcsiOHFKi90}} which provided annotation for abdominal CT data from previously published datasets: \citet{roth2015deeporgan}\footnote{\url{https://wiki.cancerimagingarchive.net/display/Public/Pancreas-CT}}~and~  \citet{btcv2015}\footnote{\url{https://www.synapse.org/\#!Synapse:syn3193805/wiki/217752}}. Furthermore, we utilize data from the \textit{Medical Segmentation Decathlon} (MSD) challenge \citep{simpson2019large}\footnote{\url{http://medicaldecathlon.com}}. From MO, we utilize the spleen, liver, pancreas, left kidney, and gallbladder segmentation masks, denoted as \mospleen, \moliver, \mopancreas, \molkidney, and \mogallbladder, respectively. From MSD, we include the spleen mask, denotes as \msdspleen. Qualitative results are shown in Fig. \ref{fig:results} for each segmentation task on example cases from the validation set. For MO, we use a constant data split of 81 training and 9 validation cases, respectively. For MSD, there are 32 training and 9 validation cases, respectively, available.
\paragraph{Experiments} 
In all cases, we iterate our algorithm for a maximum of 20 iterations as shown in Fig. \ref{fig:convergence}. In Table \ref{table:results}, we compare training with and without using random walker (RW) regularization after each round of 3D FCN learning. In addition, by running the framework with RW regularization but without the extreme points channel, we quantify the benefit of modeling the extreme points as an extra input channel to the network versus only using the bounding box as in \citet{rajchl2017deepcut}. 
It can be further observed that the greatest changes occur after initial random walker segmentation in the first round of FCN training. While the average Dice score is not always enhanced by random walker regularization alone, it helps to incorporate enough ``novelty'' into our learning system to boost the overall Dice score in later iterations as shown in Fig. \ref{fig:convergence}. We furthermore, show the average Dice scores on the validation set after convergence when utilizing the proposed point loss, point loss plus attention gates as in \citet{oktay2018attention}, and a setting when using the point information as an additional guiding feature to the attention gates. The fully supervised case using Dice loss with the strong label ground truth masks are shown for reference.
It can be observed that utilizing the point channel information in the point loss function and the attention gates generally improves the performance of the model. The addition of point loss and point attention works best in four out of six weakly supervised cases, while the addition of point loss alone showed an advantage in two out of the six tasks. Notice, that the average Dice score in the \mogallbladder~task even outperforms the fully supervised setting.
\paragraph{Implementation} 
The training and evaluation of the deep neural networks used in the proposed framework were implemented based on the \textit{NVIDIA Clara Train SDK}\footnote{\url{https://developer.nvidia.com/clara}} using 4 NVIDIA Tesla V100 GPUs with 16 GB memory for each round of training. All models were trained using the deterministic training setup in \textit{Tensorflow}\footnote{\url{https://github.com/NVIDIA/tensorflow-determinism}} with the same random seed initialization in order to guarantee comparable results between the different variations of training.
For the random walker algorithm, we use the default parameters\footnote{\url{https://scikit-image.org/docs/dev/auto_examples/segmentation/plot_random_walker_segmentation.html}}.
%%%%%%%%%%%%%%%%%%%%%%%%%%%%%%%%%%%%%%%%%%%%%%%%%%%%%%%%%%%%%%%%%
%\begin{figure*}[htbp]
%\begin{center}
    %\includegraphics[width=1.0\textwidth]{figs/chart.pdf}
%\end{center}
%\caption{Weakly supervised training from random walker initialization. Each segmentation task is shown with (w) and without (w/o) random walker regularization after each round of FCN training.}
%\label{fig:training}
%\end{figure*}

\paragraph{Analysis of point loss} 
An analysis of the impact of the point loss on our weakly supervised models' predictions is shown in Fig. \ref{fig:point_loss_results}.

\section{Discussion}
\label{sec:discussion}
\noindent We provided a method for weakly-supervised 3D segmentation from extreme points. Asking the user to simply on the surface of the organ in each spatial dimension can drastically reduce the cost of labeling. The point clicks can simultaneously identify the region of interest and simplify the 3D machine learning task. The extreme points can also be used to create an initial noisy pseudo label based on the extreme points using the random walker algorithm. From our experiments, it can be observed that this initialization is relatively robust for six different tasks from medical image segmentation.

%\paragraph{Limitations:}
Occasionally, the random walker may lack robustness for organs with very diverse interior textures or highly concave curved shapes, for example, the pancreas (see \mopancreas~task in Table \ref{table:results}). In this situation, the shortest path result might sometimes lie outside the organ. A boundary search algorithm might provide a better initial segmentation here. Still, the initial segmentation can be significantly enhanced by the first round of FCN training.
In this study, we utilized one dataset (\msdspleen) as our development set and kept the hyperparameters of the full approach constant across different segmentation tasks and datasets. One might achieve better performance when optimizing the hyperparameters, especially for the initial random walker, based on the task at hand.
In practice, we performed model selection for each round of training in our approach based on the pseudo labels $\hat{Y}$ alone. However, we do need a fully annotated validation set to practically evaluate the overall convergence of our iterative approach for it to be clinically useful. One could use the predictions of the first round of FCN training to build an ML-based annotation tool that could speed up the creation of such a hold-out ``gold standard'' validation dataset and reduce the amount of manual labeling and editing needed in total.

%\paragraph{Conclusions:}
%\label{sec:conclusions}
Previous work primarily used boundary box annotations for weakly supervised learning in 2D/3D medical imaging, such as \citet{rajchl2017deepcut}. We consider, however, that selecting extreme points on the surface of the organ is more natural than selecting corners of a bounding box outside the organ of interest and more efficient than adding scribbles within and around the organ \citep{wang2018deepigeos,can2018learning}. This is consistent with recent findings in the computer vision literature \citep{papadopoulos2017extreme}. An application of the proposed approach to the 2D case would be straightforward.

We conducted a comprehensive ablation study of our proposed method in Table \ref{table:results}. Some of these settings are similar to previous work. For example, performing the network training without the extra point channel is equivalent to studies using bounding boxes alone such as in \citet{rajchl2017deepcut}. From Table \ref{table:results}, we can see that adding the additional point-click information in the loss and as attention mechanism is however beneficial while not increasing the labeling cost.
%However, direct comparison to other approaches is difficult due to the diverse datasets and hyperparameters employed in each prior work.
%There is a large body of prior art around interactive and weakly supervised segmentation.

In summary, we proposed a weakly-supervised 3D segmentation framework based on extreme point clicks. Experimentation on six datasets showed that the approach can achieve performance close to the fully supervised setting in four tasks and even outperforms the fully supervised training in one of them (\mogallbladder).
In the future, an automatic proposal network could assist the user with the region of interest and extreme point selection to further reduce the manual burden of medical image annotation.

{
%\clearpage
%\newpage
%\small
\bibliographystyle{ieee_fullname}
\bibliography{bibliography}
}

\end{document}